\documentclass[lettersize,journal]{IEEEtran}
\usepackage{amsmath,amsfonts}
\usepackage{algorithm}
\usepackage{algorithmic}
\usepackage{array}
\usepackage{textcomp}
\usepackage{stfloats}
\usepackage{url}
\usepackage{verbatim}
\usepackage{graphicx}
\usepackage{cite}
\usepackage{csquotes}
\usepackage{rotating}
\usepackage{multirow}
\usepackage{times}
\usepackage{epsfig}
\usepackage{color}
\usepackage[normalem]{ulem}
\usepackage{booktabs}
\usepackage{subfigure}
\usepackage{ulem}
\usepackage[colorlinks=true,linkcolor=black,citecolor=blue,urlcolor=red,]{hyperref}
\usepackage{soul}
\newcommand\eg{\emph{e.g.}} 
\newcommand\ie{\emph{i.e.}}

\newcommand\etcn{\emph{etc. }}

\begin{document}

%\title{Multi-Spectral Salient Object Detection}
\title{Mutual Information Regularization for\\ Weakly-supervised RGB-D Salient Object Detection}


\author{Aixuan Li,~
        Yuxin Mao,~
        Jing Zhang,~
        Yuchao Dai*\\
% <-this % stops a space
\IEEEcompsocitemizethanks{
\IEEEcompsocthanksitem Aixuan Li, Yuxin Mao and Yuchao Dai are with School of Electronics and Information, Northwestern Polytechnical University, Xi'an, China and State Key Laboratory of Integrated Services Networks (Xidian University).
\IEEEcompsocthanksitem Jing Zhang is with School of Computing, the Australian National University, Canberra, Australia (zjnwpu@gmail.com).
\IEEEcompsocthanksitem Yuchao Dai is the corresponding author (daiyuchao@nwpu.edu.cn).
This work was supported in part by the National Natural Science Foundation of China (No. 62271410).
}% <-this % stops an unwanted space
%\thanks{Manuscript received XXX; revised XXX.}

}

\markboth{Journal of \LaTeX\ Class Files,~Vol.~14, No.~8, August~2021}%
{Shell \MakeLowercase{\textit{et al.}}: A Sample Article Using IEEEtran.cls for IEEE Journals}


\maketitle

\begin{abstract}
In this paper, we present a weakly-supervised RGB-D salient object detection model via scribble supervision.
Specifically, as a multimodal learning task, we focus on effective multimodal representation learning via inter-modal mutual information regularization.
In particular, following the principle of disentangled representation learning, we introduce a mutual information upper bound with a mutual information minimization regularizer to encourage the disentangled representation of each modality for salient object detection. Based on our multimodal representation learning framework, we introduce an asymmetric feature extractor for our multimodal data, which is proven more effective than the conventional symmetric backbone setting. We also introduce multimodal variational auto-encoder as stochastic prediction refinement techniques, which takes pseudo labels from the first training stage as supervision and generates refined prediction. 
Experimental results on benchmark RGB-D salient object detection datasets verify both effectiveness of our explicit multimodal disentangled representation learning method and the stochastic prediction refinement strategy, achieving comparable performance with the state-of-the-art fully \emph{supervised} models. Our code and data are available at: \url{https://github.com/baneitixiaomai/MIRV}.

% to ease the pixel-wise labelling burden.
% with effective RGB-Depth representation learning.

% weakly-suppervised 

% and stochastic prediction refinement. 
% For the former, 

% to model the representation quality of RGB image and depth data, which we define as the multimodal fusion process. For the latter, we incorporate a second stage training strategy via multimodal variational auto-encoder as the stochastic refinement process, 

% prototype learning to the weakly-supervised learning framework to extract discriminative feature representation of each modal. Due to the missing structure information in the scribble supervision, we further adopt minimum spanning tree (MST) to achieve online saliency map updating, with which we aim to explore the contribution of unlabeled region for model updating. Given that the MST saliency updating process may introduce noise to the network, we further introduce a deterministic uncertainty estimation module to our framework, avoiding error propagation caused by the less accurate predictions for the unlabeled region. 
% Extensive experimental results verify the effectiveness of our weakly-supervised learning strategy in achieving comparable performance with the existing fully-supervised learning based counterparts.
\end{abstract}


% \section{Introduction}
% \IEEEPARstart{S}{alient} objects refer to the object that attract human attention the most.

\section{Introduction}
\label{sec:intro}
% \section{Introduction}
\IEEEPARstart{T}{he} advance of existing saliency detection techniques relies greatly on the large-scale pixel-wise labeled training datasets, \eg~DUTS~\cite{imagesaliency} and MSRA10K dataset~\cite{msra10k} for RGB saliency detection, and NJU2K~\cite{NJU2000}, NLPR~\cite{peng2014rgbd} and COME dataset~\cite{cascaded_rgbd_sod} for RGB-D saliency detection, where the labeling is usually time-consuming and expensive.
% , which is one of the main bottlenecks for saliency detection. 
% To relief the labeling effort, a
Alternatively,
% weak labels are used to train 
saliency detection models can be trained in a weakly-supervised manner, where the weak labels can be image-level labels~\cite{piao2021mfnet,li2018weakly,hsu2019weakly,to_be_critical,noise_sensitive_adv,self_supervised_saliency}, scribble supervisions~\cite{yu2021structure,jing2020weakly,xu2022weakly}, bounding boxes~\cite{liu2021weakly} and \etcn Considering both labeling efficiency and depth's contribution as structure-preserving guidance, we work on scribble-supervised RGB-D salient object detection.

% with a \enquote{1) estimation, 2) progressive refinement and 3) second stage refinement} pipeline.

% \begin{figure}[!htp]
% \begin{center}
%   \begin{tabular}{c@{ } }
%   {\includegraphics[width=0.9\linewidth]{figures/illustration_main.png}}\\
%   \end{tabular}
% \end{center}
% \caption{Illustration of our proposed pipeline for weakly-supervised RGB-D saliency detection--placeholder}.
% \label{model_illustration}
% \end{figure}

% Most of existing weakly-supervised saliency detection models are designed for RGB saliency detection. We find that the extra depth data is informative in providing both rich object structure information and scene geometric information, where both of these two advantages can be beneficial to bridge the gap between pixel-wise labeling the the weak labeling.
 
\begin{figure*}[!htp]
%  \vspace{-0.7mm}
\begin{center}
  \begin{tabular}{c@{ }}
  {\includegraphics[width=0.95\linewidth]{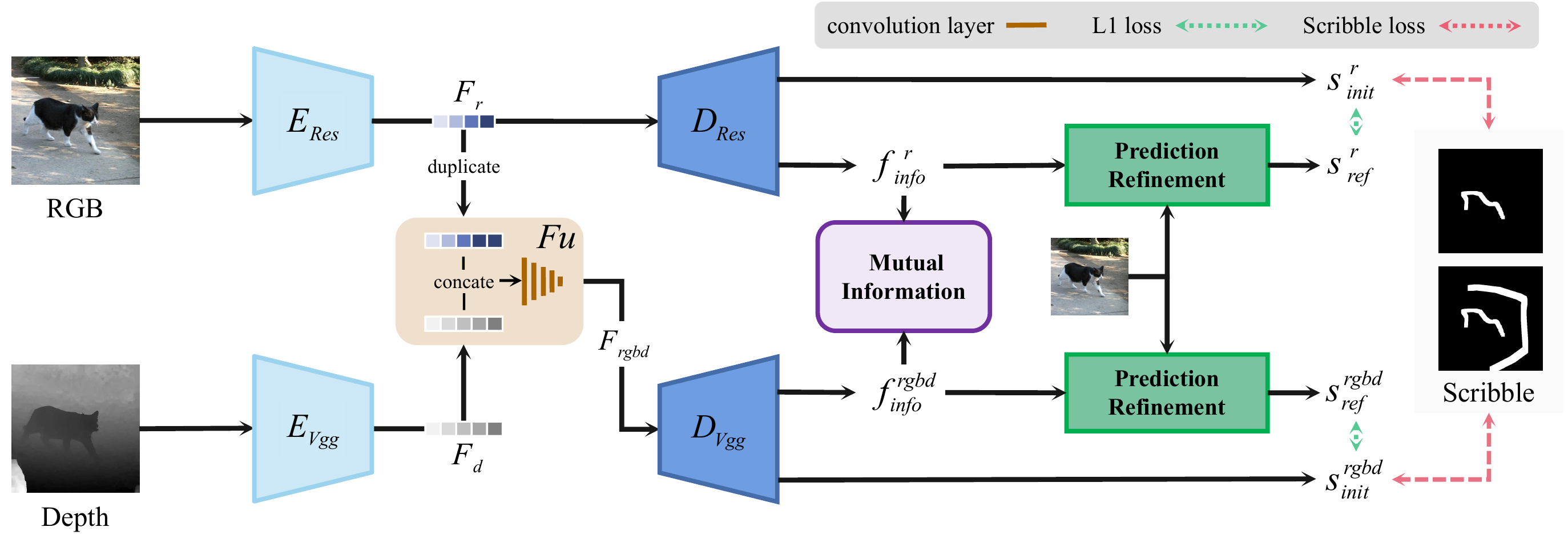}}\\
  \end{tabular}
\end{center}
%  \vspace{-5mm}
\caption{Overview of the proposed framework. We propose \textit{Asymmetric Encoders}
% ($E_{Res}$ and $E_{\mathit{Vgg}}$) 
to provide task-specific features ($F_r$ and $F_d$) for RGB and depth image. After cross-level fusion ($F_u$), we utilize \textit{Mutual Information} as a regularization term to emphasize the information specificity of the different modalities. And to optimize the dense predictions, we add \textit{Prediction Refinement} module via tree-energy loss~\cite{liang2022tree} for self-supervised learning in the training phase.}
%  \vspace{-3mm}
\label{training_overview_method}
\end{figure*}

Our method includes three cascaded steps (see Fig.~\ref{training_overview_method}): 1) initial prediction estimation; 2) progressive prediction refinement with mutual information minimization~\cite{cheng2020club,rangarajan2001mime} as a regularizer; 3) second stage prediction refinement via multimodal variational auto-encoder~\cite{vae_bayes_kumar,structure_output,multimodal_generative_models_weakly_learning}.
For the estimation process, following the conventional dense prediction models, we adopt backbone models, \ie~VGG16 or ResNet50, with a decoder to generate the initial prediction. Within the progressive refinement process, we first employ the tree-energy loss~\cite{liang2022tree}, a minimum spanning tree based online prediction refinement technique, to get the structure refined prediction. Then, to extensively explore the contribution of multimodal data for our weakly-supervised learning task, we present inter-modal mutual information minimization via mutual information regularization~\cite{cheng2020club}, achieving disentangled representation learning~\cite{chen2018isolating} to extensively explore the contribution of each modality for RGB-D saliency detection.

Different from existing RGB-D saliency detection models~\cite{xu2022weakly,SSLSOD,cascaded_rgbd_sod} that use symmetric feature extractors for both modalities, we observe the encoding abilities of different backbones differ in generating reliable feature representation of the dense segmentation task
% our model is built up existing backbones, \ie~ResNet50 backbone, VGG16 backbone. We find that different backbones are effective in providing different activation for each modality 
(see Fig.~\ref{backbone_activation_analysis}). Based on this, we introduce asymmetric feature extractors, where different backbones are used for RGB image and depth data.

 Further, unlike existing approaches that focus on enhancing the fusion of features from different modalities via attention modules~\cite{ji2021calibrated,Sun_2021_CVPR_DSA2F,hu2022multi,wang2022three,Chen_Liu_Zhang_Fu_Zhao_Du_2021},
we explore the use of mutual information estimation to explicitly explore the contribution of depth for weakly-supervised RGB-D saliency detection. Specifically, we apply a
% where mutual information minimization~\cite{cheng2020club} is presented for effective disentangled representation learning.
% help RGB-D prediction make more effective use of depth. 
% And for the first time, we provide a 
tighter upper bound~\cite{cheng2020club} of cross-modal mutual information estimation compared with~\cite{cascaded_rgbd_sod} following disentangled representation learning, achieving more reliable cross-modal complementary information exploration. 

% \sout{Further, for the first time, we provide a tighter upper bound of cross-modal mutual information estimation compared with~\cite{cascaded_rgbd_sod} following disentangled representation learning, achieving more reliable cross-modal complementary information exploration.}
 Also,
% considering the coarse predictions generated in the weakly supervised setting, 
our stochastic prediction refinement technique via multimodal variational
auto-encoder~\cite{vae_bayes_kumar,structure_output,multimodal_generative_models_weakly_learning}  is proven more effective in generating robust and refined predictions to reduce the error propagation compared with existing pseudo labeling techniques or the post-hoc refinement methods (see Table~\ref{tab:ablation_refinement}), \ie~denseCRF~\cite{dense_crf}.
% Moreover, our proposed stochastic prediction refinement model the fusion of RGB and depth on latent distribution, rather than rough blend of features, which can avoid the construction of complex fusion modules.

% \sout{Also, our stochastic prediction refinement technique via multimodal variational auto-encoder~\cite{vae_bayes_kumar,structure_output,multimodal_generative_models_weakly_learning} is proven more effective in generating robust and refined predictions compared with existing pseudo labeling techniques or the post-hoc refinement methods (see Table~\ref{tab:ablation_refinement}), \ie~denseCRF~\cite{dense_crf}.}

We summarize our main contributions as: 

\textbf{1)} We introduce a mutual information optimization method to explicitly model the contribution of RGB and depth for weakly-supervised RGB-D saliency detection; 

\textbf{2)} We present
% adopt graph-based progressive refinement strategy to encourage the contribution of unlabeled pixels for model updating; 3) our 
asymmetric feature extractors, taking advantage of different backbones' encoding abilities to achieve more reliable feature representation;

\textbf{3)} We present a multimodal variational auto-encoder framework as the second stage refinement solution to refine model prediction, which is proven more robust to error propagation issues caused by pseudo labeling. 

\section{Related Work}
\noindent\textbf{RGB-D Saliency Detection:} 
% \cite{ji2021calibrated,Sun_2021_CVPR_DSA2F,hu2022multi,wang2022three,Chen_Liu_Zhang_Fu_Zhao_Du_202}
 As a multimodal learning task, most of the existing RGB-D saliency detection models~\cite{peng2014rgbd,cheng2014depth,DANet,A2dele_cvpr2020,Fu2020JLDCF,dmra_iccv19,chen2018progressively,Li_2020_CMWNet,robust_rgbd_fusion,wu2021mobilesal,mao_generativeTransformerSOD_2021,hu2022multi,Zhang2021BTSNet,yang2022bi,gao2021unified,liu2021swinnet,jin2022moadnet,chen2022modality} focus on effective multimodal feature fusion, which can be achieved via implicit multimodal feature aggregation, or explicit modal contribution evaluation~\cite{cascaded_rgbd_sod,zhou2021specificity}. Given that depth can be noisy, one main direction exists to explore depth  contribution~\cite{zhang2020select,wang2019adaptive}, which aims to distill the more informative depth for effective multimodal learning.
For instance, \cite{ji2021calibrated} proposed to calibrate the potential noise in depth and a
% use the 
cross reference module was introduced to fuse the calibrated depth with the RGB features. \cite{Sun_2021_CVPR_DSA2F} built the depth decompose module to filter noise in RGB images using the geometric prior of the depth map.
% to help predict. 
% \cite{wang2022three,hu2022multi} designed more refined network structure to fuse depth and RGB features layer by layer. 
% \sout{As a multimodal learning task, most of the existing RGB-D saliency detection models~\cite{peng2014rgbd,cheng2014depth,DANet,A2dele_cvpr2020,Fu2020JLDCF,dmra_iccv19,chen2018progressively,Li_2020_CMWNet,robust_rgbd_fusion,wu2021mobilesal,mao_generativeTransformerSOD_2021} focus on effective multimodal feature fusion, which can be achieved via implicit multimodal feature aggregation, or explicit modal contribution evaluation~\cite{cascaded_rgbd_sod,zhou2021specificity}. Given that depth can be noisy,  one main direction exists to explore depth  contribution~\cite{zhang2020select,wang2019adaptive}, which aims to distill the more informative depth for effective multimodal learning.}
With the same goal, alternative strategies usually involve an auxiliary depth estimation module~\cite{rgbd_sod_no_depth} or refinement module~\cite{to_be_critical} via self-supervised learning to achieve better utilization of the geometric information.
Besides the deterministic
% learning pipeline of conventional 
RGB-D saliency detection models, generative model based saliency detection~\cite{jing2020uc} is also explored to explain the \enquote{subjective nature} of saliency.

\noindent\textbf{Weakly-supervised Saliency Detection:}
The primary weak annotations for saliency detection include image or class level label~\cite{piao2021mfnet,li2018weakly,hsu2019weakly,to_be_critical,noise_sensitive_adv,self_supervised_saliency,instance_saliency_weak}, saliency subitizing~\cite{weak_saliency_subitizing}, scribble~\cite{jing2020weakly,xu2022weakly,yu2021structure}, bounding box~\cite{liu2021weakly}, and noisy labels from handcrafted-feature based methods~\cite{Wang_2022_CVPR,zhang2018deep,supervision_by_fusion,zhang2020learning,deepups,zhang2020learning_tpami,ji2022promoting,Feng_2022_CVPR_lightfield_noise}. Given the close relation between fixation prediction
% (to localize the most visual attractive regions) 
and salient object detection, \cite{zhao2021weakly} treats fixation maps as weak annotations for video salient object detection.
% To extensively relax the labeling process,
Point supervision is used in \cite{weak_video_point}
% introduces  based 
for
video saliency detection. Based on class-activation map~\cite{zhou2016learning}, Image-level supervision~\cite{piao2021mfnet,li2018weakly,hsu2019weakly,instance_saliency_weak} provide coarse foreground localization, highlighting the discriminative region of the object(s). Differently, scribble supervision~\cite{jing2020weakly,xu2022weakly,yu2021structure} is usually accurate and sparse. Although bounding-box~\cite{liu2021weakly} usually cover the entire extend of the salient foreground, the extra background region leads to extensive false positive, especially for objects with concave shapes. Alternatively, noise labeling based solutions~\cite{supervision_by_fusion,zhang2020learning_tpami,ji2022promoting,Feng_2022_CVPR_lightfield_noise} are designed to identify clean supervisions from labels generated with the conventional techniques. 

\noindent\textbf{Graph-based Saliency Detection:}
Graph-based~\cite{normalized_cut,graph_theoretical_clustering,an_optimal_graph,graph_theoretical_methods} segmentation techniques usually represent an image as an undirected graph $G=(\mathcal{V},\mathcal{E})$, where vertices   $v_i\in\mathcal{V}$ corresponds to pixels, and the edges $(v_i,v_j)\in\mathcal{E}$ connect the neighborhood pixels. A weight $w(v_i,v_j)$ is associated with each edge that connects a pair of vertices, which is a non-negative measure of the dissimilarity between $v_i$ and $v_j$. Conventional graph-based segmentation methods \cite{normalized_cut} define the weight based on the intensity difference of the vertices pairs to achieve
% achieve a segmentation $S$ of the image, representing 
a partition of $\mathcal{V}$ with
% into 
different components. Given that graph-based methods can explore pixel-wise correlation, many conventional saliency detection models~\cite{yang2013saliency,aytekin2017learning,lu2014learning,gopalakrishnan2010random,real_time_saliency_mst,label_propagation_graph,two_stage_graph_saliency} construct pixel or superpixel level graph for an image to achieve salient region segmentation, where the weight is defined based on low-level image feature, \eg~color, intensity, location and \etcn Graph can also be constructed based on deep features~\cite{wang2021graph,xiao2019saliency,liang2022tree,adaptive_irregular_graph_saliency} to
% . Graph-based methods can also 
explore image-wise correlation, leading to graph-based co-saliency detection models~\cite{Hsu_2018_ECCV,co_saliency_graph_matching,Zhang_2020_CVPR_cosaliency_graph,gfn_aaai21,graph_cosaliency_instance_coseg}, video saliency detection~\cite{video_saliency_graph}.

\noindent\textbf{Mutlimodal Mutual Information Estimation:} Mutual Information (MI) captures the dependencies~\cite{Equitability_mutual_information} between variables. For a pair of random variables $X$ and $Y$, their MI $I(X;Y)$ is defined as the KL-Divergence of the joint distribution $p(X,Y)$ from the product of the marginal distributions $p(X)$ and $p(Y)$. Due to its explicit dependency modeling ability, MI is typically used as a regularizer to encourage or limit the dependency between variables via MI maximization with a lower bound~\cite{poole2019variational,ba_mmm_nips03} or MI minimization with an upper bound~\cite{cheng2020club}. 
The MI maximization is usually applied for effective self-supervised representation learning~\cite{sanchez2020learning,colombo2021novel,xin2021disentangled,ma2020decoupling,gao2021information,soleymani2021mutual,Disentangled_Representation_Learning_with_Wasserstein_Total_Correlation}, assuming each modality contains full information of the target distribution. On the contrary, the MI minimization~\cite{Private_Shared_Disentangled_Multimodal_VAE,daunhawer2020self,cheng2020club,rangarajan2001mime} acknowledge the partial information about the desired task within each modality, and it serves as a regularizer to achieve disentangled representation learning~\cite{chen2018isolating}. Mutual information can also be estimated via Bayesian analysis~\cite{wolpert1995estimating,hutter2001distribution,archer2013bayesian}, which involves a prior probability and a likelihood to compute a posterior probability. Specifically, let's define the mutual information of random variable $x$ and $y$ as $I(x,y)$, which can also be explained as the Kullback–Leibler divergence between the joint distribution $p(x,y)$ and the outer product of the marginal distribution $p(x)$ and $p(y)$. Given samples $D$ from the joint distribution, with Bayesian analysis, we have a prior $p(I(x,y))$, and a likelihood $p(D|I(x,y))$, the posterior expectation of the mutual information can be estimated via Bayesian analysis, achieving mutual information estimation.

\noindent\textbf{Prediction Refinement for Weakly-supervised Learning:}
Considering that weak supervisions are used instead of the full annotations, prediction refinement is widely adopted in weakly-supervised models either
% models have little annotation, refinement are used 
during training~\cite{shin2021all,li2022towards}, or as post-processing~\cite{lee2021anti,li2021pseudo,li2022towards,lee2021reducing} technique to obtain more accurate predictions.
% information. 
Among those methods,
% The refinement module is introduced in the training phase usually by adding 
new supervisions are usually added manually \cite{shin2021all,xu2022weakly} to the model as pseudo labels.
% oror by using low-dimensional features \cite{li2022towards} to help add detail to prediction maps.
% Based on active learning, 
\cite{shin2021all} introduced
% are continuously added 
the newly annotated pixels to optimize model predictions during training via uncertainty-aware learning.
% phase by the uncertainty of the pseudo label.
% And in order to generate a more fine-grained pseudo label for the retrain phase,
\cite{li2021pseudo} used class activation map (CAM)~\cite{zhou2016learning} as seed to generate initial predictions, which   are then refined with denseCRF~\cite{dense_crf}.
% , which is then and smooth the CAM by using coefficient of variation smoothing and introduces CRF to optimize the details of the pseudo label. 
Similarly, \cite{lee2021anti,lee2021reducing} generated coarse CAM and used seed refinement methods~\cite{ahn2019weakly,ahn2018learning} to produce better pseudo labels. \cite{li2022towards} utilized denseCRF~\cite{dense_crf} to refine the prediction, where denseCRF~\cite{dense_crf} served as both training-phrase refinement technique and post-processing method.
\section{Our Method}

% \section{Generative RGB-D Latent Representation}
Our weakly-supervised training dataset is $D=\{X_i,y_i\}_{i=1}^N$,
% as our training dataset, 
where $i$ indexes the samples and we omit it when it's clear, $X=\{x^r,x^d\}$ is the input RGB ($x^r$) and depth ($x^d$) pair, $y_{s}$ is the
% group truth saliency map, which is 
scribble annotation.
% in our case. 
% We introduce an explicit multimodal learning framework for effective weakly-supervised RGB-D saliency detection. Specifically, w
We first design asymmetric feature extractors with a simple encoder to extract features for the input data pair and generate our initial prediction, where the asymmetric feature extractors are proven more effective than the conventional symmetric feature extractors. Then, the initial prediction is further refined
% A simple cross-level feature fusion strategy is further introduced to fuse the multimodal features and generate our initial prediction, which is further refine 
during training with the minimum spanning tree based tree-energy loss  and  attention based fusion module (Sec.~\ref{sec:feat_extractors}). To achieve disentanglement representation learning for our multimodal task,
% multimodal feature representation, 
we perform mutual information minimization
% inspired by~\cite{cascaded_rgbd_sod} 
with a tighter mutual information upper bound (Sec.~\ref{sec:mi_min_reg}). Finally, we present a multimodal variational auto-encoder
% generative model
based prediction refinement strategy
% via multimodal variational auto-encoder~
\cite{vae_bayes_kumar,structure_output,multimodal_generative_models_weakly_learning} (Sec.~\ref{sec:multimodal_vae}), which is proven relatively robust to labeling noise compared with other deterministic prediction refinement techniques~\cite{dense_crf} and self-training~\cite{xu2022weakly}.

\begin{figure}[!htp]
\begin{center}
  \begin{tabular}{c@{ }c@{ }c@{ }c@{ }c@{ } }
  {\includegraphics[width=0.18\linewidth]{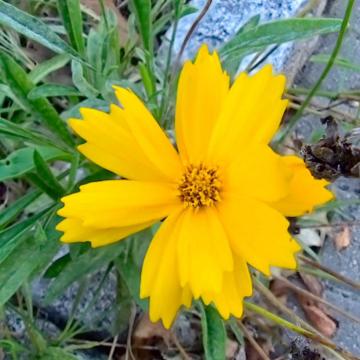}}&
  {\includegraphics[width=0.18\linewidth]{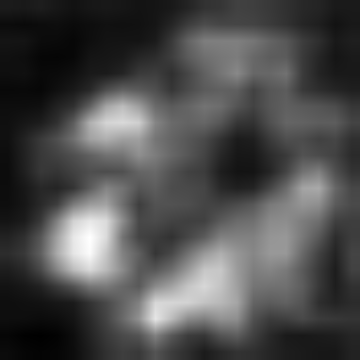}}&
  {\includegraphics[width=0.18\linewidth]{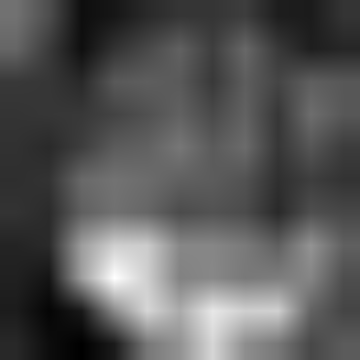}}&
  {\includegraphics[width=0.18\linewidth]{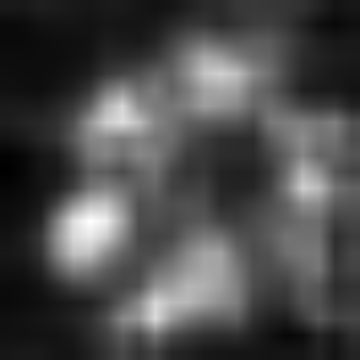}}&
  {\includegraphics[width=0.18\linewidth]{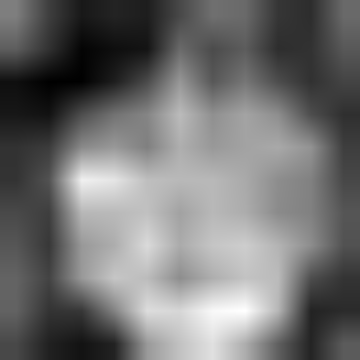}}\\
  {\includegraphics[width=0.18\linewidth]{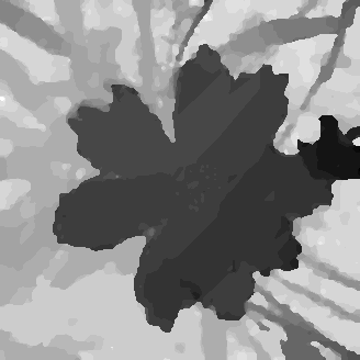}}&
  {\includegraphics[width=0.18\linewidth]{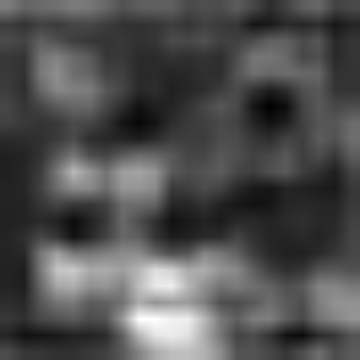}}&
  {\includegraphics[width=0.18\linewidth]{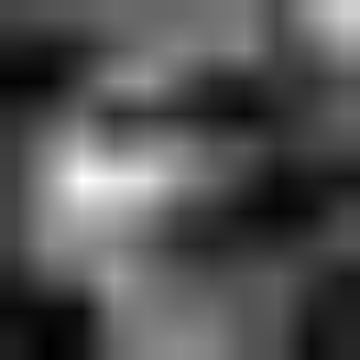}}&
  {\includegraphics[width=0.18\linewidth]{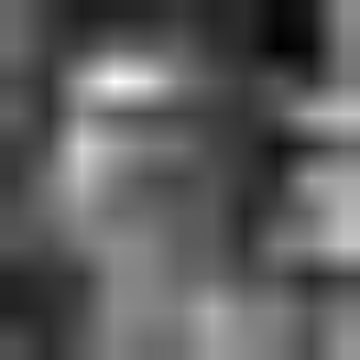}}&
  {\includegraphics[width=0.18\linewidth]{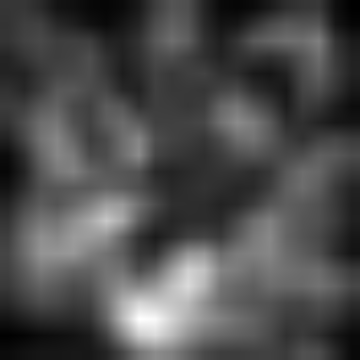}}\\
   \footnotesize{Img/Depth} & \footnotesize{VGG-VGG} & \footnotesize{Res-Res} & \footnotesize{VGG-Res} & \footnotesize{Res-VGG} 
  \end{tabular}
  \end{center}
%   \vspace{-5mm}
\caption{Feature activation with different backbone settings. From left to right: RGB, Depth pair, activation of RGB (top), depth (bottom) within VGG-VGG, ResNet-ResNet, VGG-ResNet, and ResNet-VGG settings, where left means backbone for RGB, and the right one indicates backbone for depth.}
%  \vspace{-3mm}
\label{backbone_activation_analysis}
\end{figure}
% We present a conditional multimodal variational auto-encoder (VAE)~\cite{vae_bayes_kumar,structure_output,multimodal_generative_models_weakly_learning} with explicit unimodal optimization for RGB-D latent representation learning.

\subsection{Initial Predictions via Asymmetric Encoders}
\label{sec:feat_extractors}
% \Jing{finish this part according to your current implementation. introduce backbone for each modality}
\noindent\textbf{Asymmetric Encoders:} We discovered that the RGB image and depth data respond  differently to the same backbone, where the ResNet50 backbone works better in generating more reliable foreground activation for RGB image, and the VGG16 backbone provides better activation for the depth data. In Fig.~\ref{backbone_activation_analysis}, we visualize the backbone feature (the deep level backbone feature) by performing channel-wise sum and min-max normalization.
% and from the visualization activation maps w
We observe that
% are able to observe that 
under our RGB-D SOD setting, the VGG backbone features focus more on details while ResNet backbone features highlight more
% pays more attention to the
% has larger values at the edges compared to ResNet, which is more focused on the 
target region(s). With the asymmetric encoders, 
% \enquote{Res-VGG} setting as backbones for RGB/depth, 
semantic meaningful activation can be extracted from the RGB branch, and at the same time, we obtain
% on the other hand, the depth branch can provide 
structure accurate activation from the depth branch.
% the phenomenon that intuitively fits well with our proposed asymmetric encoders. 
% We tend to let depth provide effective structural information, while RGB provides more semantic information.
% \sout{which shows that with the \enquote{Res-VGG} setting as backbones for RGB/depth, semantic meaningful activation can be extracted from the RGB branch, and on the other hand, the depth branch can provide structure accurate activation.}
% \sout{ We show more activation maps in the supplementary materials.}
% modalities 
% exhibit distinct responses at different backbones in the SOD (see), so 
In this paper, we utilized the pre-trained ResNet50 backbone to extract the RGB image features and the pre-trained VGG network to extract the depth features, where the former is defined as $F_r= E_{res}(x^r)$ and the latter is defined as $F_{d}= E_{\mathit{vgg}}(x^{d})$, where $E_{res}(\cdot)$ and $E_{\mathit{vgg}}(\cdot)$ indicate the ResNet50 and VGG16 backbone based feature extractors, respectively.
% \begin{equation}
%     \label{init_decoder}
%     \begin{aligned}
%       F_r &= E_{res}(x^r), \\ %{f^r_1,f^r_2,f^r_3,f^r_4}
%         F_{d} &= E_{vgg}(x^{d}), %{f^r_1,f^r_2,f^r_3,f^r_4}
%     \end{aligned}
% \end{equation}    
% where $ F_r = \{f^r_1,f^r_2,f^r_3,f^r_4\} $. $E_{vgg}$ means the VGG backbone encoder  and $E_{res}$ the Resnet backbone encoder.

\begin{figure}[h]
\begin{center}
  \begin{tabular}{c@{ } }
  {\includegraphics[width=0.7\linewidth]{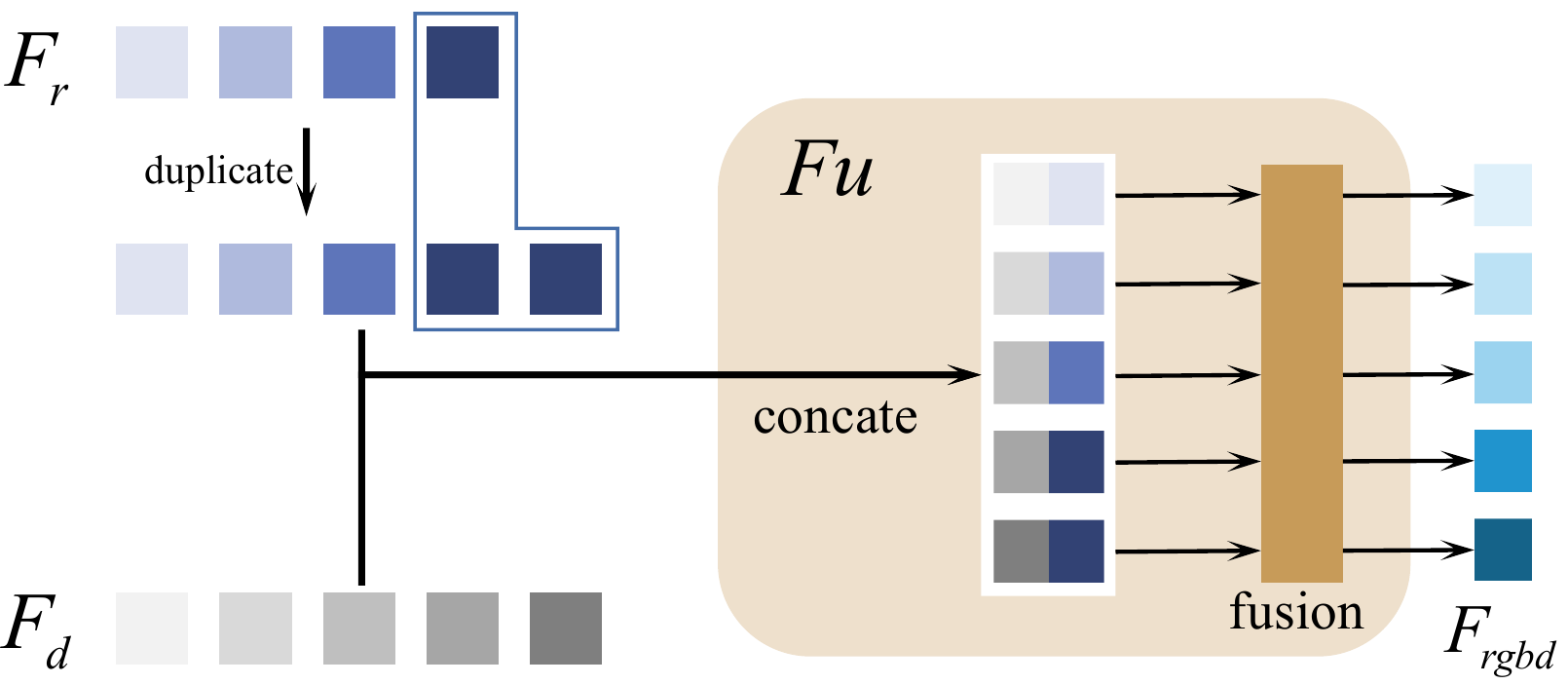}}
   % \footnotesize{Img/Depth} & \footnotesize{VGG-VGG} & \footnotesize{Res-Res} & \footnotesize{VGG-Res} & \footnotesize{Res-VGG} 
  \end{tabular}
  \end{center}
%   \vspace{-5mm}
\caption{Illustration of the feature fusion module $Fu$ for Asymmetric Encoder features. The RGB backbone feature $F_r$ is duplicated to the same channel size with depth feature $F_d$. $Fu$ consists of channel-wise concatenation and one $3 \times 3$ convolutional layers for each backbone feature stage. }
%  \vspace{-3mm}
\label{backbone_fusion}
\end{figure}

\begin{figure*}[!t]
\begin{center}
  \begin{tabular}{c@{ }c@{ } c@{ } c@{ } c@{ } c@{ } c@{ }c@{ } c@{ } c@{ } c@{ } }

  {\includegraphics[width=0.085\linewidth,height=0.066\linewidth]{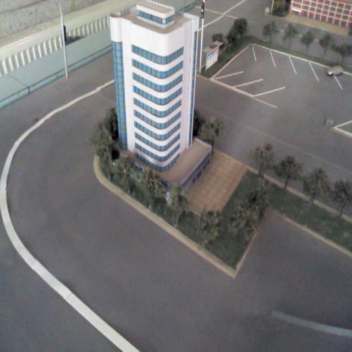}}&
  {\includegraphics[width=0.085\linewidth,height=0.066\linewidth]{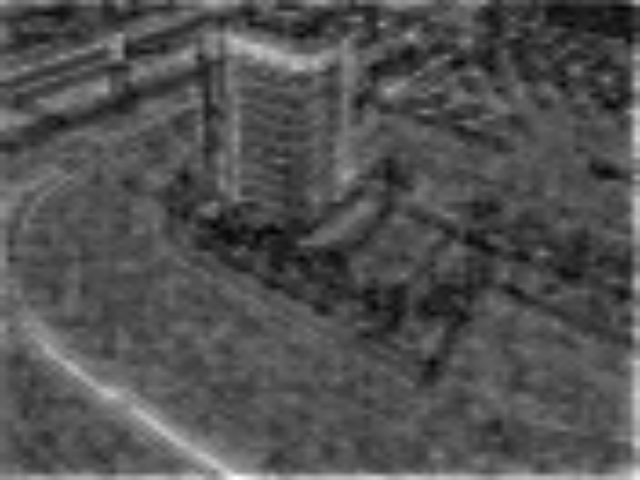}}&
  {\includegraphics[width=0.085\linewidth,height=0.066\linewidth]{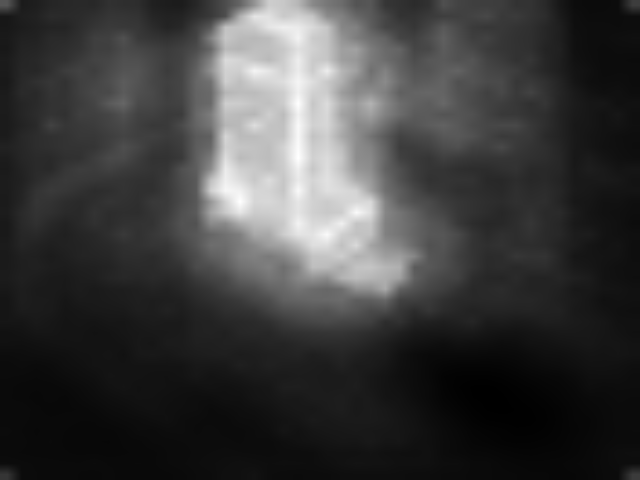}}&
  {\includegraphics[width=0.085\linewidth,height=0.066\linewidth]{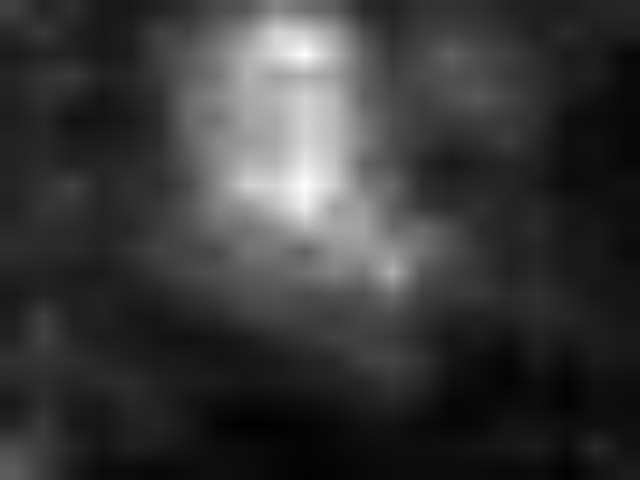}}&
  {\includegraphics[width=0.085\linewidth,height=0.066\linewidth]{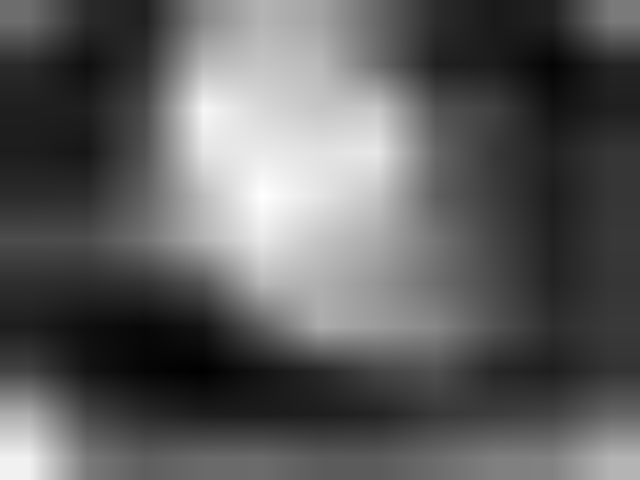}}&
  {\includegraphics[width=0.085\linewidth,height=0.066\linewidth]{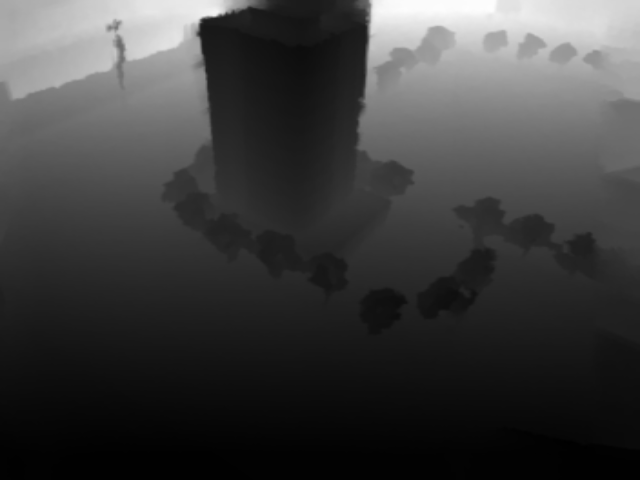}}&
  {\includegraphics[width=0.085\linewidth,height=0.066\linewidth]{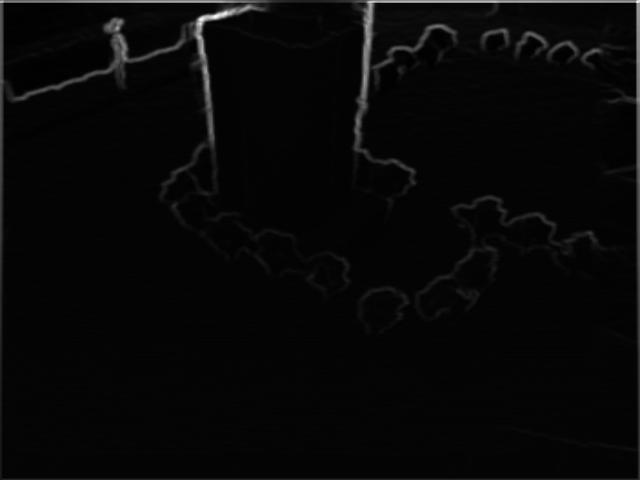}}&
  {\includegraphics[width=0.085\linewidth,height=0.066\linewidth]{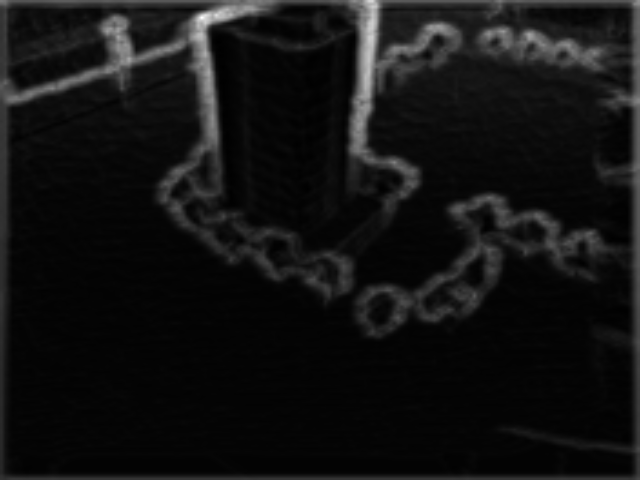}}&
  {\includegraphics[width=0.085\linewidth,height=0.066\linewidth]{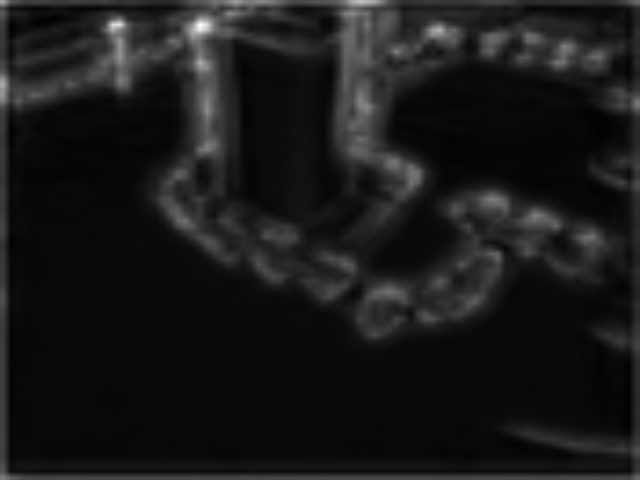}}&
  {\includegraphics[width=0.085\linewidth,height=0.066\linewidth]{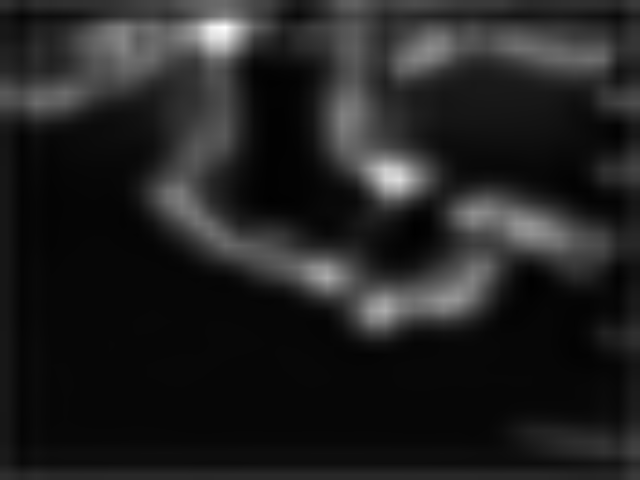}}&
  {\includegraphics[width=0.085\linewidth,height=0.066\linewidth]{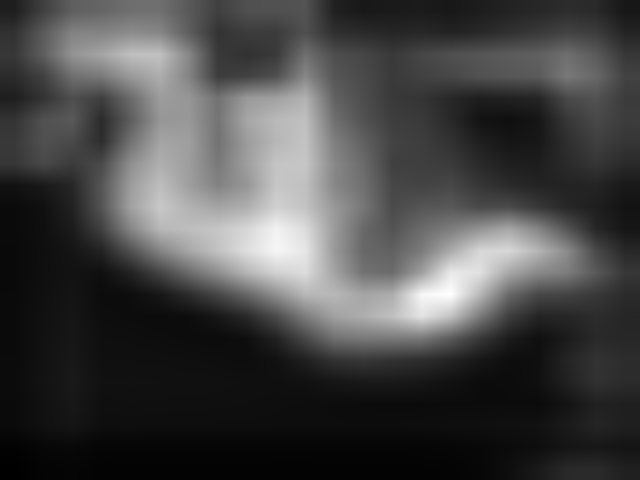}} \\

 {\includegraphics[width=0.085\linewidth,height=0.066\linewidth]{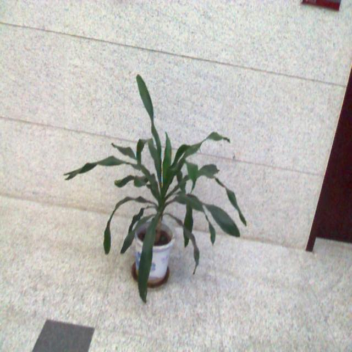}}&
  {\includegraphics[width=0.085\linewidth,height=0.066\linewidth]{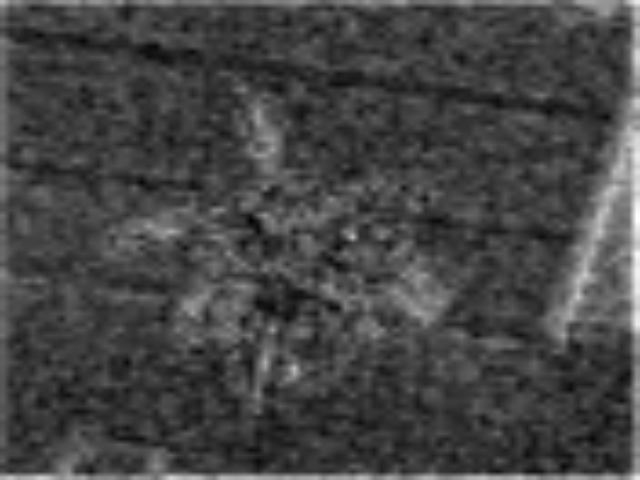}}&
  {\includegraphics[width=0.085\linewidth,height=0.066\linewidth]{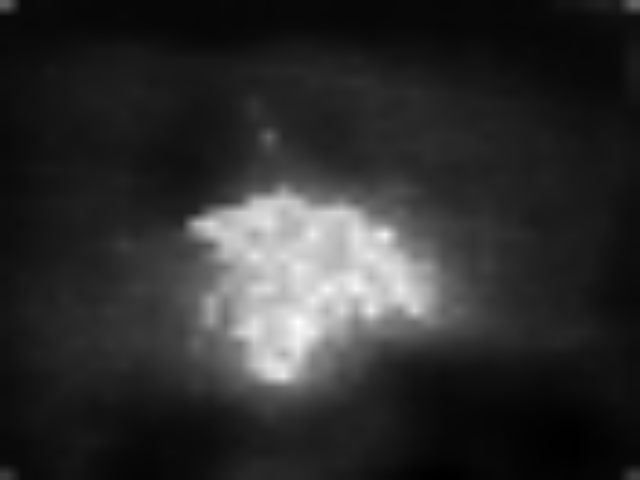}}&
  {\includegraphics[width=0.085\linewidth,height=0.066\linewidth]{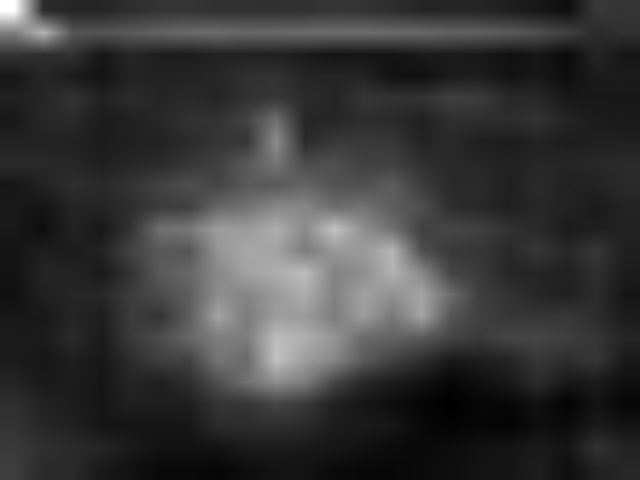}}&
  {\includegraphics[width=0.085\linewidth,height=0.066\linewidth]{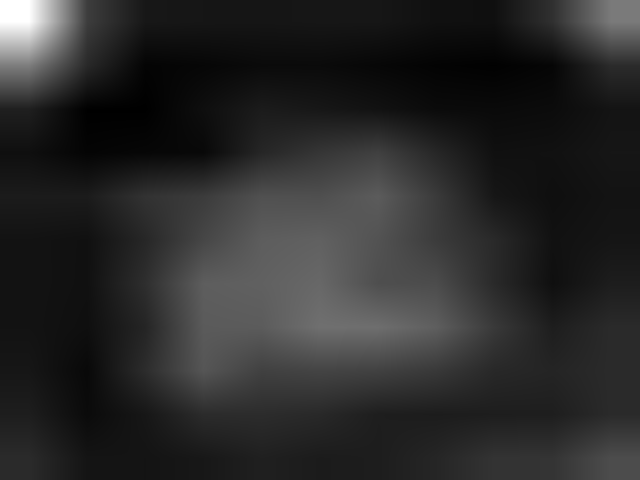}}&
  {\includegraphics[width=0.085\linewidth,height=0.066\linewidth]{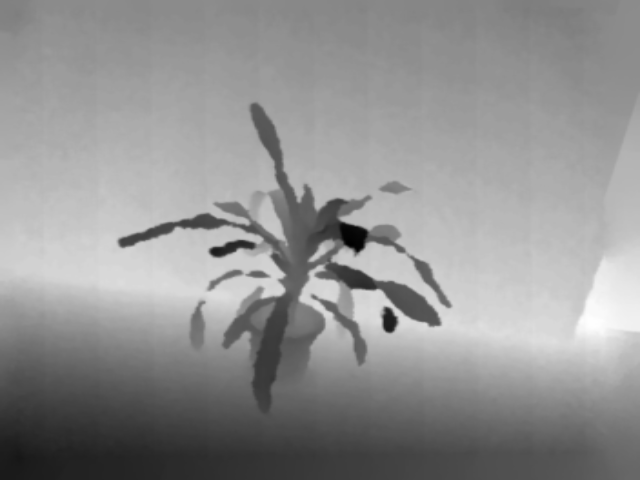}}&
  {\includegraphics[width=0.085\linewidth,height=0.066\linewidth]{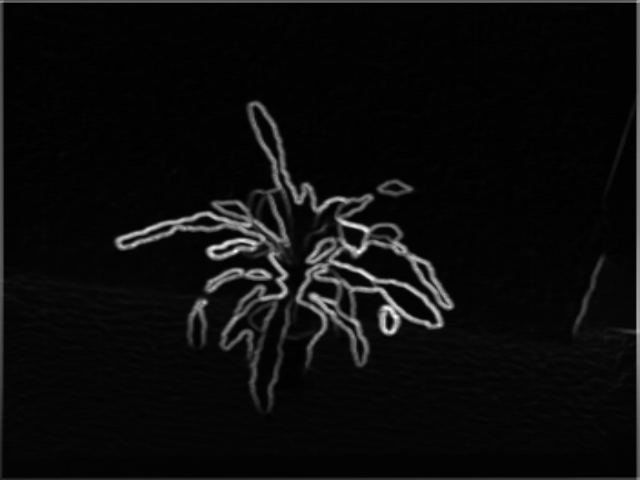}}&
  {\includegraphics[width=0.085\linewidth,height=0.066\linewidth]{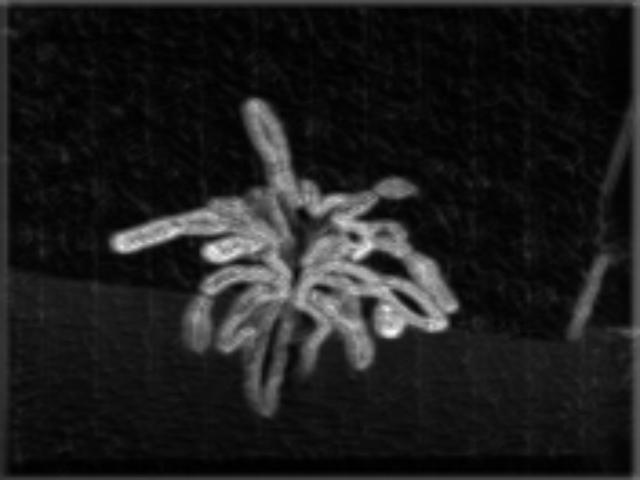}}&
  {\includegraphics[width=0.085\linewidth,height=0.066\linewidth]{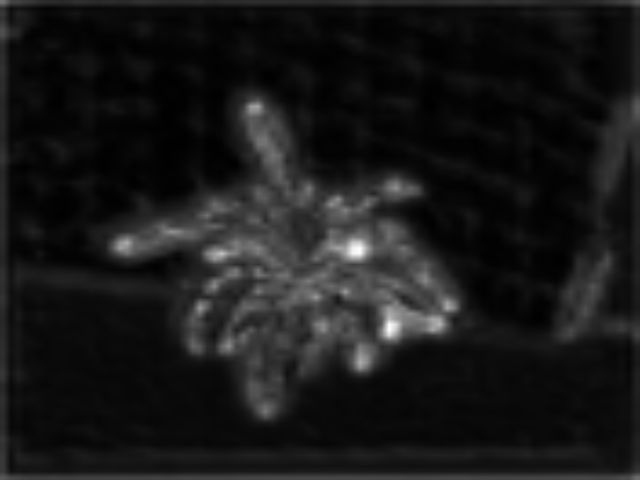}}&
  {\includegraphics[width=0.085\linewidth,height=0.066\linewidth]{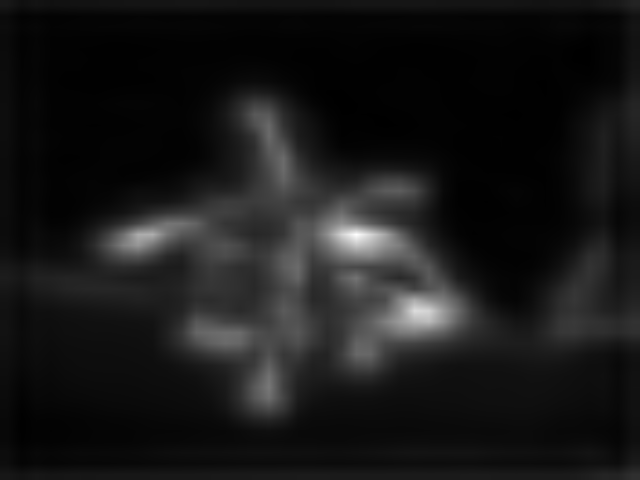}}&
  {\includegraphics[width=0.085\linewidth,height=0.066\linewidth]{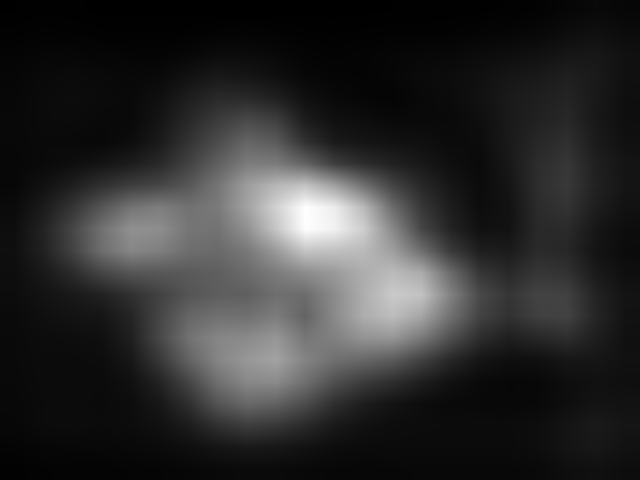}} \\
  
 {\includegraphics[width=0.085\linewidth,height=0.066\linewidth]{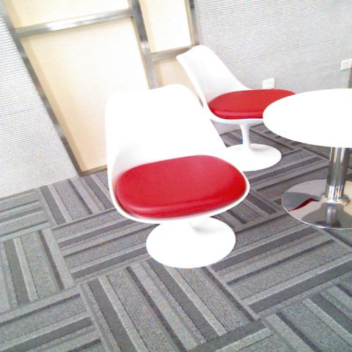}}&
  {\includegraphics[width=0.085\linewidth,height=0.066\linewidth]{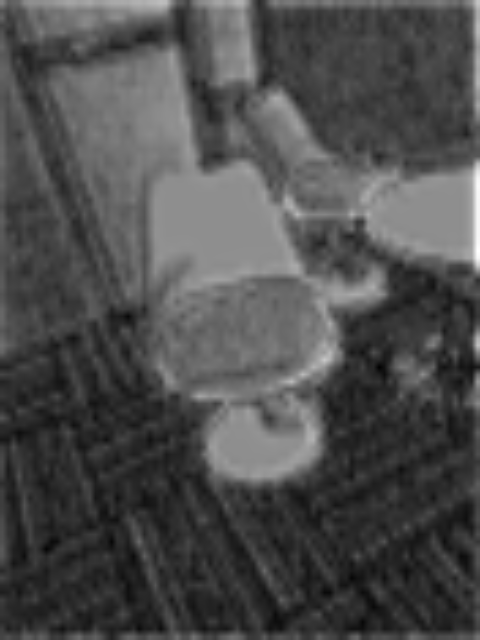}}&
  {\includegraphics[width=0.085\linewidth,height=0.066\linewidth]{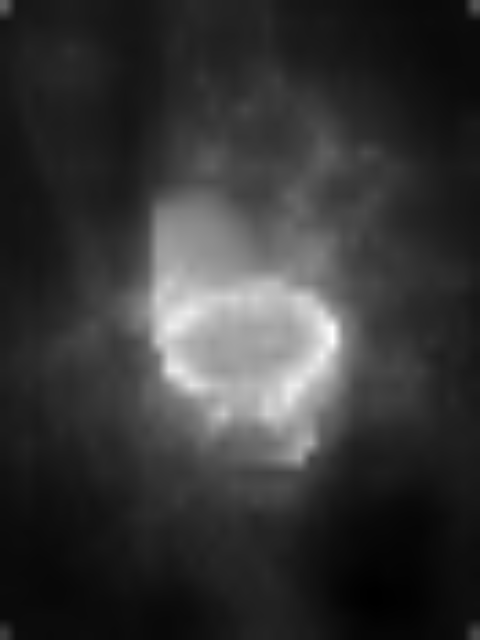}}&
  {\includegraphics[width=0.085\linewidth,height=0.066\linewidth]{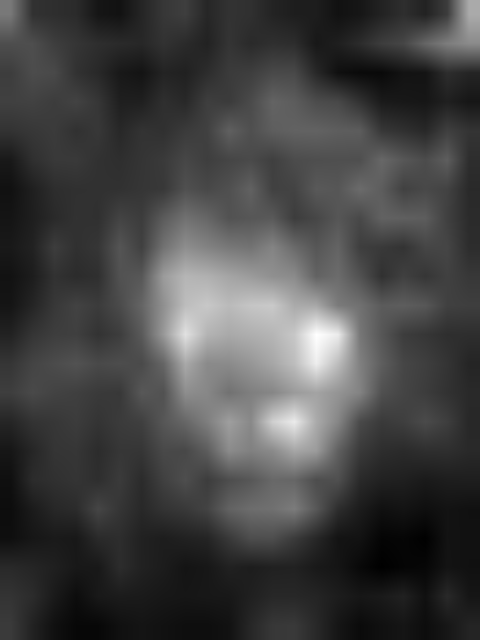}}&
  {\includegraphics[width=0.085\linewidth,height=0.066\linewidth]{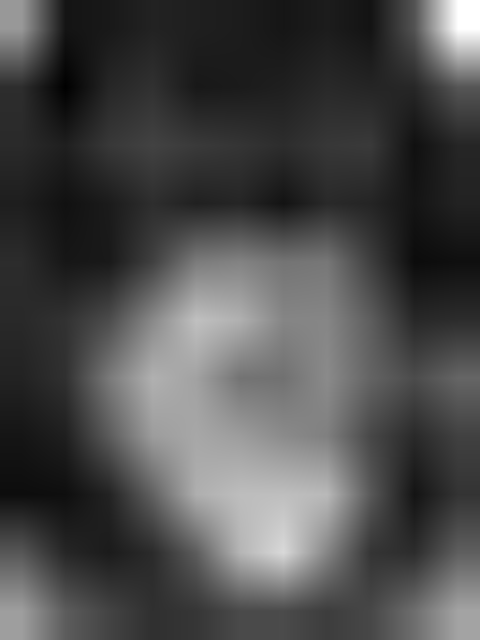}}&
  {\includegraphics[width=0.085\linewidth,height=0.066\linewidth]{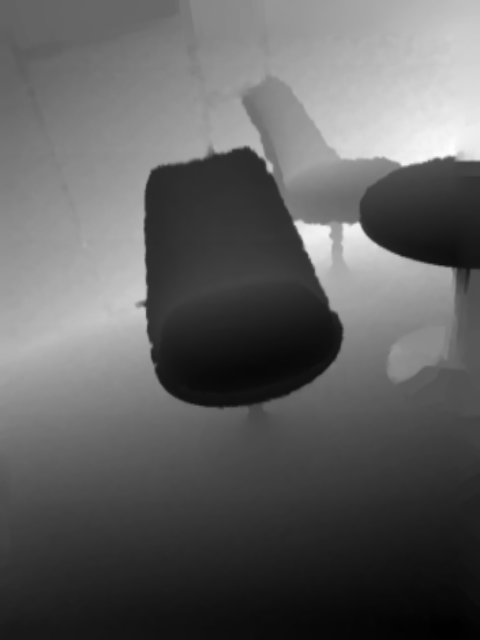}}&
  {\includegraphics[width=0.085\linewidth,height=0.066\linewidth]{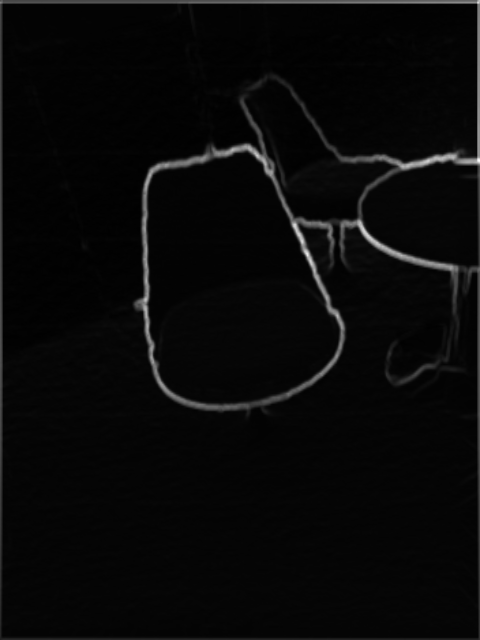}}&
  {\includegraphics[width=0.085\linewidth,height=0.066\linewidth]{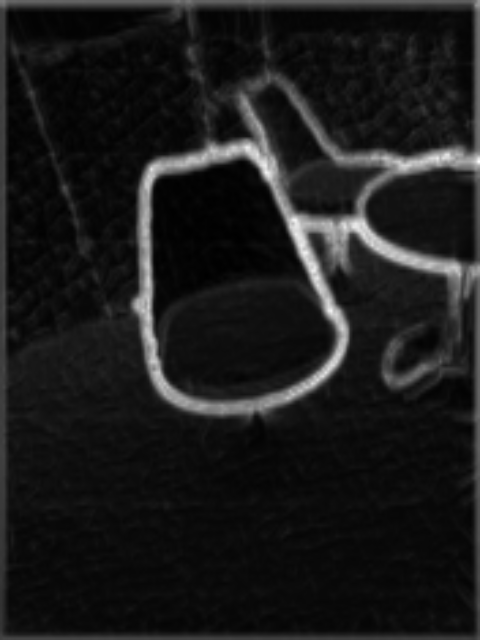}}&
  {\includegraphics[width=0.085\linewidth,height=0.066\linewidth]{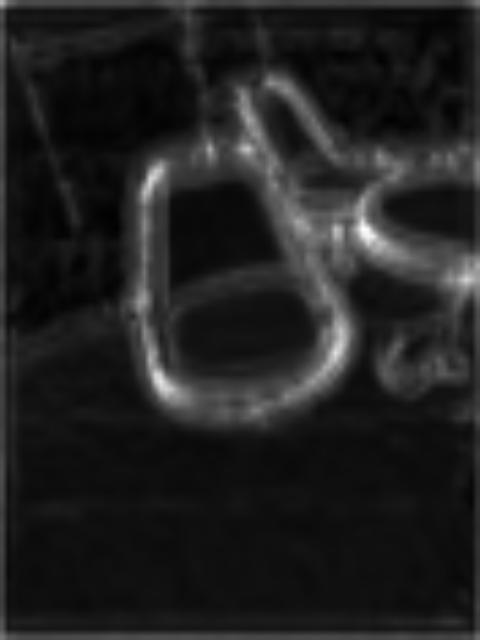}}&
  {\includegraphics[width=0.085\linewidth,height=0.066\linewidth]{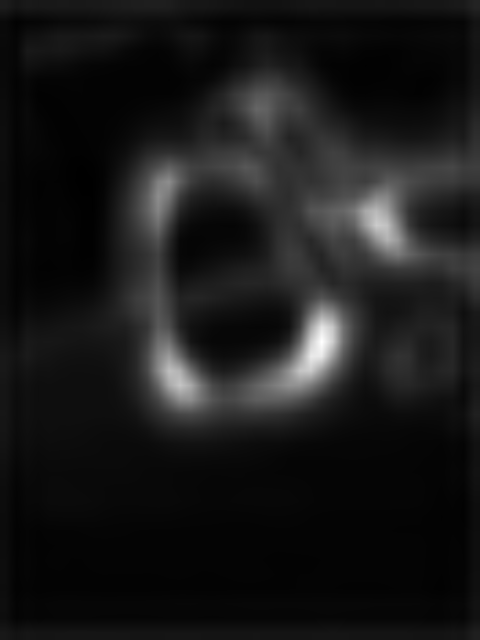}}&
  {\includegraphics[width=0.085\linewidth,height=0.066\linewidth]{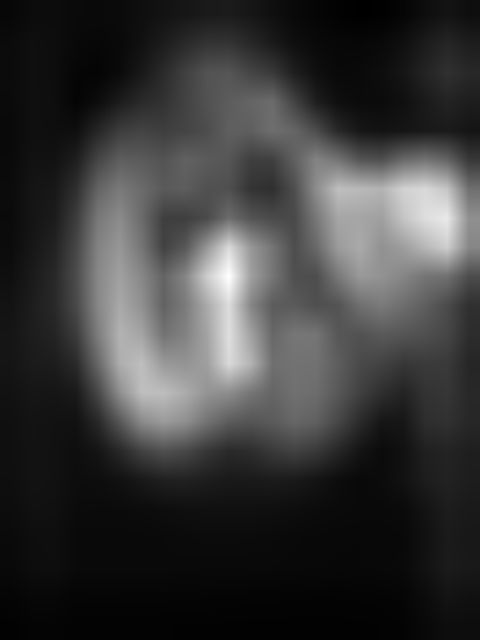}}\\

 {\includegraphics[width=0.085\linewidth,height=0.066\linewidth]{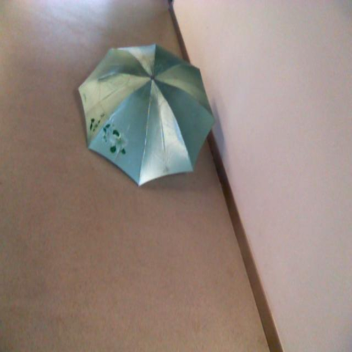}}&
  {\includegraphics[width=0.085\linewidth,height=0.066\linewidth]{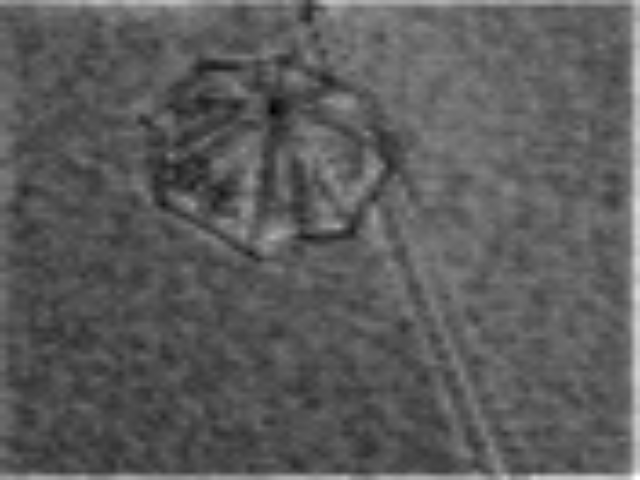}}&
  {\includegraphics[width=0.085\linewidth,height=0.066\linewidth]{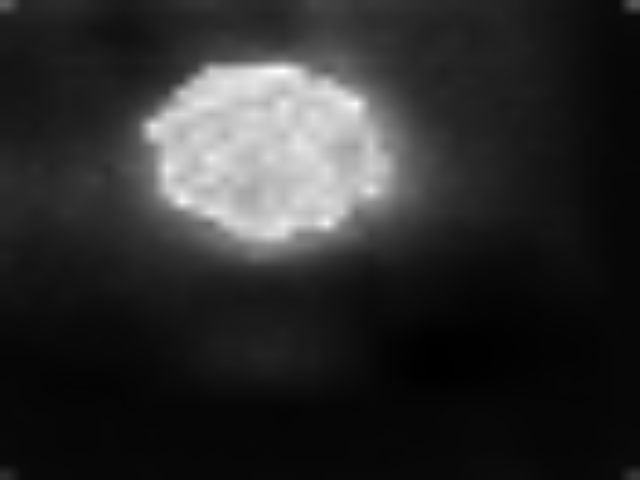}}&
  {\includegraphics[width=0.085\linewidth,height=0.066\linewidth]{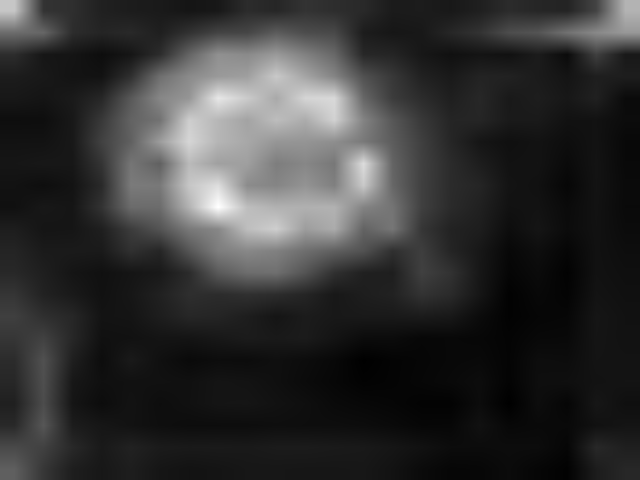}}&
  {\includegraphics[width=0.085\linewidth,height=0.066\linewidth]{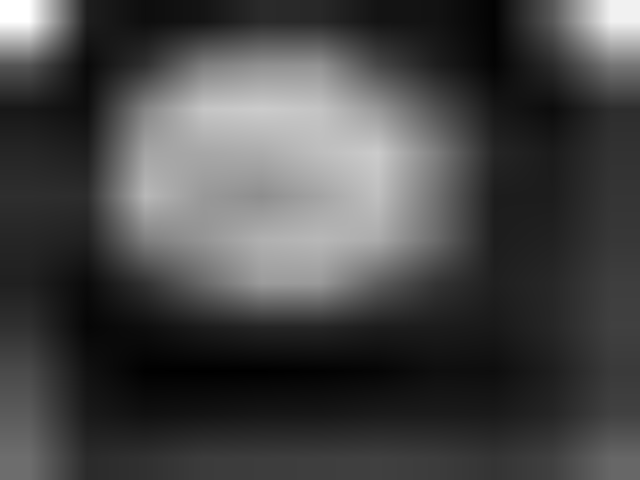}}&
  {\includegraphics[width=0.085\linewidth,height=0.066\linewidth]{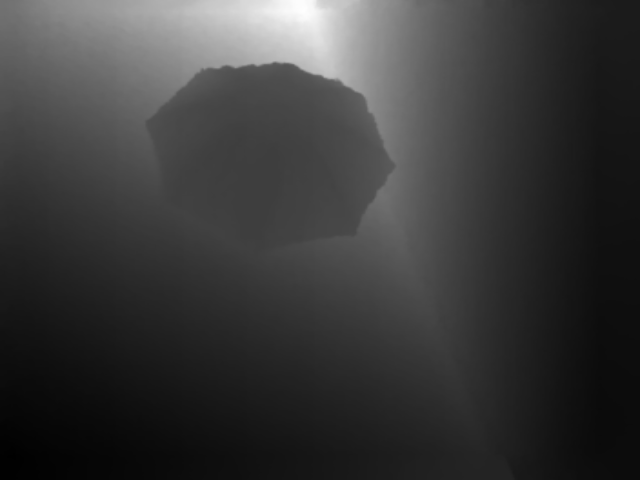}}&
  {\includegraphics[width=0.085\linewidth,height=0.066\linewidth]{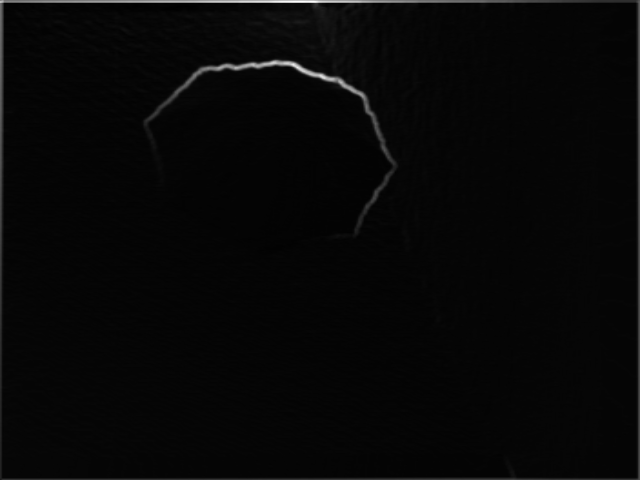}}&
  {\includegraphics[width=0.085\linewidth,height=0.066\linewidth]{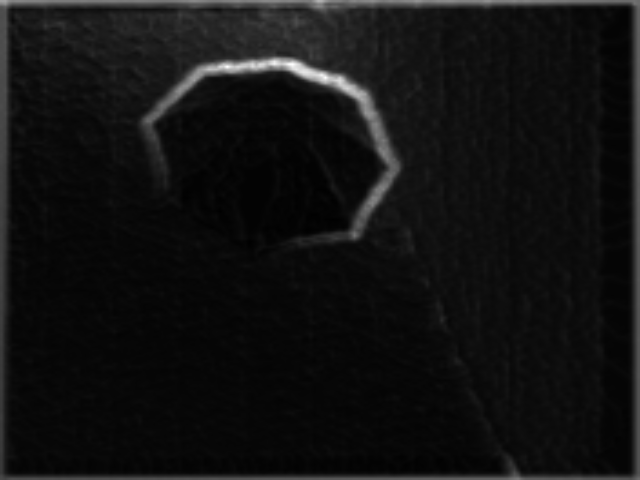}}&
  {\includegraphics[width=0.085\linewidth,height=0.066\linewidth]{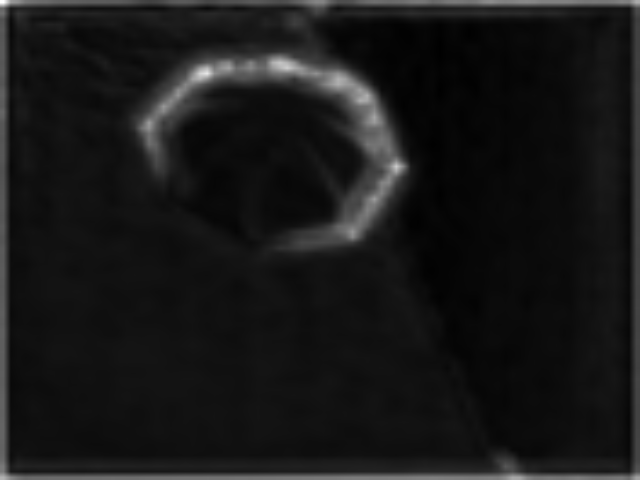}}&
  {\includegraphics[width=0.085\linewidth,height=0.066\linewidth]{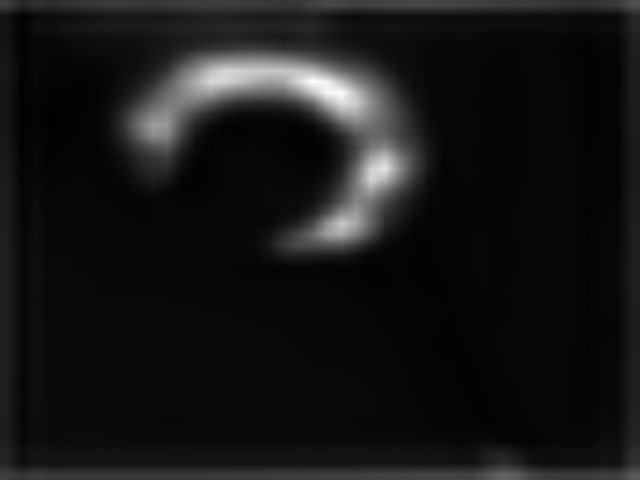}}&
  {\includegraphics[width=0.085\linewidth,height=0.066\linewidth]{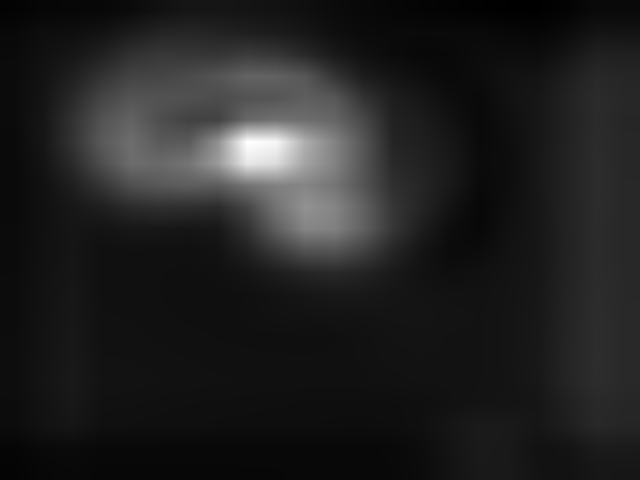}} \\

\footnotesize{Image}&\footnotesize{\textbf{$f_r^1$}}&\footnotesize{\textbf{$f_r^2$}}&\footnotesize{\textbf{$f_r^3$}}&\footnotesize{\textbf{$f_r^4$}} & \footnotesize{Depth}&\footnotesize{\textbf{$f_d^1$}}&\footnotesize{\textbf{$f_d^2$}}&\footnotesize{\textbf{$f_d^3$}}&\footnotesize{\textbf{$f_d^4$}}&\footnotesize{\textbf{$f_d^5$}}
  \end{tabular}
  \end{center}
   % \vspace{-5mm}
\caption{Visualization of our proposed asymmetric encoder produces the RGB and depth activation maps. $f_r^i, i=1,...,4$ is the RGB (ResNet) backbone feature, and $f_d^i, i=1,...,5$ means the depth (VGG) backbone feature. } 
\label{activation_map_visualization_of_encoder}
\end{figure*}

\noindent\textbf{Initial prediction generation:} For RGB and depth branch,  we use a simple decoder to generate initial predictions with coarse features via $\{s^{r}_{init},f^r_{\mathit{info}}\} = D_{res}(F_{r})$, where $D_{res}$ represents decoder for the ResNet50 backbone, consisting of four $3\times3$ convolutional layers with four sets of feature fusion modules from~\cite{Ranftl2022_midas}. $s^{r}_{init},f^r_{\mathit{info}}$ are the prediction and the corresponding coarse feature. 
% \Rev{To achieve multimodal fusion, we fuse the backbone feature of RGB image $F_{r}$ and depth feature $F_{d}$ with a channel-wise concatenation operation by a fusion module $\mathit{Fu}(\cdot)$, which consist of channel-wise concatenation operation and one $3\times 3$ convolutional layers for each stage of the backbone feature as shown in Fig.~\ref{backbone_fusion}. Note that VGG16 has five stages of backbone features $F_{d} = \{f_d^1,f_d^2,f_d^3,f_d^4,f_d^5\}$, while ResNet50 has only four $F_{r} = \{f_r^1,f_r^2,f_r^3,f_r^4\}$. Thus we duplicate the highest level feature from the RGB encoder like  $F_{r}' = \{f_r^1,f_r^2,f_r^3,f_r^4,f_r^4\}$, which has similar characteristics expressed with depth feature like Fig.~\ref{activation_map_visualization_of_encoder}.And generate the fused RGB-D data based backbone feature $F_{\mathit{rgbd}}=\mathit{Fu}(F_{r}', F_{d})$ with five stages of features, which is precisely the same size as the depth backbone features.}
 % \sout{
 To achieve multimodal fusion, we simply concatenate the backbone feature of RGB image and depth data, and feed each stage of the backbone feature to a fusion module to achieve the fused RGB-D data based backbone feature $F_{\mathit{rgbd}}=\mathit{Fu}(F_{r},F_{d})$, where $\mathit{Fu}(\cdot)$ represents a fusion module consisting of channel-wise concatenation operation and one $3\times 3$ convolutional layers for each stage of the backbone feature as shown in Fig.~\ref{backbone_fusion}. Note that VGG16 has five stages of backbone features, while ResNet50 has only four. Thus we duplicate the highest level feature from the RGB encoder, leading to five stages of features for $F_{\mathit{rgbd}}$, which is precisely the same size and similar degree of semantic information as
% the RGB-D feature branch, as the same size as 
the depth backbone features as shown in Fig.~\ref{activation_map_visualization_of_encoder}. 
% }
Based on it, we define the
% And the 
RGB-D prediction as $\{s^{\mathit{rgbd}}_{init},f^{\mathit{rgbd}}_{\mathit{info}}\} = D_{\mathit{vgg}}(F_{\mathit{rgbd}})$, where $D_{\mathit{vgg}}$ represents the decoder for VGG backbone, which is composed of five $3\times3$ convolutional layers with five sets of feature fusion modules from~\cite{Ranftl2022_midas}.
% as RGB-D encoder,
$s^{\mathit{rgbd}}_{init},f^{\mathit{rgbd}}_{\mathit{info}}$ are the coarse prediction and the corresponding feature.

% After giving the initial predictions and features, we designed the progressive feature fusion module to refine the texture features of RGB images. We use the holistic attention \cite{cpd_sal} based refine module $HaRes$  to generate optimised features $\{f^x_{2\_2},f^x_{3\_2},f^x_{4\_2}\} = HaRes(f^x_2,p_{r})$, which will be utilized in inter-modal fusion.  $HaRes$ is composed of $Ha$ module and two layers of ResNet. And The channels of $\{f^x_{2\_2},f^x_{3\_2},f^x_{4\_2}\}$ are $512,1024,2048$. Additionally, we produced predictions $p_{r\_ref}$ for the RGB branch using the optimised features.
% \begin{equation}
%     \label{init_decoder}
%     \begin{aligned}
%       P_{r\_ref} = D_{res}(f^r_1,f^r_{2\_2},f^r_{3\_2},f^r_{4\_2}) ,
%     \end{aligned}
% \end{equation} 
% where $D_{res}$ is the same decoder model in Eq.\eqref{rgb_decoder}, $P_{r\_ref} = \{p_{r\_ref},f^r_{ref}\}$, $p_{r\_ref}$ is the refined RGB prediction, $f^r_{ref}$ is the improved RGB feature, which contains location information.

% \subsection{Multimodal Feature Fusion}
% \label{sec:feat_fusion}
% \Jing{accordingly, introduce how multimodal features are fused.}
\noindent\textbf{Training-time prediction refinement:} As our scribble annotation is very sparse, directly training with it as supervision is insufficient to recover the image structure information. 
We draw on the approach in \cite{liang2022tree} to optimize predictions using input RGB images with high-dimensional features for self-supervised learning. 
% \noindent\textbf{Tree Energy Loss}
% Since the supervision of semantic segmentation of sparse annotations is weak, 
% Liang \etal~\cite{liang2022tree} introduced tree energy loss to generate soft supervision on unlabeled pixels using the tree affinity of features, which is based on the idea that when annotated and unannotated pixels belong to the same object, low-level features have similar colour information and high-level features share semantic information.
To establish a point-to-point association, the image and features are first transformed into an undirected graph $G=(V, E)$, where $V$ is the set of vertices consisting of all pixels in the image and $E$ denotes the set of edges consisting of the weights of adjacent vertices. The weight between pixels are obtained by calculating the $L2$ distance as $\omega_{u, v}^{k}=\omega_{v, u}^{k}=|I_k(u)-I_k(v)|^{2}$, where $k$ indexes the high/low level features, including coarse features (high-level feature) and the raw input image (low-level feature), $u, v$ are the index of the pixel, and 
% denotes an image or feature map,
$I_k(u)$ represent the $k^{th}$ feature of pixel $u$.

% the value of the pixel, and $i$
Following~\cite{liang2022tree},
% Similar to~\cite{liang2022tree}, 
the minimum spanning tree (MST) is calculated to obtain the distance of the shortest path ($\mathbb{E}^{k}$) between vertices, with which the distance map of the MST is defined as $D^{k}_{u,v}=D^{k}_{v,u}=\sum_{(u',v')\in \mathbb{E}^{k}_{u,v}}\omega^k_{u',v'}$. The corresponding positive affinity matrices are obtained via: $A(k)=\exp(-D^{k}/\sigma)$, where $\sigma$ is used to control the information intensity of $k$. We set $\sigma=0.02$ when $k$ is the RGB image, and $\sigma=1$ when $k$ is the high-level feature.

% by~\cite{gallager1983distributed}. And the affinity matrices can be obtained by calculating the shortest distance between two vertices:
% \begin{equation}
% \begin{aligned}
% \label{input_fea}
%     \mathbb{A}\left [k\right ] =\exp \left(-\frac{1}{\sigma} {\sum_{(m,n)\in \mathcal{E}_{i, j}}\omega_{m,n}^{k} }\right), 
% \end{aligned}
% \end{equation}
% where k denotes an RGB image or a high-level feature map, $\mathcal{E}$ is the distance of the shortest path between vertex $i$ and vertex $j$, and $m,n$ are the indexes of the vertices on edge $\mathcal{E}$. $\sigma$ is a hyperparameter used to control the information intensity of $k$. Referring to \cite{liang2022tree}, we set $\sigma=0.02$ when $k$ is the RGB image, and $\sigma=1$ when $k$ is the high-dimensional feature.

With the affinity matrices, the optimized prediction is obtained via cascaded low/high level filtering:
\begin{equation}
\begin{aligned}
\label{pseudo_eq}
    s^{r}_{ref}&=\mathbb{F}\left( A(f^{r}_{\mathit{info}}),\mathbb{F}\left(A(x^r),s^{r}_{init}\right)\right),\\
    s^{rgbd}_{ref}&=\mathbb{F}\left( A(f^{rgbd}_{\mathit{info}}),\mathbb{F}\left(A(x^r),s^{rgbd}_{init}\right)\right),
\end{aligned}
\end{equation}
where $\mathbb{F}(A,s)=\frac{1}{z}\sum_{v} A_{u,v} s_v$ 
% \sout{$\mathbb{F}(s,A)=\frac{1}{z}\sum_{v} A_{u,v} s_v$ }
is the filtering function, which performs the normalized ($z=\sum_v A_{u,v}$) weight sum across all pixels. Based on the refined prediction,
% , where $A$ is t $\mathbb{F}(M,N)= (1/ {\textstyle \sum_{j}} M_{i,j}) \sum_{j}M_{i,j}N_{j}$. 
% Then 
$L1$ loss is utilized between initial and refined predictions to encourage the unlabeled pixels into model updating via:
% to help train-time refined prediction to supervise initial prediction:
\begin{equation}
\label{tree_energy_loss}
\begin{aligned}
    \mathcal{L}_{tree}^{r}=&\frac{1}{\Omega_U}\sum_{\forall u\in \Omega_U}\left|s^{r}_{init}(u)-s^{r}_{ref}(u) \right|, \\
    \mathcal{L}_{tree}^{rgbd}=&\frac{1}{\Omega_U}\sum_{\forall u\in \Omega_U}|s^{rgbd}_{init}(u)-s^{rgbd}_{ref}(u)|,
    \end{aligned}
\end{equation} 
where $\Omega_U$ represents the unlabeled set, $s^{r}_{init}(u)$ is the prediction at location $u$.

% \Rev{
\begin{algorithm}[!t]
\caption{The First Stage Training Procedure}
\textbf{Input}:  Training dataset $D=\{X_i,y_i\}_{i=1}^N$, where $X=\{x^r,x^d\}$, and maximal number of iterations $T$. \\
\textbf{Output}:  Parameters for encoder $E_{\mathit{vgg}}$, $E_{\mathit{res}}$, decoder $D_{\mathit{vgg}}$, $D_{\mathit{res}}$, fusion module $\mathit{Fu}$, and Mutual Information Optimization module. 
\begin{algorithmic}[1]
    \STATE  Initialize encoder $E_{\mathit{vgg}}$ with  VGG16-Net~\cite{VGG}, encoder $E_{\mathit{res}}$ with ResNet50~\cite{ResHe2015}, and other parameters by default.  
    \FOR {$t \leftarrow 1$ to $T$}  
    \STATE  Extract RGB features $F_r= E_{\mathit{res}}(x^r)$ and depth features $F_{d}= E_{\mathit{vgg}}(x^{d})$;
    % , where $F_{r} = {f_r^1,f_r^2,f_r^3,f_r^4}$ and $F_{d} = {f_d^1,f_d^2,f_d^3,f_d^4,f_d^5}$.
    \STATE  Generate coarse prediction and corresponding feature ${s^{r}_{init},f^r_{\mathit{info}}} = D_{\mathit{res}}(F_{r})$ for RGB branch;
    \STATE  Perform feature fusion using the fusion module: duplicate the highest level of $F_{r}$ to generate $F_{r}'$, and obtained the fused RGB-D  feature $F_{\mathit{rgbd}}=\mathit{Fu}(F_{r}', F_{d})$.
    \STATE  Generate RGB-D prediction and information feature ${s^{\mathit{rgbd}}_{init},f^{\mathit{rgbd}}_{\mathit{info}}} = D_{\mathit{vgg}}(F_{\mathit{rgbd}})$;
    \STATE  Perform training-time prediction refinement (Eq.~\ref{pseudo_eq});
    \STATE  Obtain mutual information estimator (Eq.~\ref{club_loss});
    \STATE Update all parameters via Eq.~\ref{final_loss}.
    \ENDFOR  
     % \RETURN  All parameters
\end{algorithmic}
\label{our_alg}
\end{algorithm}

\subsection{Mutual Information Optimization}
\label{sec:mi_min_reg}
With tree-energy loss~\cite{liang2022tree} as an auxiliary loss function, we aim to achieve training-time prediction refinement. Further, as a typical multimodal learning task, we claim that effective multimodal representation learning is especially important for our task. Given RGB feature $f^{r}_{\mathit{info}}$, and RGB-D feature $f^{\mathit{rgbd}}_{\mathit{info}}$, we aim to explicitly maximize the contribution of depth for RGB-D saliency detection. In this case, we resort to mutual information minimization, which is optimized when $f^{\mathit{rgbd}}_{\mathit{info}}$ contains minimal information from the RGB branch, thus we can maximize the contribution of depth. 
This operation is reasonable as $f^{\mathit{rgbd}}_{\mathit{info}}$ by nature contains extensive information from $f^{r}_{\mathit{info}}$. By pushing them apart, depth contribution can be maximized.

Let $X$ and $Y$ be a pair of random variables, their mutual information $I(X;Y)$ is defined as:
\begin{equation}
    \label{mutual_info_defination}
    I(X;Y)=\mathbb{E}_{p(x,y)}\left[\log \frac{p(x,y)}{p(x)\otimes p(y)}\right],
\end{equation}
which is the Kullback-Leibler divergence of the joint distribution $p{(x,y)}$ from the the product of the marginal distributions $p(x)$ and $p(y)$. As mutual information measures the information that the two variables share, it is typically used as a regularizer in the loss function to encourage (via mutual information maximization) or limit dependency (via mutual information minimization) between variables.
As the log density ratio between the joint distribution $p(x,y)$ and product of marginals $p(x)\otimes p(y)$ is intractable, mutual information is usually estimated instead of computed directly.

In this paper, given RGB-D feature $x=f^{\mathit{rgbd}}_{\mathit{info}}$ and the RGB feature $y=f^{r}_{\mathit{info}}$, we aim to minimize their mutual information, achieving disentangled representation learning.
% , which is intended to allow RGB-D predictions to include more depth information.
% , thus we can achieve explicit depth contribution modeling.
To achieve this, we adopt vCLUB upper bound~\cite{cheng2020club} of mutual information.
Given $X$ and $Y$ with unknown conditional distribution $p(y|x)$ (or $p(x|y)$),~\cite{cheng2020club} introduces contrastive log-ratio upper bound (CLUB) of mutual information between $X$ and $Y$, which is defined as:
\begin{equation}
\begin{aligned}
    \label{club_upper_bound}
    I_{\mathit{vCLUB}}&=\mathbb{E}_{p(x,y)}\left[\log q_\theta(y|x)\right]-\mathbb{E}_{p(x)}\mathbb{E}_{p(y)}\left[\log q_\theta(y|x)\right]\\
    &\geq I(X;Y),
\end{aligned}
\end{equation}
$q_\theta(y|x)$ is a variational distribution with parameters $\theta$ to approximate the true distribution $p(y|x)$, which is also called the \enquote{approximation network}. As Eq.~\eqref{club_upper_bound} involves sampling from both marginal distributions, leading to $\mathcal{O}(N^2)$ computational complexity, where $N$ is the size of the samples. To accelerate the mutual information upper bound computation,~\cite{cheng2020club} uniformly sample negative pairs $(x_i,y_{k'_i})$, treating its probability $q_\theta(y_{k'_i}|x_i)$ as unbiased estimation of the mean of the probability $\mathbb{E}_{p(y)}\left[q_\theta(y|x_i)\right]$. In this case, the sampled mutual information estimator~\cite{cheng2020club} is defined as:
\begin{equation}
    \begin{aligned}
    \label{sample_vclub_upper_bound}
        I_{\mathit{vCLUBs}}=\frac{1}{N}\sum_{i=1}^N\left[\log q_\theta(y_i|x_i)-\log q_\theta(y_{k'_i}|x_i)\right],
    \end{aligned}
\end{equation}
where $N$ is the size of the training data, or the minibatch.

\textit{Technical details:} The approximation network $q_\theta(y|x)$ is modeled with Gaussian distribution following~\cite{cheng2020club}, where we define $q_\theta(y|x)=\mathcal{N}(y;\mu(x),\sigma^2(x))$.
% parameters are Gaussian families~\cite{cheng2020club}, we replace the direct use of $x$ by estimating the mean $\mu(x)$ and variance $\sigma^2(x)$ of $x$.} 
We model $\mu(x)$ with two fully connected layers with a ReLU activation function in the between. We model $\log (\sigma^2(x))$ instead of $\sigma^2(x)$ for stable training, which shares the same structure as $\mu(x)$, except that we add an extra Tanh activation function at the end of $\log (\sigma^2(x))$ to produce a relatively narrow Gaussian distribution.
We define the corresponding RGB feature and RGB-D image as positive pair, and the non-corresponding pair as negative pair for the computation of the sampled mutual information estimator in Eq.~\eqref{sample_vclub_upper_bound}.
% Further, for our multimodal learning framework, we claim that the RGB-D contains the full RGB modality information, in this case, the approximation network $q_\theta(y|x)$ should be uni-directional. 
We perform RGB-D to RGB feature transformation via $q_\theta(f^{r}_{\mathit{info}}|f^{\mathit{rgbd}}_{\mathit{info}})$, achieving mutual information estimator via:
\begin{equation}
\label{club_loss}
\begin{aligned}
    I_{mi}=I_{\mathit{vCLUBs}}(f^{\mathit{rgbd}}_{\mathit{info}},f^{r}_{\mathit{info}}).
\end{aligned}
\end{equation}

% Further, for our multimodal learning framework, we claim that both RGB image and depth data can be the dominant modality, in this case, the approximation network $q_\theta(y|x)$ should be bi-directional. We then perform both RGB-to-depth feature transformation and depth-to-RGB feature transformation, achieving bi-directional mutual information estimator via:
% \begin{equation}
% \label{club_loss}
% \begin{aligned}
%     I_{vbi}=(I_{\mathit{vCLUBs}}(f^{r}_{\mathit{info}},f^{rgbd}_{\mathit{info}})+I_{\mathit{vCLUBs}}(f^{rgbd}_{\mathit{info}},f^{r}_{\mathit{info}}))/2.
% \end{aligned}
% \end{equation}

\subsection{Objective Function of the First Stage Training}
% For the initial predictions, w
We first define partial cross-entropy loss \cite{NCut_loss} between each of the initial predictions and the scribble ground truth with
only labeled pixels used, leading to $\mathcal{L}_{pce}(s^{rgbd}_{init},y) $ and $\mathcal{L}_{pce}(s^{r}_{init},y)$, where $y$ in this case represents the combination of the foreground scribble and background scribble. Then, tree-energy loss in Eq.~\eqref{tree_energy_loss} is incorporated to encourage the contribution of unlabeled pixels. Lastly, the mutual information regularization in Eq.~\eqref{club_loss} is used to explicitly model the contribution of depth for RGB-D saliency detection. Our final loss function is then defined as:

\begin{equation}
    \label{final_loss}
    \mathcal{L} = \mathcal{L}_{r} +\mathcal{L}_{\mathit{rgbd}} + \alpha I_{mi},
\end{equation}
where $\mathcal{L}_{r}$ means the RGB branch loss, which defined as  $\mathcal{L}_{r} = \mathcal{L}_{pce}(s^{r}_{init},y)+\mathcal{L}_{tree}^r$, and  RGB-D branch loss  $\mathcal{L}_{\mathit{rgbd}} $ is defined as $\mathcal{L}_{\mathit{rgbd}} =\mathcal{L}_{pce}(s^{\mathit{rgbd}}_{init},y)+\mathcal{L}_{tree}^{\mathit{rgbd}}$. Empirically we set $\alpha=0.005$ to balance the contribution of the mutual information regularization term for stable training.
We show the first training stage pipeline of our framework in Algorithm \ref{our_alg} for better understanding of the implementation details.

% \begin{equation}
%     \label{init_decoder}
%     \begin{aligned}
%         \mathcal{L}_{r} &=  \mathcal{L}_{scr}(p_r,y_{s}) + \mathcal{L}_{tree}(p_r,x^r,f^r_{init})\\
%         &  + \mathcal{L}_{scr}(p_{r\_ref},y_{s}) + \mathcal{L}_{tree}(p_{r\_ref},x^r,f^r_{ref}) 
%     \end{aligned}
% \end{equation} 

% The tree energy loss is defined as the $L1$ distance between $Pred$ and $\tilde{Y}$ in the unlabeled region.
% \begin{equation}
% \label{tree_energy_loss}
%     \mathcal{L}_{tree}=\frac{1}{\left|\Omega_{U}\right|} \sum_{\forall i \in \Omega_{U}}\left|Pred_{i}-\tilde{Y}_{i}\right|
% \end{equation} 

\subsection{Conditional Multimodal VAE as Refinement}
\label{sec:multimodal_vae}

% \section{Stochastic Prediction Refinement via VAE}
Given dataset $D=\{X_i,y_i\}_{i=1}^N$ as our training dataset, where $i$ indexes the samples and we omit it when it's clear, $X=\{x^r,x^d\}$ is the input RGB ($x^r$) and depth ($x^d$) pair, $y$ is the ground truth saliency map. 
% \sout{$y$ is the group truth saliency map. }
We present a conditional multimodal variational auto-encoder (VAE)~\cite{vae_bayes_kumar,structure_output,multimodal_generative_models_weakly_learning} with explicit unimodal optimization for RGB-D latent representation learning.

\noindent\textbf{Conditional Unimodal VAE:}
Following the maximum likelihood training pipeline, a conditional VAE (CVAE) is trained to maximize the conditional log-likelihood of individual datapoints $\log p_\theta(y|X)=\sum_i \log p_\theta(y|x^i)$. Unfortunately, the computation of $p_\theta(y|x^i)$ involves integral over all configurations of latent variables $z$, which requires exponential time to compute.
To achieve computationally tractable learning of the generative process, \cite{vae_bayes_kumar,structure_output} introduces a recognition model $q_\phi(z|x,y)$ as an approximation of the intractable true posterior $p_\theta(z|x,y)$. The goal is then finding the variational parameters $\phi$ that minimize the Kullback–Leibler (KL) divergence between the variational posterior $q_\phi(z|x,y)$ and the true posterior $p_\theta(z|x,y)$ as:
\begin{equation}
\label{kl_vanilla_vae}
    q_{\phi^*}(z|x,y)=\arg \min_\phi D_{KL}(q_\phi(z|x,y)\|p_\theta(z|x,y)),
\end{equation}
where the KL-divergence term can be decomposed:
\begin{equation}
\begin{aligned}
\label{kl_vae_decomposition}
   &D_{KL}(q_\phi(z|x,y)\|p_\theta(z|x,y))\\
   & = \mathbb{E}_{q_\phi(z|x,y)}\log q_\phi(z|x,y)-\mathbb{E}_{q_\phi(z|x,y)}\log \frac{p_\theta(x,y,z)}{p_\theta(x,y)}.
\end{aligned}
\end{equation}
Given $p_\theta(x,y,z)=p_\theta(y|x,z)p_\theta(z|x)p(x)$, we can further decompose the second expectation term in Eq.~\eqref{kl_vae_decomposition} as:
\begin{equation}
\begin{aligned}
    \label{second_kl_decompose}
    &\mathbb{E}_{q_\phi(z|x,y)}\log \frac{p_\theta(x,y,z)}{p_\theta(x,y)}\\
    &=\mathbb{E}_{q_\phi(z|x,y)}\log \frac{p_\theta(y|x,z)p_\theta(z|x)p(x)}{p_\theta(x,y)}\\
    &=\mathbb{E}_{q_\phi(z|x,y)}\log p_\theta(y|x,z)+\mathbb{E}_{q_\phi(z|x,y)}\log p_\theta(z|x)\\
    &-\log p_\theta(y|x).
\end{aligned}
\end{equation}
We take Eq.~\eqref{second_kl_decompose} back to Eq.~\eqref{kl_vae_decomposition} and obtain:
\begin{equation}
\begin{aligned}
\label{kl_vae_decomposition_final}
   &D_{KL}(q_\phi((z|x,y)\|p_\theta(z|x,y))\\
   & = \mathbb{E}_{q_\phi(z|x,y)}\log q_\phi(z|x,y)-\mathbb{E}_{q_\phi(z|x,y)}\log p_\theta(z|x)\\
   &-\mathbb{E}_{q_\phi(z|x,y)}\log p_\theta(y|x,z)+\log p_\theta(y|x)\\
   &=\underbrace{D_{KL}(q_\phi(z|x,y)\|p_\theta(z|x))-\mathbb{E}_{q_\phi(z|x,y)}\log p_\theta(y|x,z)}_{\text{-ELBO}(x,y,\theta,\phi)}\\
   &+\log p_\theta(y|x).
\end{aligned}
\end{equation}
We simplify Eq.~\eqref{kl_vae_decomposition_final}, and obtain:
\begin{equation}
\begin{aligned}
    \label{simplified_cvae}
    &\log p_\theta(y|x) \\
    &= \text{ELBO}(x,y,\theta,\phi)+D_{KL}(q_\phi(z|x,y)\|p_\theta(z|x,y)).
\end{aligned}
\end{equation}
By Jensen's inequality, $D_{KL}(q_\phi(z|x,y)\|p_\theta(z|x,y))$ in Eq.~\eqref{simplified_cvae} is always greater or equal to zero. In this case, minimizing it can be achieved by maximizing $\text{ELBO}(x,y,\theta,\phi)$, which is the evidence lower bound (ELBO). With the reparameterization trick~\cite{vae_bayes_kumar}, the KL-divergence term in Eq.~\eqref{simplified_cvae} can be solved in closed-form given that both the prior and posterior are Gaussian.

\noindent\textbf{Conditional Multimodal VAE:}
\label{conditional_multimodal_vae}
In our multimodal setting, we obtain the same derivation as in Eq.~\eqref{simplified_cvae}, except that we change $x$ to $X$, representing the multimodal data, leading to ELBO of conditional multimodal VAE as:

\begin{equation}
\begin{aligned}
\label{multimodal_vae_elbo}
    &\text{ELBO}(X,y,\theta,\phi)\\
    &=\mathbb{E}_{q_\phi(z|X,y)}\log p_\theta(y|X,z)-D_{KL}(q_\phi(z|X,y)\|p_\theta(z|X)),
\end{aligned}
\end{equation}

% \begin{equation}
% \begin{aligned}
% \label{multimodal_vae_elbo}
%     &\text{ELBO}(X,y,\theta,\phi)\\
%     &=D_{KL}(q_\phi(z|X,y)\|p_\theta(z|X))-\mathbb{E}_{q_\phi(z|X,y)}\log p_\theta(y|X,z),
% \end{aligned}
% \end{equation}
where $q_\phi(z|X,y)$ and $p_\theta(z|X)$ represent the joint posterior and prior respectively.
To achieve closed-form solution for the KL-divergence term in Eq.~\eqref{multimodal_vae_elbo}, MVAE~\cite{multimodal_generative_models_weakly_learning} uses a product of experts (PoE)~\cite{product_of_expert,Generalized_Product_of_Experts} based on the reparameterization trick, namely the \enquote{prior expert} as estimation of the joint prior and \enquote{posterior expert} as estimation of the joint posterior. Specifically, when each of the experts is Gaussian, their product is still Gaussian, leading to a closed form solution for the KL-divergence term in Eq.~\eqref{multimodal_vae_elbo}. The product of the Gaussian expert will have mean $\mu = (\sum_x \mu_x T_x)(\sum_x T_x)^{-1}$, and covariance $\Sigma = (\sum_x T_x)^{-1}$, where $\mu_x$ and $\Sigma_x$ are the statistics of the Gaussian expert of modality $x$, and $T_x=\Sigma_x^{-1}$ is the inverse covariance or the precision of the Gaussian expert with modality $x$.
With the PoE based joint variational posterior/prior factorizing, we can solve the KL-divergence term in Eq.~\eqref{multimodal_vae_elbo} in closed form.

\noindent\textbf{Joint Prior and Posterior via PoE:} For our conditional generation setting, $X=\{x^r,x^d\}$, including the appearance information from the RGB image $x^r$ and geometric information from the depth data $x^d$. The joint prior is then:
\begin{equation}
\begin{aligned}
\label{joint_prior}
&p(z|X)=p(z|x^r,x^d)=\frac{p(x^r,x^d|z)p(z)}{p(x^r,x^d)}\\
&=\frac{p(z)}{p(x^r,x^d)}\prod_{x}p(x|z)\\
&=\frac{p(z)}{p(x^r,x^d)}\prod_{x}\frac{p(z|x)p(x)}{p(z)}\\
&=\frac{\prod_{x}p(z|x)}{p(z)}\cdot \frac{\prod_x p(x)}{p(x^r,x^d)}\\
&\propto \frac{\prod_{x}p(z|x)}{p(z)}
\end{aligned}
\end{equation}
In practice, the $p(z|x)$ is approximated with $p(z|x)=\hat{q}(z|x)p(z)$ to avoid the quotient term, leading to the final joint prior as:
\begin{equation}
    p_\theta(z|X)\propto \frac{\prod_{x}p(z|x)}{p(z)}\approx p(z)\prod_x \hat{q}(z|x),
\end{equation}
where $\hat{q}(z|x)$ is modeled with a neural network, which is Gaussian. In this case, the joint prior is modeled with the product of Gaussian distributions, and PoE~\cite{product_of_expert} is used to obtain the statistics of $p_\theta(z|X)$.

Similarly, the joint posterior $q(z|X,y)$ is obtained via:
\begin{equation}
\begin{aligned}
    &q(z|X,y)=q(z|x^r,x^d,y)=\frac{p(x^r,x^d, y|z)p(z)}{p(x^r,x^d, y)}\\
    &=\frac{p(z)}{p(x^r,x^d, y)}\prod_x p(x,y|z)\\
    &=\frac{p(z)}{p(x^r,x^d, y)}\prod_x \frac{p(z|x,y)p(x,y)}{p(z)}\\
    &=\frac{\prod_x p(z|x,y)}{p(z)}\cdot \frac{\prod_x p(x,y)}{p(x^r,x^d, y)}\\
    &\propto \frac{\prod_{x}p(z|x,y)}{p(z)}
\end{aligned}
\end{equation}
We also approximate $p(z|x,y)$ with $p(z|x,y)=\hat{q}(z|x,y)p(z)$, and define the joint posterior as:
\begin{equation}
    p_\theta(z|X,y)\propto \frac{\prod_{x}p(z|x,y)}{p(z)}\approx p(z)\prod_x \hat{q}(z|x,y),
\end{equation}
where $\hat{q}(z|x,y)$ is Gaussian and modeled with a neural network, leading to the product of Gaussian distribution for the joint posterior distribution, which can be computed in closed form with PoE~\cite{product_of_expert}.

% \section{Weakly-Supervised Conditional VAE}
\noindent\textbf{Conditional Multimodal VAE as Refinement:}
In order to refine the prediction and mitigate error propagation issues, we explore the potential of Conditional Multimodal VAE~\cite{multimodal_generative_models_weakly_learning} and apply it to our weakly-supervised learning setting, leading to a two-stage training pipeline.
% propose to construct a more robust optimization model based on potential distribution by balancing RGB and depth using Conditional Multimodal VAE.
Within the fully-supervised conditional VAE framework, the conditional variable $x$ is used to produce the conditional prediction $p(y|x,z)$. Given our weakly supervised setting with scribble annotation, the label $y$ is partial, directly modeling the posterior distribution $q_\phi(z|x,y)$ is problematic. Instead, we use the 
% Although 
pseudo label from the first step training as $y$ in this case, and all the other derivations are the same as the conditional multimodal VAE in Sec.~\ref{conditional_multimodal_vae}.

\begin{figure*}[h]
\begin{center}
  \begin{tabular}{c@{ }c@{ } c@{ } c@{ } c@{ } c@{ } }

{\includegraphics[width=0.152\linewidth,height=0.10\linewidth]{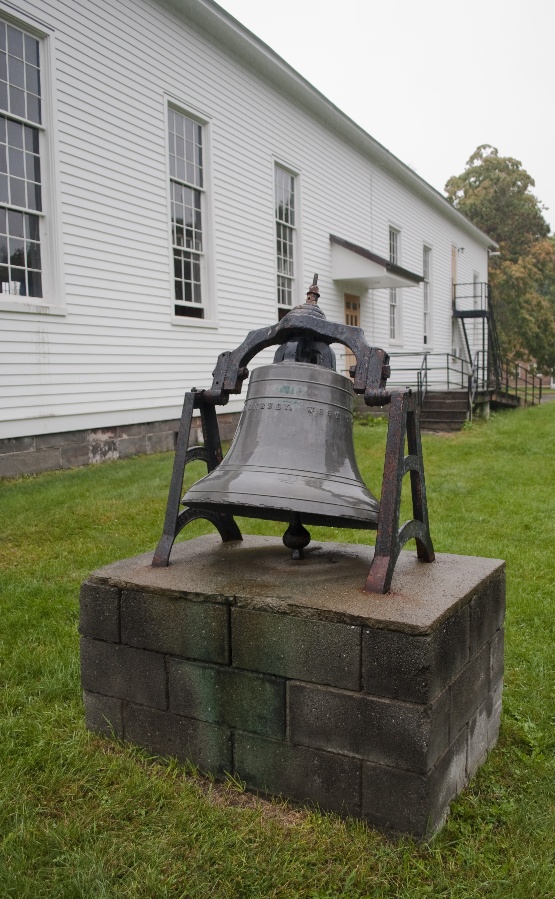}}&
{\includegraphics[width=0.152\linewidth,height=0.10\linewidth]{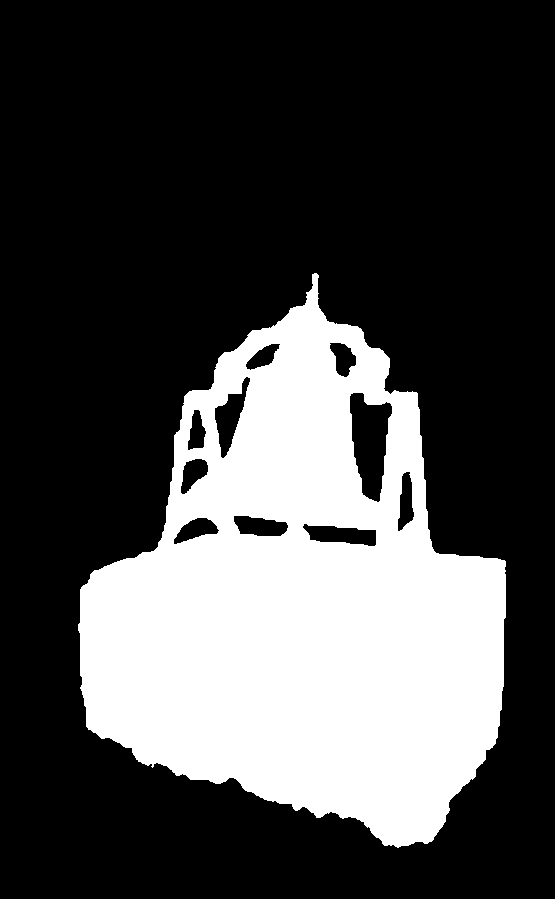}}&
{\includegraphics[width=0.152\linewidth,height=0.10\linewidth]{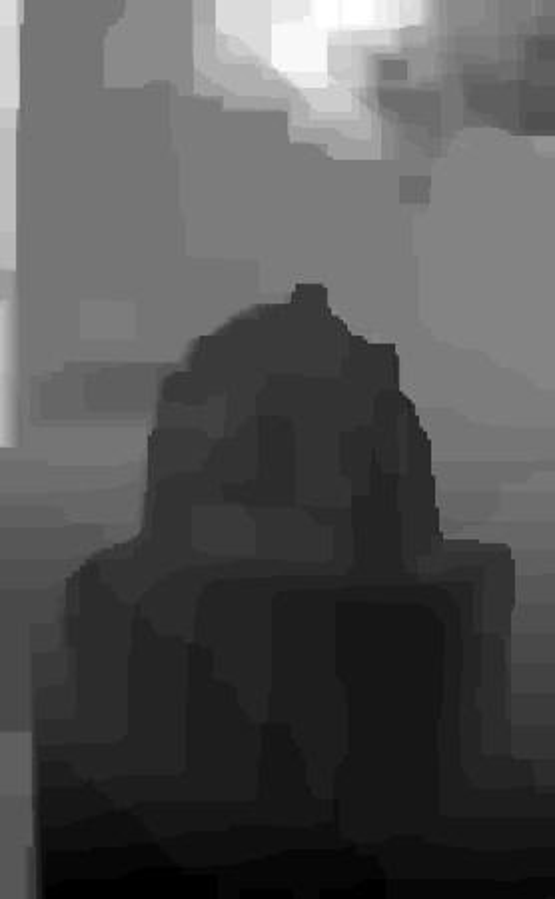}}&
{\includegraphics[width=0.152\linewidth,height=0.10\linewidth]{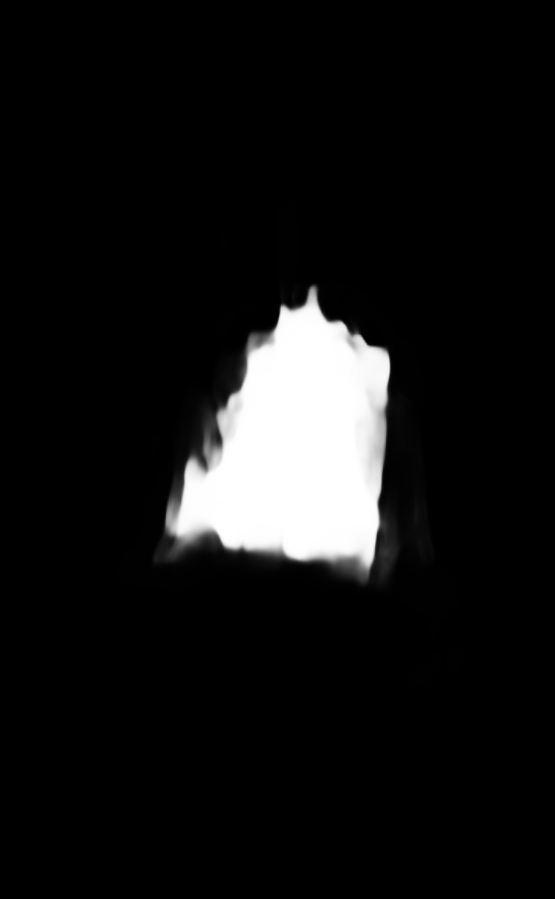}}&
{\includegraphics[width=0.152\linewidth,height=0.10\linewidth]{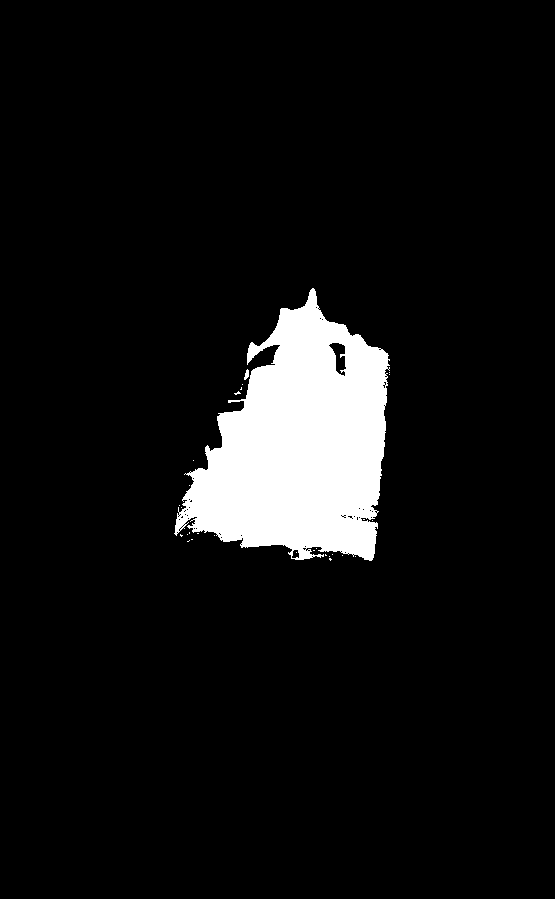}}&
{\includegraphics[width=0.152\linewidth,height=0.10\linewidth]{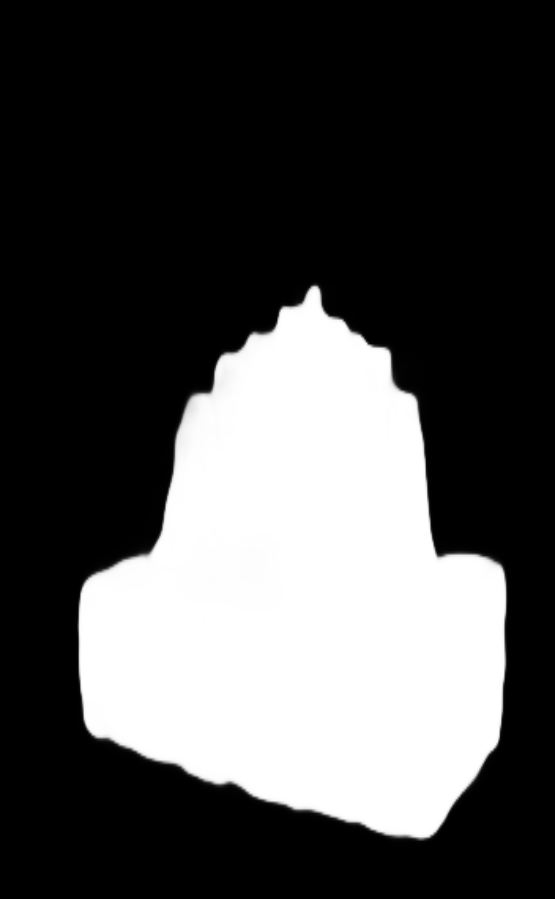}} \\

{\includegraphics[width=0.152\linewidth,height=0.10\linewidth]{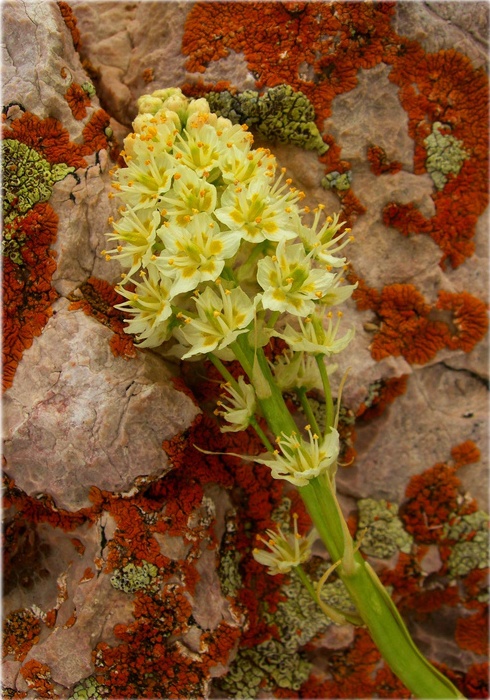}}&
{\includegraphics[width=0.152\linewidth,height=0.10\linewidth]{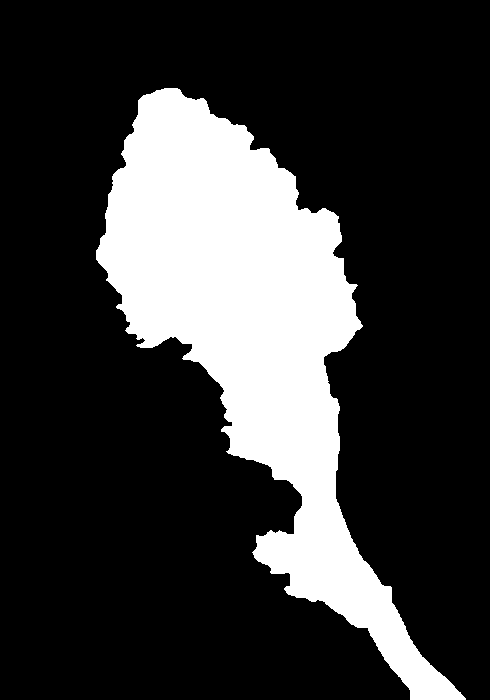}}&
{\includegraphics[width=0.152\linewidth,height=0.10\linewidth]{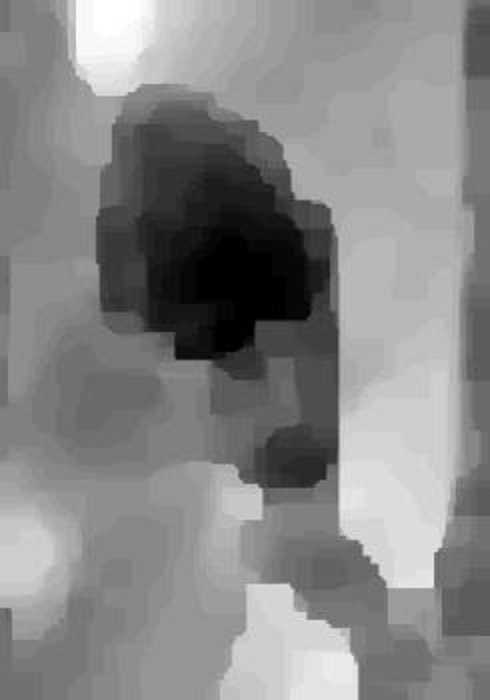}}&
{\includegraphics[width=0.152\linewidth,height=0.10\linewidth]{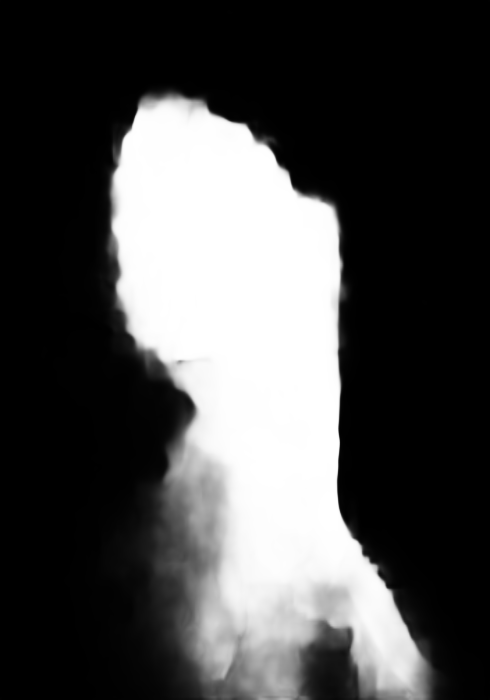}}&
{\includegraphics[width=0.152\linewidth,height=0.10\linewidth]{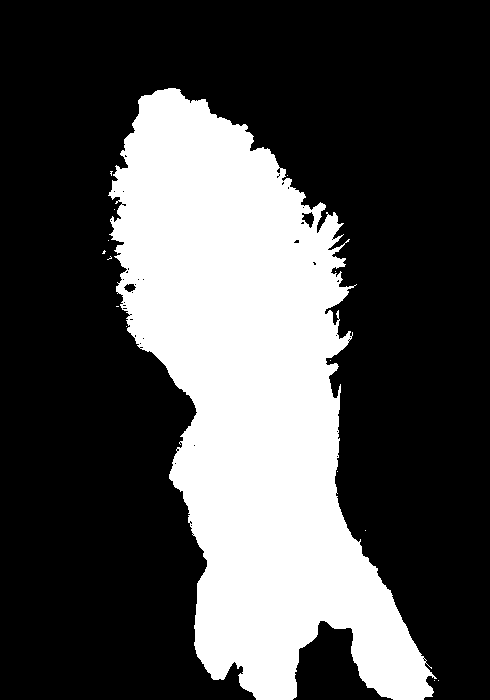}}&
{\includegraphics[width=0.152\linewidth,height=0.10\linewidth]{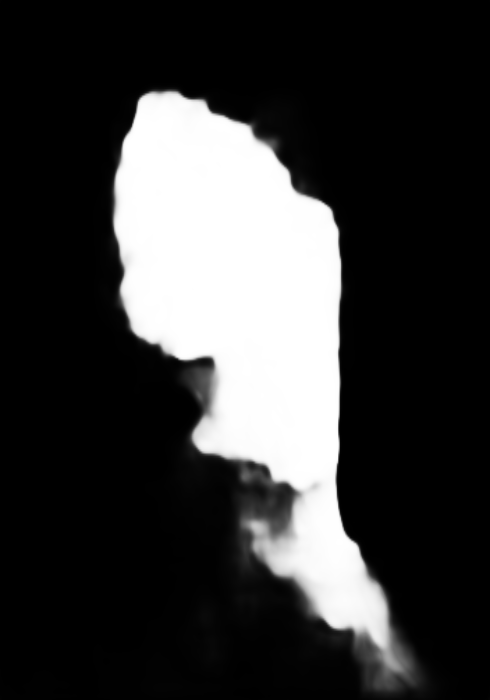}} \\

\footnotesize{Image}&\footnotesize{GT}&\footnotesize{Depth}&\footnotesize{$Pred_{\mathit{raw}}$}& \footnotesize{$Pred_{\mathit{crf}}$}&\footnotesize{$Pred_{\mathit{ours}}$}
  \end{tabular}
  \end{center}
   % \vspace{-5mm}
\caption{Prediction comparison,
% of the CRF refine and our proposed Conditional Multimodal VAE refine, 
where \enquote{Image}, \enquote{GT} and \enquote{Depth} are the RGB image, the ground truth map, and the depth map.  \enquote{$Pred_{\mathit{raw}}$}, \enquote{$Pred_{\mathit{crf}}$} and \enquote{$Pred_{\mathit{ours}}$} represent prediction after the first stage training, refinement with denseCRF~\cite{dense_crf} and the proposed Conditional Multimodal VAE, respectively.
% means the prediction of the first stage,  \enquote{$Pred_{\mathit{crf}}$} means the prediction of CRF refinement, and  \enquote{$Pred_{\mathit{ours}}$} means the prediction of Conditional Multimodal VAE refinement.
}
\label{refine_visualization_init}
\end{figure*}

With the first stage of training, we obtain relatively reasonable predictions on the training dataset. However, we observe less accurate predictions happen along object boundaries, which can be refined by post-processing techniques, \ie~denseCRF~\cite{dense_crf}
 as shown in Fig.~\ref{refine_visualization_init}.
However, the performance of denseCRF~\cite{dense_crf} relies on carefully picked hyper-parameters, making it not robust. According to the noise robustness learning~\cite{neelakantan2015adding}, where adding noise to the network can prevent the model from overfitting, we introduce
% due to the spar
% We further introduce 
stochastic prediction refinement by adding structured noise
% as the second stage training.
% With the proposed explicit weakly-supervised RGB-D saliency detection model, we can obtain similar predictions compared with the state-of-the-art models. We further introduce a refinement strategy 
via a 
conditional multimodal variational auto-encoder (CVAE)~\cite{vae_bayes_kumar,structure_output,multimodal_generative_models_weakly_learning} as the second stage training,
% to train a pixel-level supervised model, 
which is proven relatively robust to labeling noise, suffering less error propagation issues.

The goal of CVAE is finding the variational parameters $\phi$ that minimize the Kullback–Leibler (KL) divergence ($D_{\text{KL}}$) between the variational posterior $q_\phi(z|X,y)$ and the true posterior $p_\theta(z|X,y)$ via maximizing the conditional log-likelihood, which is usually achieved via optimizing the
% \begin{equation}
% \label{kl_vanilla_vae}
%     q_{\phi^*}(z|X,y)=\arg \min_\phi D_{KL}(q_\phi(z|X,y)\|p_\theta(z|X,y)),
% \end{equation}
% where $X$ is the multimodal data.
% Based on Jensen's inequality and Bayes' rule, we
% optimize the 
evidence lower bound (ELBO)\footnote{The complete derivation is provided in the supplementary material}:
% which is defined as: 
% Following the conventional practice of learning a variation auto-encoder, we optimize the evidence lower bound (EBLO) instead of the real conditional log-likelihood ($\log p_\theta(y|X)\geq\text{ELBO}(X,y,\theta,\phi)$),
% where the ELBO is defined as:
\begin{equation}
\begin{aligned}
\label{multimodal_vae}
\text{ELBO}(X,y,\theta,\phi)&=\underbrace{\lambda\mathbb{E}_{q_\phi(z|X,y)}\log p_\theta(y|X,z)}_{\text{conditional generation}}\\
&-\beta\text{D}_{\text{KL}}(q_\phi(z|X,y)||p_\theta(z|X)),
\end{aligned}
\end{equation}
where $\lambda$ and $\beta$ are hyper-parameters to balance the two parts~\cite{beta_vae}. For our multimodal data setting, both the prior and posterior of $z$ are based on the multimodal data $X$, which are defined as the \enquote{joint prior} and \enquote{joint posterior} respectively. We achieve them via product of experts (PoE)~\cite{product_of_expert,Generalized_Product_of_Experts}, leading to both the \enquote{prior expert} as estimation of the \enquote{joint prior expert} and the \enquote{joint posterior expert}, which are both Gaussian. In this case, the KL-divergence term in Eq.~\eqref{multimodal_vae} can be solved in closed form.

\textit{Multimodal VAE as Refinement:}
% \Jing{describe how this is achieved.}
The multimodal VAE refinement process is based on the conditional VAE framework, where we use pseudo label $s_{ref}^{rgbd}$ from the first stage training
% and scribble 
as supervisions. We follow a simpler training pipeline than the first stage training, each modality has a ResNet50 encoder with a similar multimodal fusion $F_u$ operation and the decoder for the ResNet50 backbone is used to generate the refined prediction.
% and directly concatenate feeding into the decoder. 
Further, we use the same structure neural network to obtain both the \enquote{prior expert} and \enquote{posterior expert}, and generate our prediction only for the RGB-D branch for simplicity. 
% and sampled from the latter to generate our prediction for the RGB branch and RGB-D branch. 
To obtain the \enquote{prior expert} $p_\theta(z|X)$, we follow PoE~\cite{product_of_expert,Generalized_Product_of_Experts} and obtain:
\begin{equation}
\label{prior_expert}
    p_\theta(z|X)\propto \frac{\prod_{x}p(z|x)}{p(z)}\approx p(z)\prod_x \hat{q}(z|x),
\end{equation}
where $\hat{q}(z|x)$ is modeled with a neural network, which is Gaussian. In this case, the joint prior is modeled with the product of Gaussian distributions.
% , and PoE~\cite{product_of_expert} is used to obtain the statistics of $p_\theta(z|X)$.

Similarly, the \enquote{posterior expert} can be obtained as:
\begin{equation}
\label{posterior_expert}
    p_\theta(z|X,y)\propto \frac{\prod_{x}p(z|x,y)}{p(z)}\approx p(z)\prod_x \hat{q}(z|x,y),
\end{equation}
where $\hat{q}(z|x,y)$ is Gaussian and modeled with a neural network, leading to the product of Gaussian distribution for the joint posterior distribution.
% which can be computed in closed form with PoE~\cite{product_of_expert}.

\begin{table*}[t!]
  \centering
  \scriptsize
  \renewcommand{\arraystretch}{1.25}
  \renewcommand{\tabcolsep}{0.2mm}
  \caption{Benchmarking results of SOD models.
  %NB 3 
%   leading fully/weakly/un-supervised deep RGB-D saliency detection models
%   on six RGB-D saliency datasets.
  $\uparrow \& \downarrow$ denote the larger and smaller is better, respectively and
%   Here, we adopt mean $F_{\beta}$ and mean $E_{\xi}$.
\enquote{Bkb}: the backbone. V16, R50, PVTV2 and Swin are VGG16, ResNet50 , PVTV2\cite{wang2021pyramid} and Swin~\cite{liu2021Swin} backbones, respectively. Mob: MobileNetV2 backbone\cite{sandler2018mobilenetv2}. CNN: the backbone from \cite{Fu2020JLDCF} with six hierarchies. 3DC: the backbone from \cite{Chen_Liu_Zhang_Fu_Zhao_Du_2021} with 3D convolutions. .
% ,
% which is basically a ResNet/VGG-like backbone with 3D convolutions. 
ViT: T2T-ViT backbone~\cite{yuan2021tokens}.
  } % \Rev{SFNet: ShuffeNet backbone\cite{ma2018shufflenet}}
  \begin{tabular}{ll|l|cccc|cccc|cccc|cccc|cccc|cccc}
  \hline
%   \toprule
  &&&\multicolumn{4}{c|}{NJU2K\cite{NJU2000}}&\multicolumn{4}{c|}{SSB \cite{niu2012leveraging}}&\multicolumn{4}{c|}{DES \cite{cheng2014depth}}&\multicolumn{4}{c|}{NLPR \cite{peng2014rgbd}}&\multicolumn{4}{c|}{LFSD \cite{li2014saliency}}&\multicolumn{4}{c}{SIP \cite{sip_dataset}} \\
    Method &Year&BkB
    & $S_{\alpha}\uparrow$ & $F_{\beta}\uparrow$ & $E_{\xi}\uparrow$ & $\mathcal{M}\downarrow$
    & $S_{\alpha}\uparrow$ & $F_{\beta}\uparrow$ & $E_{\xi}\uparrow$ & $\mathcal{M}\downarrow$
    & $S_{\alpha}\uparrow$ & $F_{\beta}\uparrow$ & $E_{\xi}\uparrow$ & $\mathcal{M}\downarrow$
    & $S_{\alpha}\uparrow$ & $F_{\beta}\uparrow$ & $E_{\xi}\uparrow$ & $\mathcal{M}\downarrow$
    & $S_{\alpha}\uparrow$ & $F_{\beta}\uparrow$ & $E_{\xi}\uparrow$ & $\mathcal{M}\downarrow$
    & $S_{\alpha}\uparrow$ & $F_{\beta}\uparrow$ & $E_{\xi}\uparrow$ & $\mathcal{M}\downarrow$ \\
  \hline
  \multicolumn{27}{c}{Early Fusion Models} \\ \hline
    % DF~\cite{qu2017rgbd} & 2017& . & . & . & . & . & . & . & . & . & . & . & . & . & . & . & . & . & . & . & . & . & . & . & .    \\
    DANet~\cite{DANet} & 2020 & V16 & .897 & .877 & .926 & .046 & .892 & .857 & .915 & .048 & .905 & .848 & .961 & .028 & .908 & .850 & .945 & .031 & .845 & .826 & .872 & .082 & .878 & .829 & .914 & .054    \\
    UCNet~\cite{jing2020uc} & 2020& R50 & .897 & .886 & .930 & .043 & .903 & .884 & .938 & .039 & .934 & .919 & .967 & .019 & .920 & .891 & .951 & .025 & .864 & .855 & .901 & .066 & .875 & .867 & .914 & .051    \\
    JLDCF~\cite{Fu2020JLDCF} & 2020& CNN & .902 & .885 & .935 & .041 & .903 & .873 & .936 & .040 & .931 & .907 & .959 & .021 & .925 & .894 & .955 & .022 & .862 & .848 & .894 & .070 & .880 & .873 & .918 & .049    \\
    RD3D~\cite{Chen_Liu_Zhang_Fu_Zhao_Du_2021}& 2021& 3DC & .927 & .909 & .943 & .033 & .914 & .881 & .932 & .039 & .950 & .923 & .968 & .017 & .933 & .898 & .953 & .022 & - & - & - & - & .892 & .881 & .917 & .046    \\
    \hline
   \multicolumn{27}{c}{Late Fusion Models} \\ \hline
    % LHM~\cite{peng2014rgbd}  & 2014& . & . & . & . & . & . & . & . & . & . & . & . & . & . & . & . & . & . & . & . & . & . & . & .    \\
    % DESM~\cite{cheng2014depth} & 2014& . & . & . & . & . & . & . & . & . & . & . & . & . & . & . & . & . & . & . & . & . & . & . & .    \\
    % CDB~\cite{liang2018stereoscopic} & 2018&  & . & . & . & . & . & . & . & . & . & . & . & . & . & . & . & . & . & . & . & . & . & . & . & .    \\
    AFNet~\cite{wang2019adaptive} & 2019& V16 & .822 & .827 & .867 & .077 & .825 & .806 & .872 & .075 & .770 & .713 & .809 & .068 & .799 & .755 & .851 & .058 & .738 & .736 & .796 & .134 & .720 & .702 & .793 & .118    \\
    A2dele~\cite{A2dele_cvpr2020} & 2020& V16 & .873 & .867 & .913 & .051 & .876 & .874 & .925 & .044 & .881 & .868 & .913 & .030 & .887 & .871 & .933 & .031 & .831 & .829 & .872 & .076 & .826 & .827 & .887 & .070    \\ \hline
    % CTMF~\cite{han2017cnns} & 2017& . & . & . & . & . & . & . & . & . & . & . & . & . & . & . & . & . & . & . & . & . & . & . & .    \\ \hline
    \multicolumn{27}{c}{Cross-level Fusion Models} \\ \hline
    MSal~\cite{wu2021mobilesal}&2021 & Mob & .910 & .891 & .939 & .039 & .903 & .874 & .930 & .041 & .929 & .908 & .957 & .021 & .920 & .885 & .950 & .025 & .847 & .825 & .879 & .080 & .873 & .861 & .908 & .053    \\
    DSA2F~\cite{Sun_2021_CVPR_DSA2F} & 2021&  V19 &.903 &.901 &.923 &.039  &.904 &.898 &.933 &.036 &.920 &.896 &.962 &.021 &.918 &.897 &.950 &.024 &.882 &.878 &.919 &.055  &- &- &- &- \\ 
    VST~\cite{Liu_2021_ICCV_VST} & 2021 & ViT &.922 &.898 &.939 &.035 &.913 &.879 &.937 &.038 &.943 &.920 &.965 &.017 &.932 &.897 &.951 &.024 &.890 &.871 &.917 &.054  &.904 &.894 &.933 &.040 \\ 
    GTSOD~\cite{jingnips2021} &2021& Swin &.929 &.924 &.956 &.028 &.916 &.898 &.950 &.032&.945 &.928 &.971 &.016 &.938 &.921 &.966 &.018&.872 &.862 &.901 &.066  &.906 &.908 &.940 &.037  \\
    C2DF~\cite{criss_cross_tmm2022}&2022 & R50 & .908 & .898 & .936 & .038 & .902 & .881 & .936 & .038 & .921 & .908 & .945 & .020 & .927 & .904 & .955 & .021 & .863 & .859 & .897 & .065 & .871 & .865 & .912 & .053    \\
    SSLSOD~\cite{SSLSOD}&2022 & V16  & .902 & .887 & .929 & .043 & .886 & .867 & .921 & .048 & .933 & .919 & .966 & .019 & .914 & .881 & .941 & .027 & - & - & - & - & .870 & .862 & .900 & .059    \\
    CLNet~\cite{cascaded_rgbd_sod} &2021 &  R50& .939 & .925 & .956 & .032 & .921 & .895 & .959 & .034 & .953 & .926 & .970 & .015 & .941 & .909 & .964 & .019 & .877 & .862 & .911 & .064 & .894 & .887 & .933 & .044    \\ 
    
    % \Rev{ MLFFL~\cite{huang2022middle}} &2022 &  SFNet& .898 & .885 & .925 & .042 & -   & -   & -   & -     & -   & -   & -   & -   & .917 & .887 & .943 & .027  & -   & -   & -   & -   & .882 & .871 & .919 & .048  \\  % STEREO is Not STERE1K(SSB)
    % \Rev{ DCF~\cite{ji2021calibrated}} &2022 &  R50 & -   & -   & -   & -  & .902 & .886 & .939 & .039 & .905 & .880 & .938 & .024 & .924 & .897 & .957 & .022 & .842 & .835 & .877 & .075 & .876 & .874 & .915 & .052 \\% predtrained map (trained on NJUD & NLPR & DUT)
     DCF~\cite{ji2021calibrated} &2022 & R50 & .890 & .880 & .924 & .045 & .886 & .870 & .926 & .044 & .895 & .868 & .924 & .026 & .907 & .878 & .941 & .028 & .807 & .801 & .842 & .094 & .848& .841 & .892 & .067  \\ % code
     CIRNet~\cite{cong2022cir} &{2022 }&{  R50 }&{ .901 }&{ .880 }&{ .917 }&{ .047 }&{ .901 }&{ .872 }&{ .914 }&{ .046 }&{ .906 }&{ .871 }&{ .913 }&{ .029 }&{ .920 }&{ .881 }&{ .937 }&{ .028 }&{ .822 }&{ .803 }&{ .834 }&{ .096 }&{ .861 }&{ .840 }&{ .886 }&{ .069 } \\
    % \Rev{ DCF~[24]} &2022 &  R50 & .890 & .880 & .924 & .045 & .886 & .870 & .926 & .044 & .895 & .868 & .924 & .026 & .907 & .878 & .941 & .028 & .807 & .801 & .842 & .094 & .848 & .841 & .892 & .067 \\ % code
    % \Rev{ CIRNet~[116]} &2022 &  R50 & .901 & .880 & .917 & .047 & .901 & .872 & .914 & .046 & .906 & .871 & .913 & .029 & .920 & .881 & .937 & .028 & .822 & .803 & .834 & .096 & .861 & .840 & .886 & .069 \\    

    \hline
    \multicolumn{27}{c}{Weak/Un-supervised Models} \\ \hline
    SSOD~\cite{xu2022weakly} & 2022& V16  & .899 & .878 & .922 & .044 & .887 & .851 & .914 & .044 & .916 & .887 & .939 & .029 & .913 & .874 & .936 & .028 & .832 & .814 & .860 & .085 & .865 & .841 & .900 & .056 \\
    %   SSOD\_noboost~\cite{xu2022weakly} & 2022& V16 & .877 & .864 & .922 & .051 & .876 & .854 & .926 & .048 & .885 & .861 & .921 & .031 & .900 & .869 & .944 & .031 & .826 & .815 & .874 & .086 & .846 & .827 & .905 & .063    \\ \hline
    DSU~\cite{ji2022promoting} &2022 &  R50& .734 & .693 & .752 & .133 & .797 & .759 & .826 & .097 & .826 & .764 & .828 & .061 & .809 & .739 & .836 & .064 & .780 & .761 & .795 & .128 & .693 & .613 & .734 & .155 \\
    WSSOD~\cite{jing2020weakly} &2020 & V16 & .851 & .840 & .905 & .062 & .860 & .843 & .915 & .054 & .858 & .840 & .918 & .039 & .872 & .842 & .925 & .038 & .821 & .821 & .871 & .087 & .811 & .795 & .868 & .082  \\
    SCWS~\cite{yu2021structure} &2021 & R50 & .843 & .822 & .896 & .065 & .841 & .808 & .895 & .064 & .862 & .835 & .919 & .035 & .868 & .827 & .916 & .042 & .794 & .776 & .845 & .103 & .825 & .792 & .885 & .074 \\
    JSM~\cite{li2021joint} &2021 & R50 & .724 & .689 & .746 & .130 & .790 & .758 & .822 & .093 & .826 & .772 & .823 & .056 & .810 & .749 & .841 & .059 & .766 & .756 & .790 & .128 & .707 & .640 & .740 & .142 \\ 
   {DLM~\cite{yang2022depth}} &{2022} &  {PVTV2} &{ .808 }&{ .765 }&{ .837 }&{ .097 }&{ .833 }&{ .790 }&{ .869 }&{ .079 }&{ .833 }&{ .749 }&{ .871 }&{ .060 }&{ .796 }&{ .668 }&{ .811 }&{ .081 }&{ .824 }&{ .807 }&{ .851 }&{ .098 }&{ .777 }&{ .727 }&{ .830 }&{ .109 } \\ \hline
    % \Rev{ DLM~[118]} &2022 &  PVTV2 & .808 & .765 & .837 & .097 & .833 & .790 & .869 & .079 & .833 & .749 & .871 & .060 & .796 & .668 & .811 & .081 & .824 & .807 & .851 & .098 & .777 & .727 & .830 & .109 \\ \hline
    
    Ours & 2022& RV &.890 & .880 & .929 & .046 & .891 & .873 & .934 & .042 & .928 & .921 & .965 & .019 & .914 & .895 & .953 & .025 & .849 & .844 & .889 & .072 & .876 & .863 & .924 & .049 
    \\
   \hline
  \end{tabular}
  \label{tab:BenchmarkResults_old_training}
%   \vspace{-5mm}
\end{table*}

\textit{PoE statistics computation:}
Specifically, when each of the experts is Gaussian, their product is still Gaussian. % leading to closed form solution for the KL-divergence term in Eq.~\eqref{multimodal_vae}. 
The product of the Gaussian expert will have mean $\mu = ({\textstyle \sum_k} \mu_k T_k)({\textstyle \sum_k} T_k)^{-1}$, and covariance $\sigma^{2}  = ({\textstyle \sum_k} T_k)^{-1}$, where $\mu_k$ and $\sigma_k$ are the statistics of the Gaussian expert of modality $k$, and $T_k=1/\sigma_k^2$ is the inverse covariance or the precision of the Gaussian expert. % with modality $k$.
With the PoE based joint variational posterior/prior factorizing, we can solve the KL-divergence term in Eq.~\eqref{multimodal_vae} in closed form.

\textit{Technical details:} Within both the prior expert in Eq.~\eqref{prior_expert} and posterior expert in Eq.~\eqref{posterior_expert}, we need to compute the unimodal prior and posterior respectively. In this paper, we design them using the same structure, which consists of five $4\times4$ convolutional layers with batch normalization and Leaky Relu, and two linear layers with the same structure to obtain $\mu$ and $\log({\sigma^{2}})$ respectively.
% xxx.\Jing{finish the structure introduction}.
During training, the reconstruction loss
(to maximize the log-likelihood of the conditional distribution in Eq.~\eqref{multimodal_vae}) is defined as the pixel-wise structure-aware loss~\cite{F3Net_aaai2020} between prediction from the RGB-D branch and the pseudo label $s^{\mathit{rgbd}}_{\mathit{ref}}$ as:
\begin{equation}
\begin{aligned}
    &\arg\max_{\theta,\phi}\mathbb{E}_{q_\phi(z|X,y)}\log p_\theta(y|X,z)\\
    =&\arg\min_{\theta,\phi}\mathcal{L}(\mathbb{E}_{q_\phi(z|X,y)} \left[p_\theta(y|X,z)\right],s^{\mathit{rgbd}}_{\mathit{ref}})
\end{aligned}
\end{equation}
% the sum of the
% loss function for pixel-level supervision is the  
% structure-aware loss \cite{F3Net_aaai2020} between prediction from the RGB-D branch and the pseudo label $s^{\mathit{rgbd}}_{\mathit{ref}}$. 
% \sout{and the mutual information loss $I_{mi}$ in Eq.~\eqref{club_loss}.}
The KL-divergence loss within the ELBO in Eq.~\eqref{multimodal_vae} is the divergence of the joint \enquote{posterior expert} from the joint \enquote{prior expert}, where we set $\lambda=1$ and $\beta=5$
% and our first stage training loss function in Eq.~\eqref{club_loss}\eqref{partial_ce_loss}. 
% Moreover, we introduce auxiliary $\beta D_{\text{KL}}$ from Eq.~\eqref{multimodal_vae} and set $\beta=5$ 
with linear annealing for stable training~\cite{beta_vae}.
% \Rev{We did not include the mutual information loss $I_{mi}$ in Eq.~\eqref{club_loss} because in the first training stage, it was difficult to accurately utilize RGB information due to the use of scribble as supervision. Therefore, we maximized the contribution of depth. In this stage, we utilized not-bad pseudo labels to jointly combine RGB and depth information in the distribution based on Eq.~\eqref{multimodal_vae}.}
Note that, for the generation process, latent code is involved.
% Also, as latent code is involved during training, w
We adjust the previous decoder for ResNet50 to take the latent code as input, where we simply tile the latent code to the same size as the top level feature, and concatenate them, which is then fed to another $3\times3$ convolution layer to obtain the new top level feature of the same size as the previous one. The other structure is the same as the decoder for the ResNet50 backbone discussed in Sec.~\ref{sec:feat_extractors}.

\begin{figure*}[!htp]
\begin{center}
  \begin{tabular}{c@{ }c@{ } c@{ } c@{ } c@{ } c@{ } c@{ } c@{ } }
%   {\includegraphics[width=0.116\linewidth]{figures/compare_sod/RGBD_data_42_a_img.png}}&
%   {\includegraphics[width=0.116\linewidth]{figures/compare_sod/RGBD_data_42_a_dep.pdf}}&
%   {\includegraphics[width=0.116\linewidth]{figures/compare_sod/RGBD_data_42_a_gt.png}}&
%   {\includegraphics[width=0.116\linewidth]{figures/compare_sod/RGBD_data_42_f_SSLSOD.png}}&
%   {\includegraphics[width=0.116\linewidth]{figures/compare_sod/RGBD_data_42_f_VST.png}}&
%   {\includegraphics[width=0.116\linewidth]{figures/compare_sod/RGBD_data_42_w_DES.png}}&
%   {\includegraphics[width=0.116\linewidth]{figures/compare_sod/RGBD_data_42_w_SCWS.png}}&
%   {\includegraphics[width=0.116\linewidth]{figures/compare_sod/RGBD_data_42_w_our.png}} \\
%   {\includegraphics[width=0.116\linewidth]{figures/compare_sod/RGBD_data_108_a_img.png}}&
%   {\includegraphics[width=0.116\linewidth]{figures/compare_sod/RGBD_data_108_a_dep.pdf}}&
%   {\includegraphics[width=0.116\linewidth]{figures/compare_sod/RGBD_data_108_a_gt.png}}&
%   {\includegraphics[width=0.116\linewidth]{figures/compare_sod/RGBD_data_108_f_SSLSOD.png}}&
%   {\includegraphics[width=0.116\linewidth]{figures/compare_sod/RGBD_data_108_f_VST.png}}&
%   {\includegraphics[width=0.116\linewidth]{figures/compare_sod/RGBD_data_108_w_DES.png}}&
%   {\includegraphics[width=0.116\linewidth]{figures/compare_sod/RGBD_data_108_w_SCWS.png}}&
%   {\includegraphics[width=0.116\linewidth]{figures/compare_sod/RGBD_data_108_w_our.png}} \\
  {\includegraphics[width=0.116\linewidth]{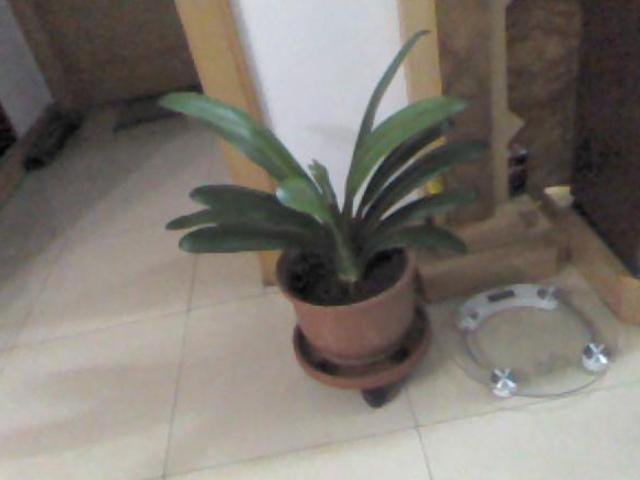}}&
  {\includegraphics[width=0.116\linewidth]{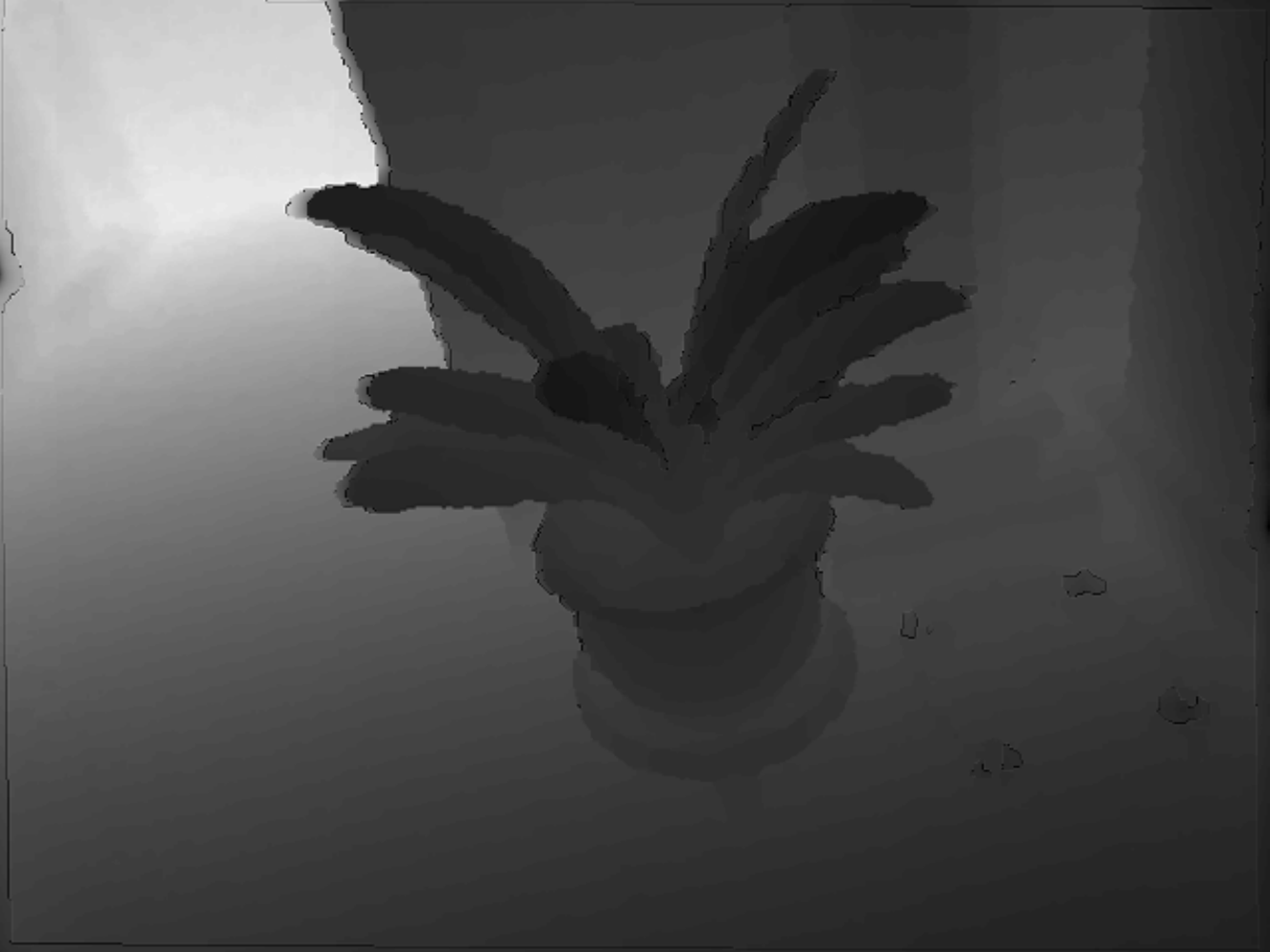}}&
  {\includegraphics[width=0.116\linewidth]{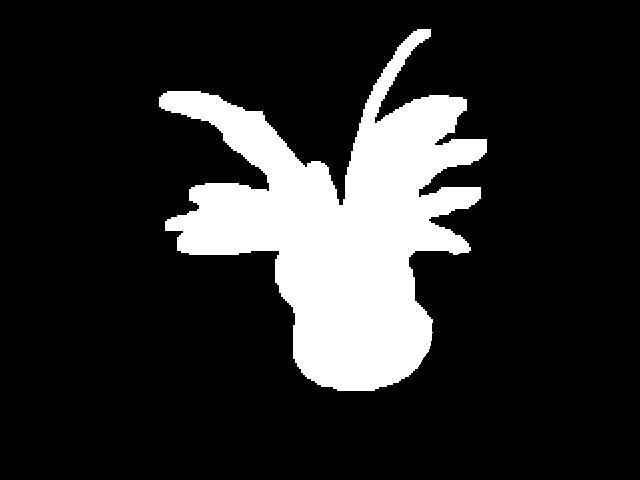}}&
  {\includegraphics[width=0.116\linewidth]{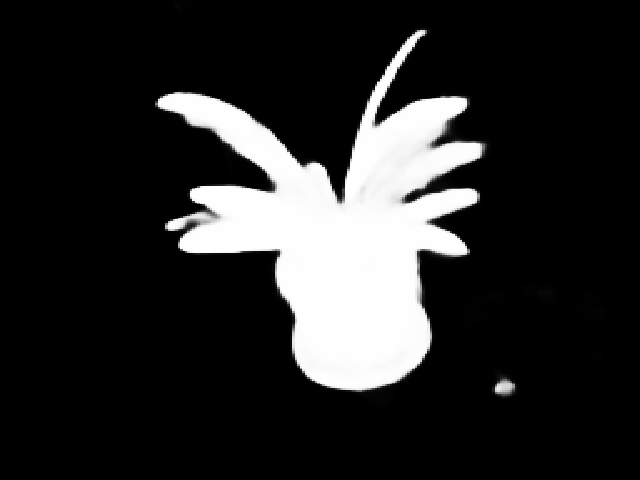}}&
  {\includegraphics[width=0.116\linewidth]{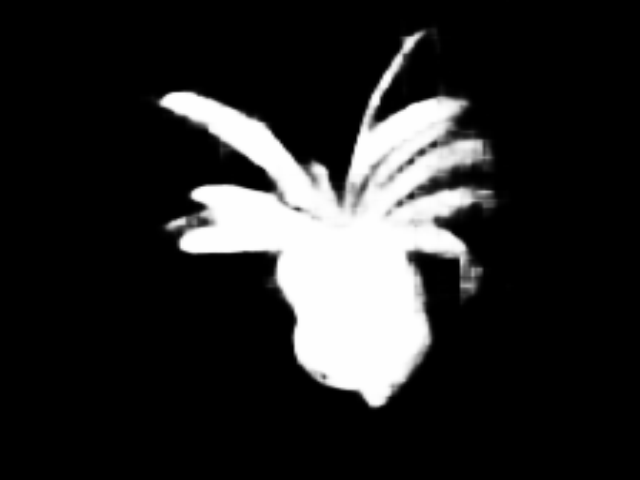}}&
  {\includegraphics[width=0.116\linewidth]{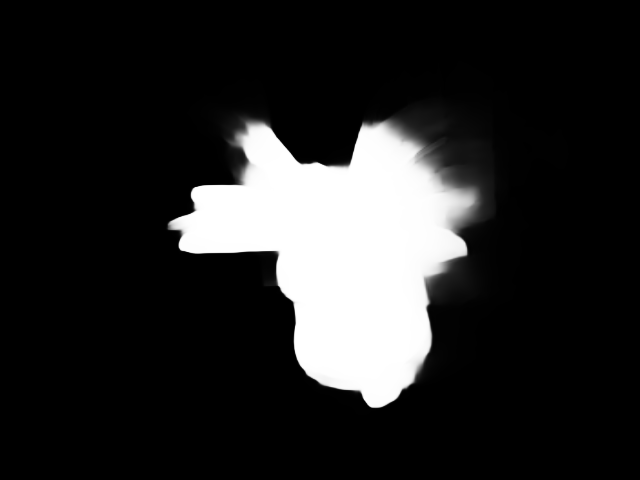}}&
  {\includegraphics[width=0.116\linewidth]{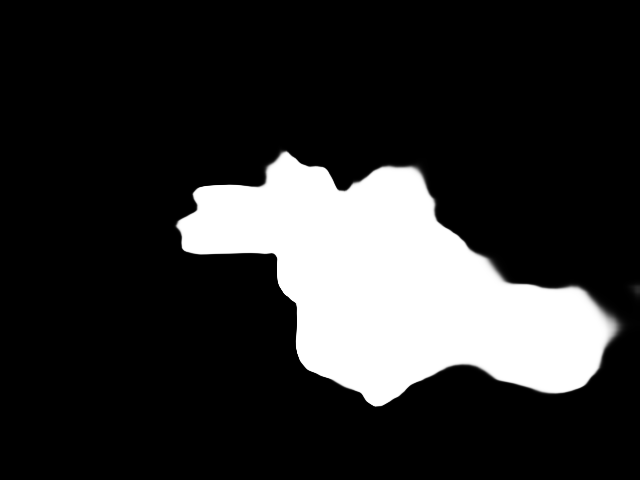}}&
  {\includegraphics[width=0.116\linewidth]{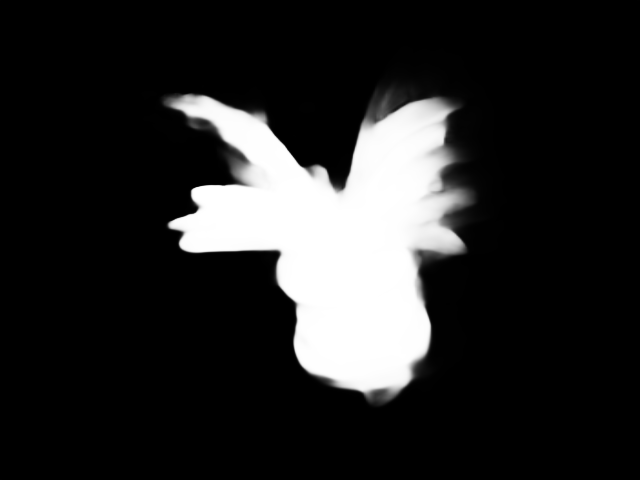}} \\
  {\includegraphics[width=0.116\linewidth]{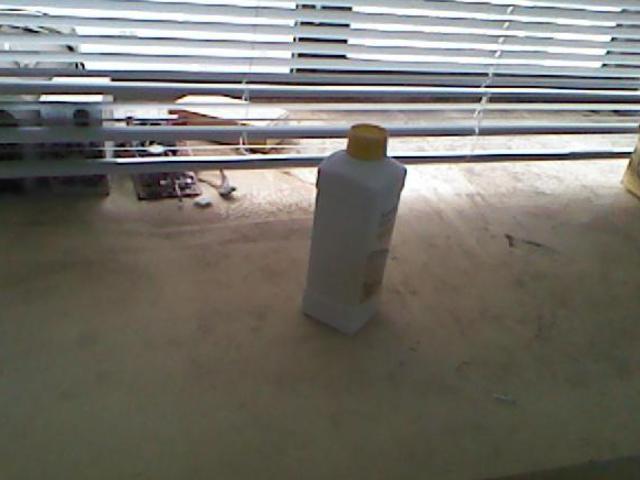}}&
  {\includegraphics[width=0.116\linewidth]{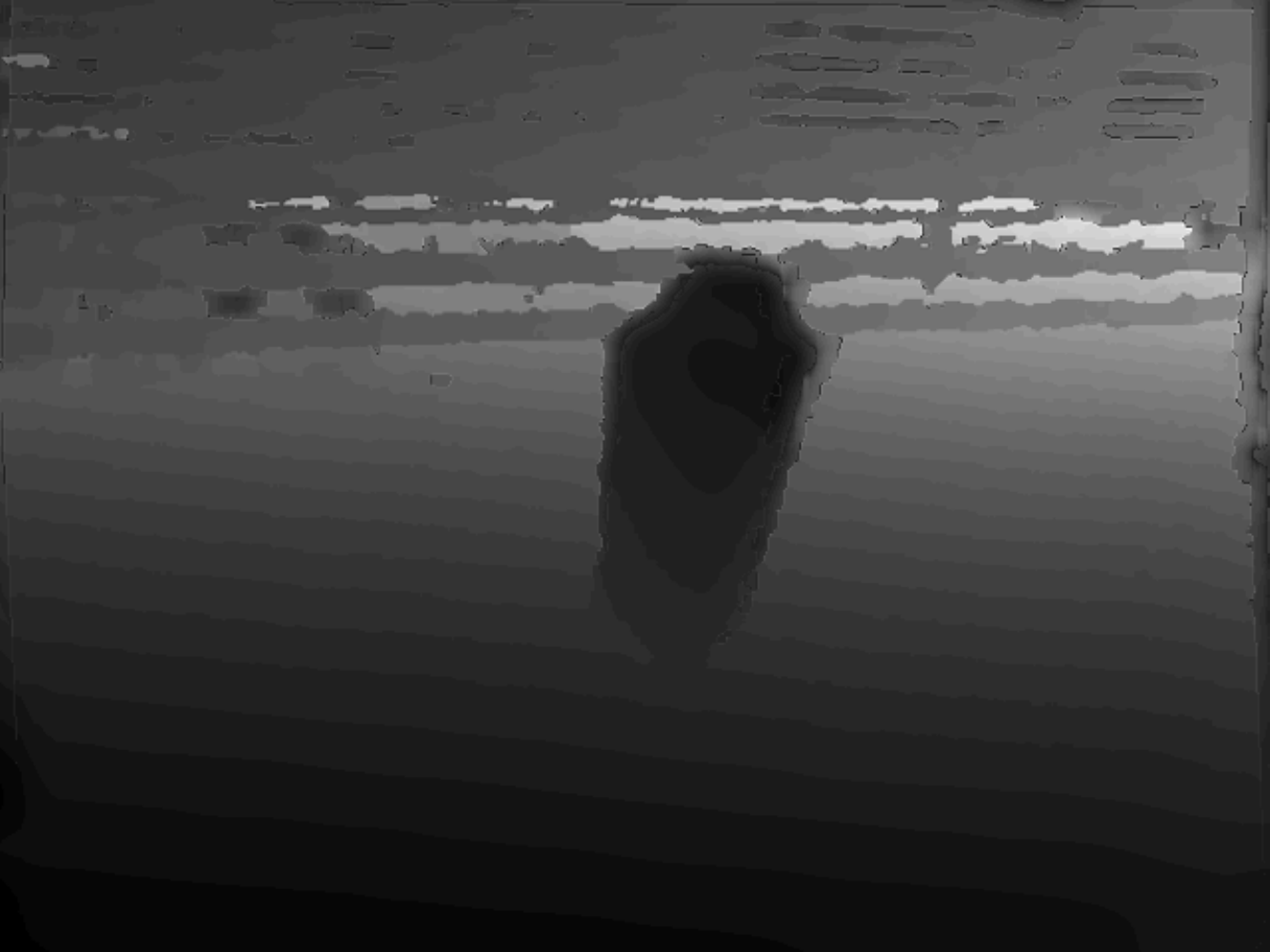}}&
  {\includegraphics[width=0.116\linewidth]{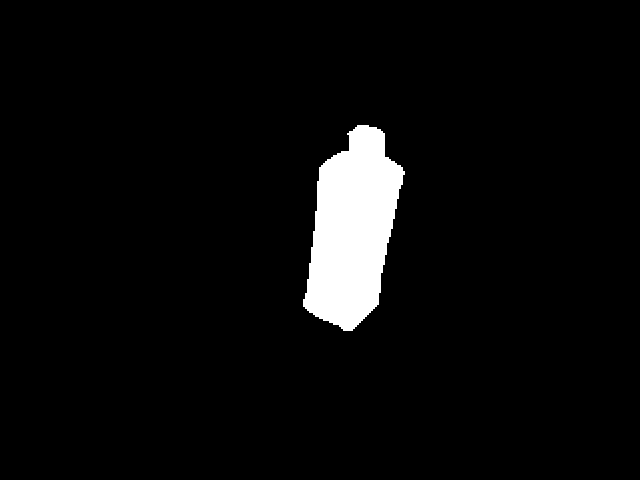}}&
  {\includegraphics[width=0.116\linewidth]{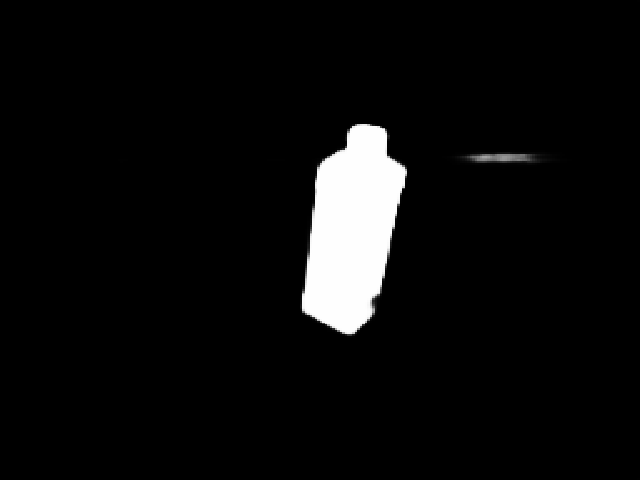}}&
  {\includegraphics[width=0.116\linewidth]{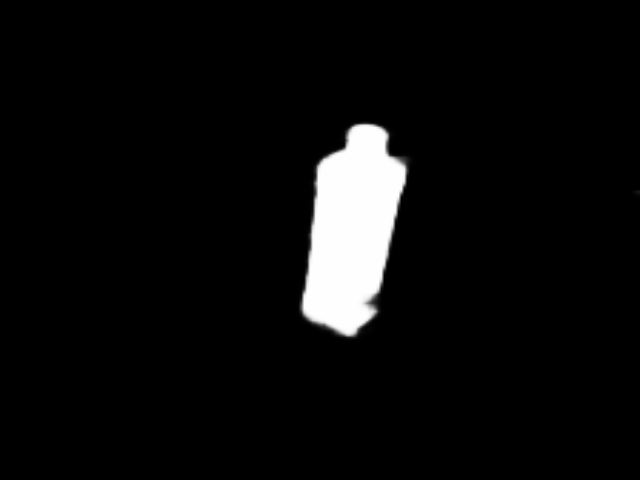}}&
  {\includegraphics[width=0.116\linewidth]{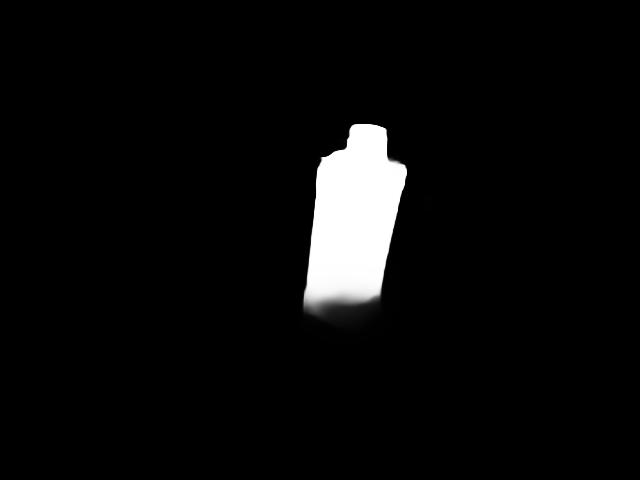}}&
  {\includegraphics[width=0.116\linewidth]{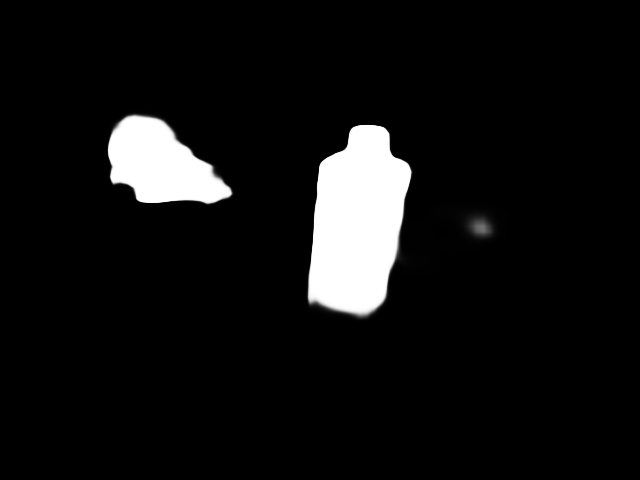}}&
  {\includegraphics[width=0.116\linewidth]{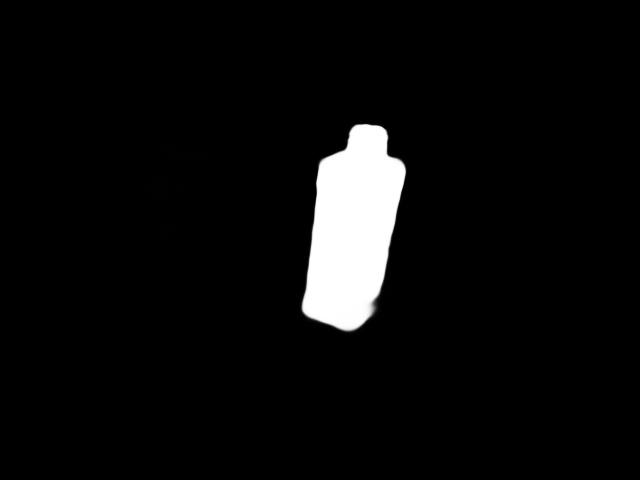}} \\
  {\includegraphics[width=0.116\linewidth]{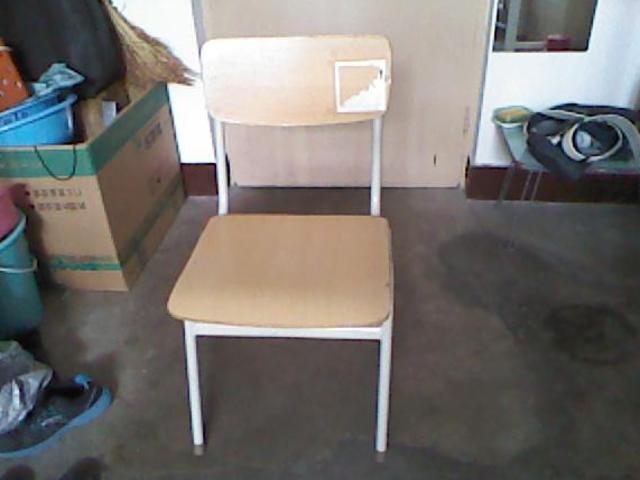}}&
  {\includegraphics[width=0.116\linewidth]{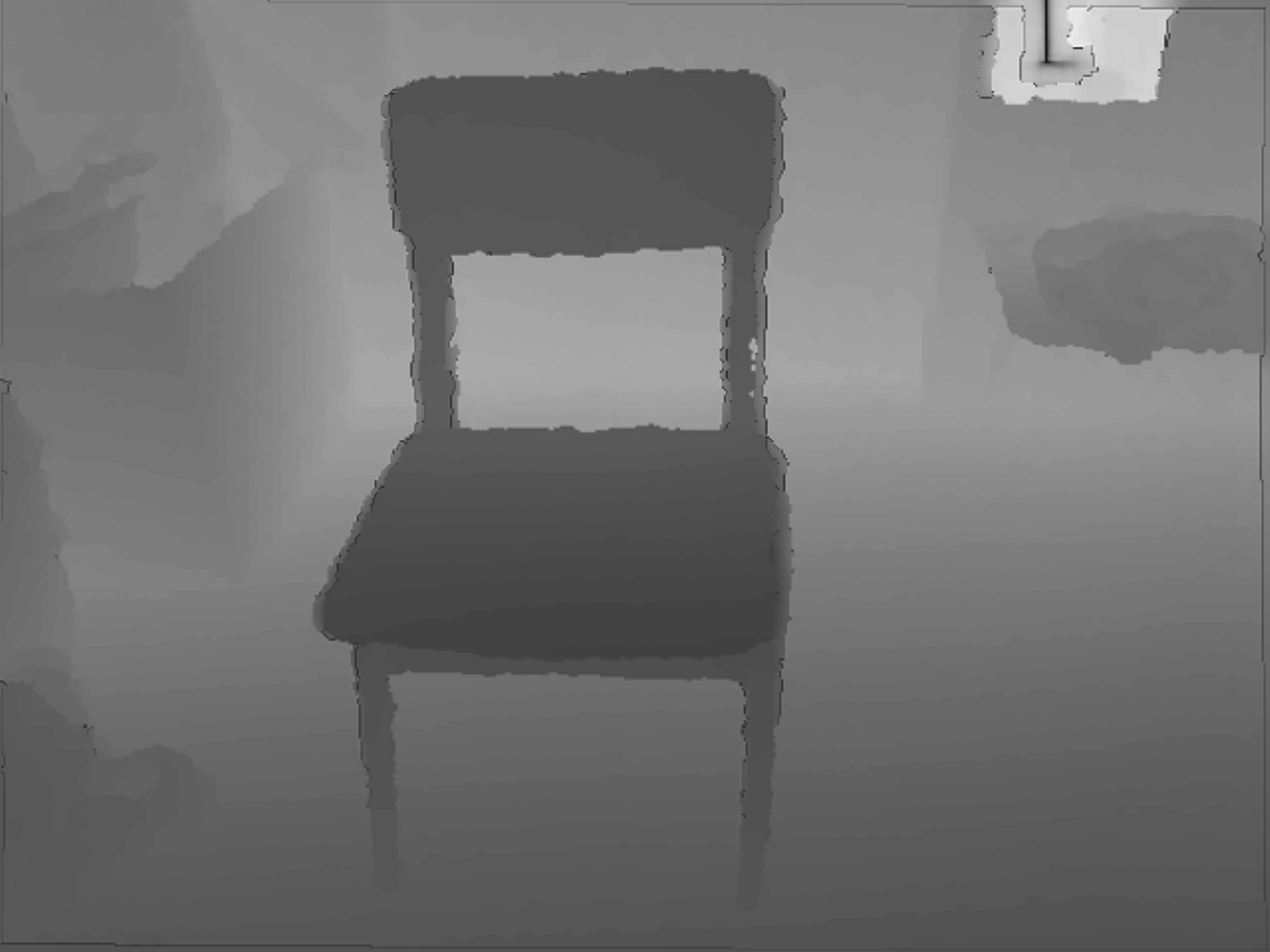}}&
  {\includegraphics[width=0.116\linewidth]{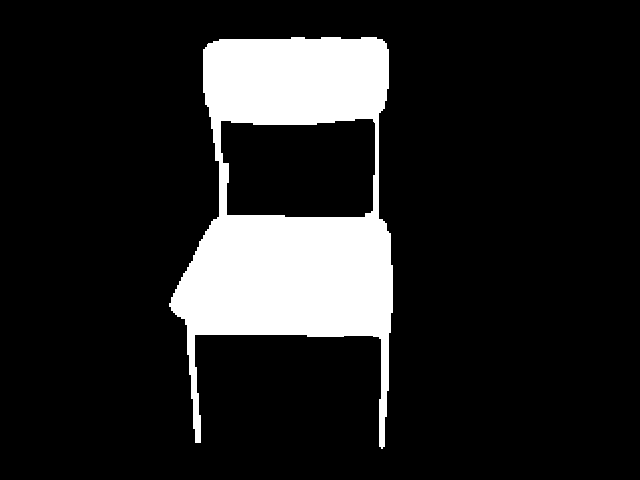}}&
  {\includegraphics[width=0.116\linewidth]{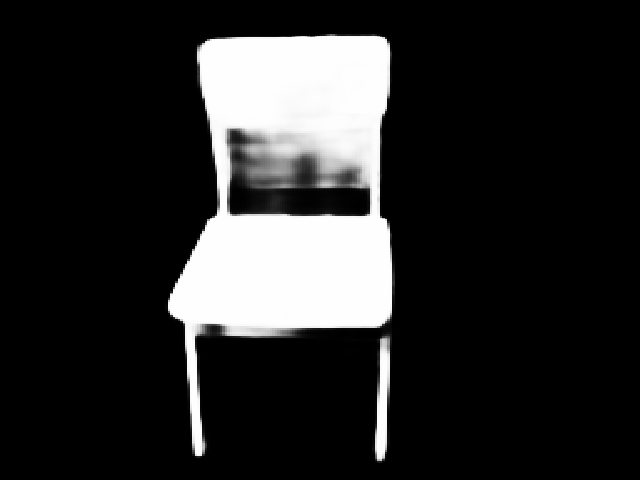}}&
  {\includegraphics[width=0.116\linewidth]{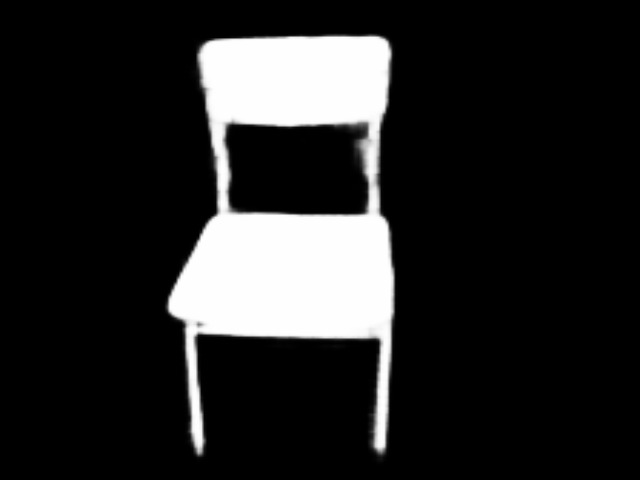}}&
  {\includegraphics[width=0.116\linewidth]{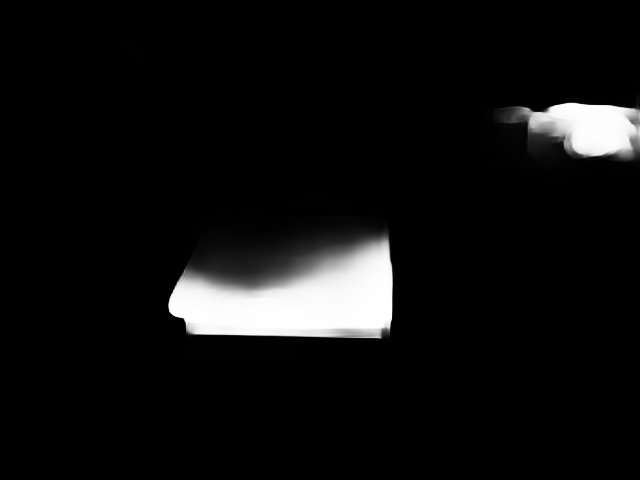}}&
  {\includegraphics[width=0.116\linewidth]{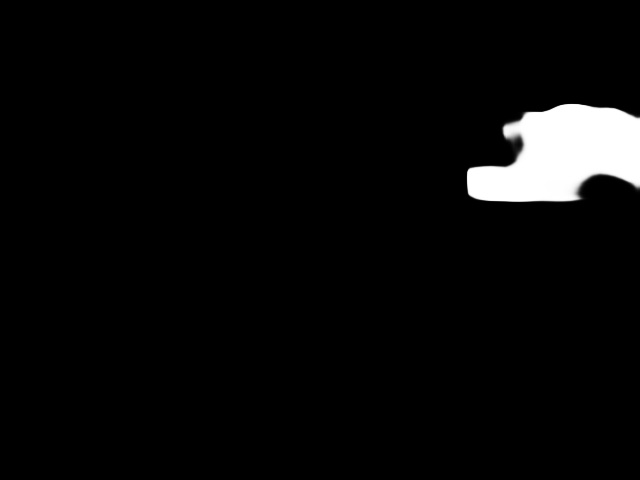}}&
  {\includegraphics[width=0.116\linewidth]{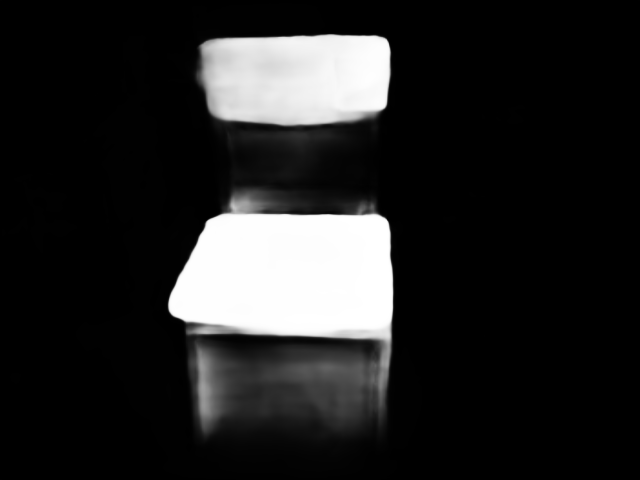}} \\
  \footnotesize{Image}&\footnotesize{Depth}&\footnotesize{GT}& \footnotesize{SSLSOD\cite{SSLSOD}}& \footnotesize{VST\cite{Liu_2021_ICCV_VST}}& \footnotesize{WSSOD\cite{jing2020weakly}}& \footnotesize{SCWS\cite{yu2021structure}}& \footnotesize{Ours}
  \end{tabular}
  \end{center}
   % \vspace{-5mm}
\caption{Visualization of the generated saliency maps from benchmark RGB-D saliency detection models and ours.}
\label{saliency_map_visualization}
\end{figure*}

\begin{table*}[t!]
  \centering
  \scriptsize
  \renewcommand{\arraystretch}{1.25}
  \renewcommand{\tabcolsep}{0.55mm}
  \caption{Cross-model fusion analysis by comparing with both the base RGB model and base RGB-D model via early-fusion.
%   where  $B_\text{rgb}$ means the base model using only RGB modalities. $B_\text{rgbd}$ represents the early-fusion base RGB-D model. 
% $r\_d$ represent the model with our proposed cross-level fusion strategy.
%   And the multimodal fusion can be achieved between RGB and depth ($r\_d$) . % or RGBand RGB-D data ($r\_rd$)
  }
  \begin{tabular}{l|cccc|cccc|cccc|cccc|cccc|cccc}
  \hline
%   \toprule
  &\multicolumn{4}{c|}{NJU2K\cite{NJU2000}}&\multicolumn{4}{c|}{SSB \cite{niu2012leveraging}}&\multicolumn{4}{c|}{DES \cite{cheng2014depth}}&\multicolumn{4}{c|}{NLPR \cite{peng2014rgbd}}&\multicolumn{4}{c|}{LFSD \cite{li2014saliency}}&\multicolumn{4}{c}{SIP \cite{sip_dataset}} \\
    Method 
    & $S_{\alpha}\uparrow$ & $F_{\beta}\uparrow$ & $E_{\xi}\uparrow$ & $\mathcal{M}\downarrow$
    & $S_{\alpha}\uparrow$ & $F_{\beta}\uparrow$ & $E_{\xi}\uparrow$ & $\mathcal{M}\downarrow$
    & $S_{\alpha}\uparrow$ & $F_{\beta}\uparrow$ & $E_{\xi}\uparrow$ & $\mathcal{M}\downarrow$
    & $S_{\alpha}\uparrow$ & $F_{\beta}\uparrow$ & $E_{\xi}\uparrow$ & $\mathcal{M}\downarrow$
    & $S_{\alpha}\uparrow$ & $F_{\beta}\uparrow$ & $E_{\xi}\uparrow$ & $\mathcal{M}\downarrow$
    & $S_{\alpha}\uparrow$ & $F_{\beta}\uparrow$ & $E_{\xi}\uparrow$ & $\mathcal{M}\downarrow$ \\
  \hline
%   SSOD & .899 & .878 & .922 & .044 & .887 & .851 & .914 & .044 & .916 & .887 & .939 & .029 & .913 & .874 & .936 & .028 & .832 & .814 & .860 & .085 & .865 & .841 & .900 & .056    \\ \hline
%   SSOD\_noboost & .877 & .864 & .922 & .051 & .876 & .854 & .926 & .048 & .885 & .861 & .921 & .031 & .900 & .869 & .944 & .031 & .826 & .815 & .874 & .086 & .846 & .827 & .905 & .063    \\ \hline
 $B_\text{rgb}$ &.810 & .782 & .878 & .079 & .820 & .786 & .892 & .070 & .784 & .722 & .858 & .062 & .815 & .749 & .893 & .054 & .740 & .728 & .815 & .127 & .784 & .745 & .866 & .089 \\
 $B_\text{rgbd}$ &.809 & .785 & .877 & .080 & .801 & .758 & .876 & .080 & .846 & .797 & .924 & .043 & .822 & .757 & .897 & .051 & .785 & .775 & .852 & .104 & .786 & .758 & .869 & .089\\ 
%  \hline
 $C_\text{rgbd}$ & .839 & .807 & .894 & .068 & .831 & .783 & .890 & .070 & .873 & .825 & .929 & .035 & .842 & .769 & .905 & .048 & .792 & .773 & .843 & .110 & .822 & .793 & .889 & .075 \\   
%  $r\_rd$ & .838 & .804 & .893 & .069 & .831 & .782 & .890 & .070 & .858 & .803 & .923 & .043 & .828 & .745 & .893 & .052 & .785 & .757 & .836 & .112 & .816 & .782 & .883 & .079 \\
%  $cF$ & .829 & .811 & .895 & .069 & .832 & .802 & .904 & .064 & .822 & .783 & .883 & .050 & .850 & .802 & .922 & .043 & .785 & .782 & .855 & .104 & .811 & .794 & .893 & .073 \\  
%   $rR\_dR$ & .883 & .867 & .923 & .050 & .879 & .856 & .924 & .049 & .915 & .900 & .953 & .023 & .907 & .878 & .947 & .028 & .827 & .815 & .864 & .088 & .853 & .827 & .901 & .063 \\  
%   $rV\_dV$ & .886 & .872 & .926 & .048 & .889 & .866 & .931 & .045 & .901 & .885 & .935 & .027 & .901 & .868 & .942 & .031 & .823 & .815 & .863 & .089 & .854 & .838 & .903 & .062 \\  
%   $rR\_dV$ & .888 & .878 & .927 & .047 & .894 & .877 & .934 & .043 & .906 & .889 & .938 & .026 & .911 & .884 & .947 & .027 & .836 & .827 & .873 & .084 & .865 & .848 & .914 & .055 \\  
%   $rV\_dR$ & .886 & .871 & .926 & .047 & .885 & .857 & .926 & .047 & .912 & .893 & .952 & .025 & .907 & .877 & .948 & .029 & .823 & .817 & .863 & .091 & .849 & .822 & .901 & .064  \\  
% $rR\_rdR$ & .884 & .867 & .923 & .050 & .884 & .855 & .924 & .049 & .908 & .884 & .947 & .028 & .902 & .870 & .943 & .031 & .824 & .805 & .862 & .092 & .855 & .830 & .903 & .062 \\
% $rV\_rdV$ & \\
% $rR\_rdV$ & .889 & .874 & .925 & .048 & .887 & .859 & .926 & .047 & .895 & .868 & .926 & .034 & .909 & .877 & .945 & .028 & .836 & .820 & .870 & .086 & .851 & .818 & .897 & .066 \\
% $rV\_rdR$ &  \\
   \hline
  \end{tabular}
  \label{tab:cross_level_fusion_analysis}
%   \vspace{-5mm}
\end{table*}

\begin{table*}[t!]
  \centering
  \scriptsize
  \renewcommand{\arraystretch}{1.25}
  \renewcommand{\tabcolsep}{0.55mm}
  \caption{Backbone analysis to verify the effectiveness of the proposed asymmetric feature extractors.
  }
  \begin{tabular}{l|cccc|cccc|cccc|cccc|cccc|cccc}
  \hline
%   \toprule
  &\multicolumn{4}{c|}{NJU2K\cite{NJU2000}}&\multicolumn{4}{c|}{SSB \cite{niu2012leveraging}}&\multicolumn{4}{c|}{DES \cite{cheng2014depth}}&\multicolumn{4}{c|}{NLPR \cite{peng2014rgbd}}&\multicolumn{4}{c|}{LFSD \cite{li2014saliency}}&\multicolumn{4}{c}{SIP \cite{sip_dataset}} \\
    Method 
    & $S_{\alpha}\uparrow$ & $F_{\beta}\uparrow$ & $E_{\xi}\uparrow$ & $\mathcal{M}\downarrow$
    & $S_{\alpha}\uparrow$ & $F_{\beta}\uparrow$ & $E_{\xi}\uparrow$ & $\mathcal{M}\downarrow$
    & $S_{\alpha}\uparrow$ & $F_{\beta}\uparrow$ & $E_{\xi}\uparrow$ & $\mathcal{M}\downarrow$
    & $S_{\alpha}\uparrow$ & $F_{\beta}\uparrow$ & $E_{\xi}\uparrow$ & $\mathcal{M}\downarrow$
    & $S_{\alpha}\uparrow$ & $F_{\beta}\uparrow$ & $E_{\xi}\uparrow$ & $\mathcal{M}\downarrow$
    & $S_{\alpha}\uparrow$ & $F_{\beta}\uparrow$ & $E_{\xi}\uparrow$ & $\mathcal{M}\downarrow$ \\
  \hline
%   SSOD & .899 & .878 & .922 & .044 & .887 & .851 & .914 & .044 & .916 & .887 & .939 & .029 & .913 & .874 & .936 & .028 & .832 & .814 & .860 & .085 & .865 & .841 & .900 & .056    \\ \hline
%   SSOD\_noboost & .877 & .864 & .922 & .051 & .876 & .854 & .926 & .048 & .885 & .861 & .921 & .031 & .900 & .869 & .944 & .031 & .826 & .815 & .874 & .086 & .846 & .827 & .905 & .063    \\ \hline
$r\_d(RV)$ & .852 & .827 & .906 & .063 & .845 & .803 & .900 & .066 & .900 & .858 & .949 & .033 & .865 & .810 & .922 & .042 & .802 & .779 & .847 & .106 & .839 & .815 & .902 & .067 \\ \hline
%%%% without refined with HA and tree
$rR\_dR$ & .849 & .817 & .896 & .066 & .843 & .793 & .894 & .068 & .898 & .860 & .944 & .031 & .861 & .796 & .916 & .042 & .798 & .771 & .836 & .112 & .844 & .816 & .901 & .067 \\
% $rR\_dV$ & .852 & .827 & .906 & .063 & .845 & .803 & .900 & .066 & .900 & .858 & .949 & .033 & .865 & .810 & .922 & .042 & .802 & .779 & .847 & .106 & .839 & .815 & .902 & .067 \\
$rV\_dR$ & .838 & .805 & .894 & .070 & .829 & .778 & .885 & .071 & .881 & .827 & .936 & .035 & .844 & .772 & .903 & .047 & .781 & .752 & .831 & .113 & .827 & .791 & .891 & .073 \\
$rV\_dV$ &.836 & .802 & .893 & .068 & .832 & .786 & .893 & .068 & .870 & .818 & .932 & .038 & .833 & .756 & .895 & .050 & .789 & .765 & .842 & .111 & .817 & .784 & .886 & .077 \\

  \hline
  \end{tabular}
  \label{tab:ablation_asymmetric_backbone_analysis}
%   \vspace{-5mm}
\end{table*}

\section{Experimental Results}
\noindent\textbf{Dataset:} For fair comparisons with existing RGB-D saliency detection models, we follow the conventional training setting, in which the training set is a combination of 1,485 images from the NJU2K
% \fdp{NJN2K$->$NJU2K} 
dataset \cite{NJU2000} and 700 images from the NLPR dataset \cite{peng2014rgbd}. We then test the performance of our model and competing models on the NJU2K testing set, NLPR, testing set 
LFSD \cite{li2014saliency}, DES \cite{cheng2014depth}, SSB \cite{niu2012leveraging} SIP \cite{sip_dataset} and DUT \cite{dmra_iccv19} testing set.

\noindent\textbf{Evaluation Metrics:} We evaluate the performance of the models on four golden evaluation metrics, \ie Mean Absolute Error ($\mathcal{M}$), Mean F-measure ($F_{\beta}$), Mean E-measure ($E_{\xi}$)~\cite{Fan2018Enhanced} and S-measure ($S_{\alpha}$)~\cite{fan2017structure}.

\noindent\textbf{Implementation Details:} Our model is implemented using \textit{Pytorch}.
For the first training stage, the RGB encoder is initialized with ResNet50~\cite{ResHe2015} backbone and
the depth encoder is initialized with VGG16-Net~\cite{VGG}. Both of the two backbones are
% trained on ImageNet\cite{Imagenet}, the depth encoder are 
initialized with the corresponding parameters
% ResNet50\cite{ResHe2015} 
trained on ImageNet\cite{Imagenet}. We set the maximum epoch as 30, and the initial learning rate as $3\times 10^{-5}$. 
For the VAE refinement stage, both encoders are initialized with ResNet50~\cite{ResHe2015}, and set the maximum epoch as 50, and the initial learning rate is $2.5\times 10^{-5}$. 
The other newly added layers are initialized by default. For both stages, We resize all the images and ground truth to the same spatial size of $352\times352$ pixels. We adopt the \enquote{step} learning rate decay policy, and set the decay size as 1 and decay rate as 0.95. The models are trained with batch size 3 on an NVIDIA GeForce RTX 3090 GPU, and the training takes 5.5 hours, VAE refinement stage takes 7.5 hours.
The features and predictions were optimized by the holistic attention \cite{cpd_sal} to obtain finer predictions that were also generated in each mode, but were omitted for brevity.
% For the first training stage, the RGB encoders are initialized with VGG16-Net\cite{VGG} trained on ImageNet\cite{Imagenet}, the depth encoder are initialized with ResNet50\cite{ResHe2015} trained on ImageNet\cite{Imagenet}. And the other newly added layers are initialized by default. We resize all the images and ground truth to the same spatial size of $352\times352$ pixels. We set the maximum epoch as 30, and initial learning rate as 3e-5. We adopt the \enquote{step} learning rate decay policy, and set the decay size as 1 and decay rate as 0.95. % The whole training takes 5.5 hours with batch size 3 on an NVIDIA GeForce RTX 3090 GPU. % ResNet50 \cite{ResHe2015}
% For VAE refinement stage, both encoder are initialized with ResNet50\cite{ResHe2015}. We set the maximum epoch as 50, initial learning rate as 2.5e-5, and the other settings are consistent with the first training stage.
% The training takes 5.5 hours with batch size 3 for the first stage, and takes 7.5 hours with batch size 3 for VAE refinement stage on an NVIDIA GeForce RTX 3090 GPU.

\subsection{Performance comparison}
\noindent\textbf{Quantitative comparison:} We compare our method with state-of-the-art fully-supervised and weakly-supervised RGB-D saliency detection models, and show performance in Table~\ref{tab:BenchmarkResults_old_training}. As a comparison method, DSU \cite{ji2022promoting} is an unsupervised method that disentangles the saliency regions from the depth and RGB images. JSM \cite{li2021joint} uses the traditional SOD algorithm as a pseudo-label and updates the annotation with the help of semantic labels. SCWS \cite{yu2021structure}, SSOD \cite{xu2022weakly} and WSSOD \cite{jing2020weakly} are based on scribble annotations. The results show that we achieve the best results on 5/6 dataset, except the NJU2K dataset \cite{NJU2000}. The reason is that our mutual information optimization may accidentally encourage noisy depth for the final prediction. Note that both WSSOD \cite{jing2020weakly} and SCWS \cite{yu2021structure} are RGB saliency detection models with scribble supervision. We directly perform early fusion and introduce another $3\times 3$ convolutional layers to map the RGB-D data to 3 channel feature map to fit their original implementations.
% some disturbing depth information, and SSOD \cite{xu2022weakly} provides extra scribble annotations in the training phase by active learning.
%\Jing{introduce the settings of those weakly supervised saliency detection models in Table~\ref{tab:BenchmarkResults_old_training} and compare both performance and setting with them.}

\noindent\textbf{Qualitative comparison:} We also show the generated saliency maps of our method and existing techniques in Fig.~\ref{saliency_map_visualization}. It can be observed that our model makes effective use of depth information and learns more details than other weakly supervised methods.

\begin{table*}[t!]
  \centering
  \scriptsize
  \renewcommand{\arraystretch}{1.25}
  \renewcommand{\tabcolsep}{0.55mm}
  \caption{Module effectiveness analysis, including the progressive fusion module and the mutual information minimization regularization.
  $r\_d$ means the baseline with the asymmetric feature extractor( same as $r\_d(RV)$ in Table ~\ref{tab:ablation_asymmetric_backbone_analysis}). $pr\_rd$ means the result of the graph based progressive refinement module with tree-energy loss~\cite{liang2022tree}. $pr\_lmi$ means mutual information approximation from \cite{cascaded_rgbd_sod}. $pr\_mi$ is the result based on
   $pr\_rd$ with the mutual information regularization term.
%   and the bi-directional version ($pr\_bmi$).
  }
  \begin{tabular}{l|cccc|cccc|cccc|cccc|cccc|cccc}
  \hline
%   \toprule
  &\multicolumn{4}{c|}{NJU2K\cite{NJU2000}}&\multicolumn{4}{c|}{SSB \cite{niu2012leveraging}}&\multicolumn{4}{c|}{DES \cite{cheng2014depth}}&\multicolumn{4}{c|}{NLPR \cite{peng2014rgbd}}&\multicolumn{4}{c|}{LFSD \cite{li2014saliency}}&\multicolumn{4}{c}{SIP \cite{sip_dataset}} \\
    Method 
    & $S_{\alpha}\uparrow$ & $F_{\beta}\uparrow$ & $E_{\xi}\uparrow$ & $\mathcal{M}\downarrow$
    & $S_{\alpha}\uparrow$ & $F_{\beta}\uparrow$ & $E_{\xi}\uparrow$ & $\mathcal{M}\downarrow$
    & $S_{\alpha}\uparrow$ & $F_{\beta}\uparrow$ & $E_{\xi}\uparrow$ & $\mathcal{M}\downarrow$
    & $S_{\alpha}\uparrow$ & $F_{\beta}\uparrow$ & $E_{\xi}\uparrow$ & $\mathcal{M}\downarrow$
    & $S_{\alpha}\uparrow$ & $F_{\beta}\uparrow$ & $E_{\xi}\uparrow$ & $\mathcal{M}\downarrow$
    & $S_{\alpha}\uparrow$ & $F_{\beta}\uparrow$ & $E_{\xi}\uparrow$ & $\mathcal{M}\downarrow$ \\
  \hline
  $r\_d$ & .852 & .827 & .906 & .063 & .845 & .803 & .900 & .066 & .900 & .858 & .949 & .033 & .865 & .810 & .922 & .042 & .802 & .779 & .847 & .106 & .839 & .815 & .902 & .067 \\ \hline
%   $Ours$ & . & . & . & . & . & . & . & . & . & . & . & . & \\ \hline
%   $p\_rd$ & .849 & .819 & .899 & .064 & .841 & .795 & .897 & .066 & .886 & .836 & .944 & .033 & .848 & .775 & .905 & .046 & .801 & .773 & .842 & .109 & .827 & .790 & .888 & .076 \\
  $pr\_rd$ & .891 & .879 & .929 & .046 & .887 & .862 & .926 & .047 & .917 & .901 & .953 & .023 & .909 & .881 & .945 & .029 & .834 & .819 & .865 & .088 & .860 & .835 & .906 & .060 \\ 
%   $pr\_mi$ & .894 & .887 & .932 & .044 & .890 & .869 & .931 & .045 & .908 & .884 & .942 & .026 & .913 & .891 & .951 & .026 & .827 & .816 & .863 & .089 & .860 & .843 & .907 & .058 \\ %0.003
%   $pr\_mi$ &  .891 & .880 & .927 & .046 & .892 & .870 & .929 & .044 & .918 & .901 & .952 & .024 & .906 & .882 & .942 & .028 & .829 & .816 & .864 & .086 & .866 & .846 & .911 & .057 \\%0.005
%   $pr\_bmi$ & .891 & .882 & .928 & .046 & .890 & .869 & .928 & .045 & .905 & .892 & .936 & .027 & .906 & .884 & .947 & .028 & .834 & .821 & .867 & .087 & .861 & .839 & .905 p` & .060\\  %0.003

  $pr\_lmi$ & .883 & .863 & .921 & .049 & .884 & .852 & .921 & .050 & .912 & .888 & .948 & .026 & .906 & .873 & .942 & .030 & .828 & .808 & .856 & .088 & .858 & .832 & .899 & .063\\
%   $pr\_bmi$ & .890 & .885 & .928 & .046 & .888 & .871 & .929 & .045 & .916 & .904 & .954 & .024 & .901 & .878 & .938 & .030 & .824 & .813 & .864 & .086 & .859 & .846 & .905 & .060\\  %0.005
$pr\_mi$ &.891 & .880 & .927 & .046 & .892 & .870 & .929 & .044 & .918 & .901 & .952 & .024 & .906 & .882 & .942 & .028 & .829 & .816 & .864 & .086 & .866 & .846 & .911 & .057\\
%   $pr\_bmi$ & .890 & .885 & .928 & .046 & .888 & .871 & .929 & .045 & .916 & .904 & .954 & .024 & .901 & .878 & .938 & .030 & .824 & .813 & .864 & .086 & .859 & .846 & .905 & .060\\  %0.005
%  $B_\text{rgb}$ &.810 & .782 & .878 & .079 & .820 & .786 & .892 & .070 & .784 & .722 & .858 & .062 & .815 & .749 & .893 & .054 & .740 & .728 & .815 & .127 & .784 & .745 & .866 & .089 \\
%  $B_\text{rgbd}$ &.809 & .785 & .877 & .080 & .801 & .758 & .876 & .080 & .846 & .797 & .924 & .043 & .822 & .757 & .897 & .051 & .785 & .775 & .852 & .104 & .786 & .758 & .869 & .089\\
%   $cF$ & .829 & .811 & .895 & .069 & .832 & .802 & .904 & .064 & .822 & .783 & .883 & .050 & .850 & .802 & .922 & .043 & .785 & .782 & .855 & .104 & .811 & .794 & .893 & .073 \\  \hline
%   $rR\_rdV$ & .889 & .874 & .925 & .048 & .887 & .859 & .926 & .047 & .895 & .868 & .926 & .034 & .909 & .877 & .945 & .028 & .836 & .820 & .870 & .086 & .851 & .818 & .897 & .066\\
%   $rR\_dV$ &  .888 & .878 & .927 & .047 & .894 & .877 & .934 & .043 & .906 & .889 & .938 & .026 & .911 & .884 & .947 & .027 & .836 & .827 & .873 & .084 & .865 & .848 & .914 & .055 \\ 
%   $MiMin$ &  .889 & .880 & .927 & .048 & .893 & .877 & .931 & .044 & .909 & .891 & .948 & .026 & .906 & .882 & .946 & .028 & .833 & .821 & .866 & .086 & .861 & .843 & .908 & .059 \\  
   \hline
  \end{tabular}
  \label{tab:ablation_fusion_mutual_mi_min}
%   \vspace{-5mm}
\end{table*}

\begin{table*}[t!]
  \centering
  \scriptsize
  \renewcommand{\arraystretch}{1.25}
  \renewcommand{\tabcolsep}{0.55mm}
  \caption{Performance of prediction refinement techniques, where $D\_s$ is obtained by directly applying first stage output as pseudo label, and $D\_crf$ is achieved via applying denseCRF~\cite{dense_crf} as post-processing technique.
  }
  \begin{tabular}{l|cccc|cccc|cccc|cccc|cccc|cccc}
  \hline
%   \toprule
  &\multicolumn{4}{c|}{NJU2K\cite{NJU2000}}&\multicolumn{4}{c|}{SSB \cite{niu2012leveraging}}&\multicolumn{4}{c|}{DES \cite{cheng2014depth}}&\multicolumn{4}{c|}{NLPR \cite{peng2014rgbd}}&\multicolumn{4}{c|}{LFSD \cite{li2014saliency}}&\multicolumn{4}{c}{SIP \cite{sip_dataset}} \\
    Method 
    & $S_{\alpha}\uparrow$ & $F_{\beta}\uparrow$ & $E_{\xi}\uparrow$ & $\mathcal{M}\downarrow$
    & $S_{\alpha}\uparrow$ & $F_{\beta}\uparrow$ & $E_{\xi}\uparrow$ & $\mathcal{M}\downarrow$
    & $S_{\alpha}\uparrow$ & $F_{\beta}\uparrow$ & $E_{\xi}\uparrow$ & $\mathcal{M}\downarrow$
    & $S_{\alpha}\uparrow$ & $F_{\beta}\uparrow$ & $E_{\xi}\uparrow$ & $\mathcal{M}\downarrow$
    & $S_{\alpha}\uparrow$ & $F_{\beta}\uparrow$ & $E_{\xi}\uparrow$ & $\mathcal{M}\downarrow$
    & $S_{\alpha}\uparrow$ & $F_{\beta}\uparrow$ & $E_{\xi}\uparrow$ & $\mathcal{M}\downarrow$ \\
  \hline
  $pr\_mi$ &.891 & .880 & .927 & .046 & .892 & .870 & .929 & .044 & .918 & .901 & .952 & .024 & .906 & .882 & .942 & .028 & .829 & .816 & .864 & .086 & .866 & .846 & .911 & .057\\\hline
%   $pr\_bmi$ & .890 & .885 & .928 & .046 & .888 & .871 & .929 & .045 & .916 & .904 & .954 & .024 & .901 & .878 & .938 & .030 & .824 & .813 & .864 & .086 & .859 & .846 & .905 & .060\\ \hline
%  $B_\text{rgb}$ &.810 & .782 & .878 & .079 & .820 & .786 & .892 & .070 & .784 & .722 & .858 & .062 & .815 & .749 & .893 & .054 & .740 & .728 & .815 & .127 & .784 & .745 & .866 & .089 \\
%  $MiMin$ &  .889 & .880 & .927 & .048 & .893 & .877 & .931 & .044 & .909 & .891 & .948 & .026 & .906 & .882 & .946 & .028 & .833 & .821 & .866 & .086 & .861 & .843 & .908 & .059 \\  \hline
%   $pr\_mi$ & .891 & .880 & .927 & .046 & .892 & .870 & .929 & .044 & .918 & .901 & .952 & .024 & .906 & .882 & .942 & .028 & .829 & .816 & .864 & .086 & .866 & .846 & .911 & .057\\
  $D\_s$ &.894 & .885 & .930 & .044 & .888 & .868 & .929 & .045 & .908 & .894 & .942 & .025 & .907 & .883 & .947 & .027 & .837 & .826 & .873 & .085 & .875 & .863 & .919 & .051 \\ % .895 & .886 & .932 & .043 & .890 & .869 & .931 & .044 & .921 & .907 & .962 & .021 & .913 & .890 & .952 & .025 & .830 & .809 & .863 & .090 & .877 & .861 & .921 & .050 \\% .896 & .880 & .926 & .048 & .889 & .854 & .921 & .051 & .919 & .889 & .942 & .026 & .911 & .875 & .943 & .029 & .847 & .828 & .875 & .082 & .873 & .850 & .911 & .057 \\ 
%   $D\_s\_CE$ & \\ % .898 & .876 & .926 & .047 & .887 & .848 & .917 & .052 & .920 & .889 & .947 & .026 & .913 & .875 & .943 & .029 & .846 & .818 & .868 & .088 & .880 & .858 & .915 & .053 \\
  $D\_crf$ &  .885 & .886 & .929 & .045 & .887 & .880 & .930 & .040 & .896 & .892 & .938 & .024 & .897 & .886 & .939 & .027 & .823 & .824 & .863 & .083 & .864 & .855 & .915 & .054\\ \hline
  Ours &.890 & .880 & .929 & .046 & .891 & .873 & .934 & .042 & .928 & .921 & .965 & .019 & .914 & .895 & .953 & .025 & .849 & .844 & .889 & .072 & .876 & .863 & .924 & .049\\  
   \hline
  \end{tabular}
  \label{tab:ablation_refinement}
%   \vspace{-5mm}
\end{table*}

\subsection{Ablation Studies}
% We extensively analyse the proposed framework with the following experiments.
% % , and show performance of the related models in Table~\ref{tab:ablation_study_experiments}. 
% Note that u
Unless stated otherwise, we define the output from the RGB-D branch as model prediction.
% and asymmetric feature extractors are used for multimodal data feature extraction.\Jing{i'm here}

\noindent\textbf{Base models:}
% Given the weakly supervised training dataset $\mathcal{D}$, 
We train two base models with only RGB image used ($B_\text{rgb}$) or both RGB-D data ($B_\text{rgbd}$) used for weakly-supervised RGB-D saliency detection with partial cross-entropy loss.
% leading to the base model $B_\text{rgb}$ for the former, and $B_\text{rgbd}$ for the latter.
Specifically, we use the same ResNet50 encoder and decoder and ResNet50 as discussed in Sec.~\ref{sec:feat_extractors}.
% structure. 
For $B_{rgbd}$, early fusion is used to fuse RGB and depth at the input, and an extra $3\times 3$ convolutional layer is used to map the concatenated feature to a 3-channel feature map, which will then be fed to the encoder-decoder framework.
% Partial cross-entropy loss function is used to train both base models.
Performance is shown in Table~\ref{tab:cross_level_fusion_analysis}, which explains that early fusion can be effective to some extent, especially for DES, NLPR, and LFSD datasets, verifying the effectiveness of depth for RGB-D saliency detection.

\noindent\textbf{Cross-level fusion:}  %  or RGB-RGBD
% With the proposed simple cross-level fusion strategy as discussed in Sec.~\ref{sec:feat_extractors}, we obtain $r\_d$, and show its performance in
With the proposed simple cross-level fusion strategy as discussed in Sec.~\ref{sec:feat_extractors}, we obtain $C_\text{rgbd}$, and show its performance in
% achieve cross-level fusion, and show performance of RGB-D branch as $r\_d$ in 
Table~\ref{tab:cross_level_fusion_analysis}. The significant performance improvement verifies the effectiveness of our cross-level fusion strategy. 
% With For feature fusion, we use VGG as the backbone network to extract RGB and depth features as the base model, and we employ separate five  $3\times 3$  convolutional layers to combine the five layers of information extracted by the VGG encoder. Similarly, we supervise the model using partial cross-entropy loss. The performance is shown in Table~\ref{tab:cross_level_fusion_analysis}.

\noindent\textbf{Asymmetric feature extractor analysis:} 
We discovered that different backbones activate differently to the multimodal data.
% focus on different portions of the image, and the activation maps for depth and RGB images correlate to backbone networks. 
Based on the cross-level fusion model, we compare the backbone configurations
% of different backbones 
for RGB and depth, and show
% and choose ResNet as the RGB encoder and VGG as the depth encoder.  
% The
model performance in Table~\ref{tab:ablation_asymmetric_backbone_analysis}, the activation is shown in Fig.~\ref{backbone_activation_analysis}, where $rR\_dR$ indicates ResNet50 backbone ($R$) for both the RGB image ($r$) and the depth $d$.
The results show that the selection of a suitable backbone facilitates the integration of multimodal information.
The performance improvement of $r\_d(RV)$ compared to other methods demonstrates the effectiveness of our proposed asymmetric feature extractor.
Note that we supervise the above model with partial cross-entropy loss.
% backbone of SCWS~\cite{yu2021structure}: ResNet50 backbone
% backbone of WSSOD~\cite{jing2020weakly}: vgg16 backbone
% backbone of SSOD~\cite{xu2022weakly}: vgg16 backbone of DSU~\cite{ji2022promoting}: ResNet50
% \Ai{compare different fusion method: done}\\
% \Ai{activation map: Done}

\noindent\textbf{Graph based progressive refinement analysis:}
To better fuse RGB and depth structural and localization information on the prediction, we introduce the graph based progressive refinement module with tree-energy loss~\cite{liang2022tree} based on $r\_d$($r\_d(RV)$ in Table \ref{tab:ablation_asymmetric_backbone_analysis}). The performance of the model with tree-energy loss
% \sout{The model performance}
is $pr\_rd$ in Table~\ref{tab:ablation_fusion_mutual_mi_min}. We observe significant performance improvement compared with $r\_d$, verifying its effectiveness.
% , which combine the $Ha$ module 

% the $Ha$ module helps to perform uni-modal and cross-modal feature fusion, and the graph refine module aligns the prediction with the input image and depth features. 

% \Ai{ Add HA, add tree energy loss:done} \\

% \YC{A conclusion is needed for the above experiments.}

% \begin{figure}[!htp]
% \begin{center}
%   \begin{tabular}{c@{ }c@{ }}
%   {\includegraphics[width=0.4\linewidth,height=0.26\linewidth]{figures/latent_dim/nomin_bar.png}}&
%   {\includegraphics[width=0.4\linewidth,height=0.26\linewidth]{figures/latent_dim/min_bar.png}}\\
% \footnotesize{(a) w/o Info} & \footnotesize{(b) w/ Info}
%   \end{tabular}
%   \end{center}
% %   \vspace{-5mm}
% \caption{Feature representation of RGB and RGB-D before (w/o Info) and after (w/ Info) using the mutual information miminization regularization by define the latent dimension as size 2 for easier feature visualization. The red and blue points mean RGB and RGB-D represently.}
% %  \vspace{-5mm}
% \label{mutual_infor_reg_effect}
% \end{figure}
% % \caption{Feature representation of RGB and depth before (w/o min) and after (w/ min) using the mutual information miminization regularization by define the latent dimension as size 2 for easier feature visualization. The red and blue region mean the activation map of depth and rgb represently.}

\noindent\textbf{Mutual information regularization analysis:}
% \Ai{bi-information} \\
Built upon $pr\_rd$, we further encourage disentangled representation across modalities with
% introduce uni-directional
mutual information minimization
% and bi-directional mutual information minimization, 
and show its performance as $pr\_mi$
% the prediction to include additional depth information, we present bi-mutual information regularization, and de, for $pr\_mi$ and 
% $pr\_bmi$ 
in Table~\ref{tab:ablation_fusion_mutual_mi_min}. To explain the superiority of our mutual information upper bound compared with an existing solution in~\cite{cascaded_rgbd_sod},
% that the proposed bi-directional mutual information estimation is a tighter upper bound of cross-modal mutual information, we also perform experiment with mutual information estimation from~\cite{cascaded_rgbd_sod}, and
we replace our $I_{mi}$ term from Eq.~\eqref{final_loss} with mutual information approximation from~\cite{cascaded_rgbd_sod}. The performance is shown as $pr\_{lmi}$ in Table~\ref{tab:ablation_fusion_mutual_mi_min}. 
% To clearly explain the contribution of
% the proposed
% uni/bi-directional 
% mutual information minimization in achieving cross-modal disentangled representation, we define latent feature dimension as 2, and show the feature correlation of RGB image and depth in Fig.~\ref{mutual_infor_reg_effect}, 
The improved performance of $pr\_{mi}$ compared with both $pr\_{rd}$ and $pr\_{lmi}$ explains the contribution of the mutual information regularization term.

\noindent\textbf{Prediction refinement strategies analysis:} 
% \Rev{ After the first phase of training is completed, we are able to obtain coarse pixel-level labels, and this enables the use of the joint probability density of depth and RGB to achieve a balance between the contribution of depth and RGB to the prediction.}
Following the conventional prediction refinement pipeline, we refine the first stage prediction via both pseudo labeling as second stage training and denseCRF~\cite{dense_crf} as post-processing technique, and show the performance in Table~\ref{tab:ablation_refinement} as $D_s$ and $D_{\mathit{crf}}$ respectively.
Similar to the proposed MVAE refinement method, $D_s$ uses trained model prediction  on the training dataset in the first stage as a pseudo-label. The network of $D_s$ uses the same network structure as the first training network. Same as our MVAE refine model, using structure-aware loss function \cite{wei2020f3net} to emphasize the structural details of the predictions.
% and partial cross entropy loss provides definitive supervisor from scribble.
Moreover, we post-process the predictions of $pr\_mi$ with denseCRF~\cite{dense_crf} and compare it with our proposed refinement method, where parameters of denseCRF are set in line with~\cite{melas2022deep}. The results are shown as $D_{\mathit{crf}}$. Table~\ref{tab:ablation_refinement} shows that the proposed solution
% proposed optimization model can 
works better than the existing techniques, achieving boost performance.
% fully exploit the inter-modal correlation information and is robust to the error propagation problem th
% \YC{Conclusions.}
% \Jing{explain how each modal ($D_s$ and $D_{crf}$) is implemented.}

\section{Conclusion}
We introduced a weakly-supervised RGB-D salient object detection model with scribble supervision. Five main strategies are introduced in our framework to achieve accurate predictions. 
Firstly, we find that the RGB and depth modalities respond differently to the same backbone, and we propose to extract features of the two modalities using asymmetric encoders.
Secondly, we design a simple cross-modal fusion module, achieving significant performance improvement compared with the baseline RGB based model and early fusion RGB-D model. Thirdly, we incorporate minimum spanning tree based tree-energy loss into our framework to achieve training time prediction refinement, encouraging unlabeled pixels in model updating. Fourthly, we present mutual information minimization as a regularizer to extensively model the contribution of RGB image and depth data for effective multimodal fusion. Lastly, our stochastic multimodal VAE based prediction refinement strategy is proven effective and robust for our weakly-supervised learning task.

\noindent\textbf{Limitations:}
According to Table~\ref{tab:ablation_fusion_mutual_mi_min}, mutual information minimization fails to consistently improve model performance. We analyze predictions before and after using $\mathcal{L}_{mi}$ and find that noisy depth signal can be enhanced with the proposed regularizer. Our next step is then to further incorporate a noise robust regularizer to prevent the model from being influenced by the noisy modality. 
Comparing \enquote{$D\_crf$} and \enquote{Our} of Table \ref{tab:ablation_refinement}, the performance of our proposed MVAE refinement is slightly lower than CRF refinement for some metrics in the NJU2K and SSB datasets.
% and we show some examples in the last two rows of Fig.~\ref{refine_visualization_init}. 
It can be found that learning-based optimization is a double-edged sword in that MVAE can balance the contribution of RGB and depth to prediction while can also be misled by both of them.
In addition, due to extra parameters involved, MVAE~\cite{vae_bayes_kumar,structure_output,multimodal_generative_models_weakly_learning} will be further investigated to extensively explore its contribution as a prediction refinement technique.

% \newpage
\bibliographystyle{IEEEtran}
\bibliography{Reference}

% \vspace{-20pt}
\begin{IEEEbiography}[{\includegraphics[width=1in,keepaspectratio]{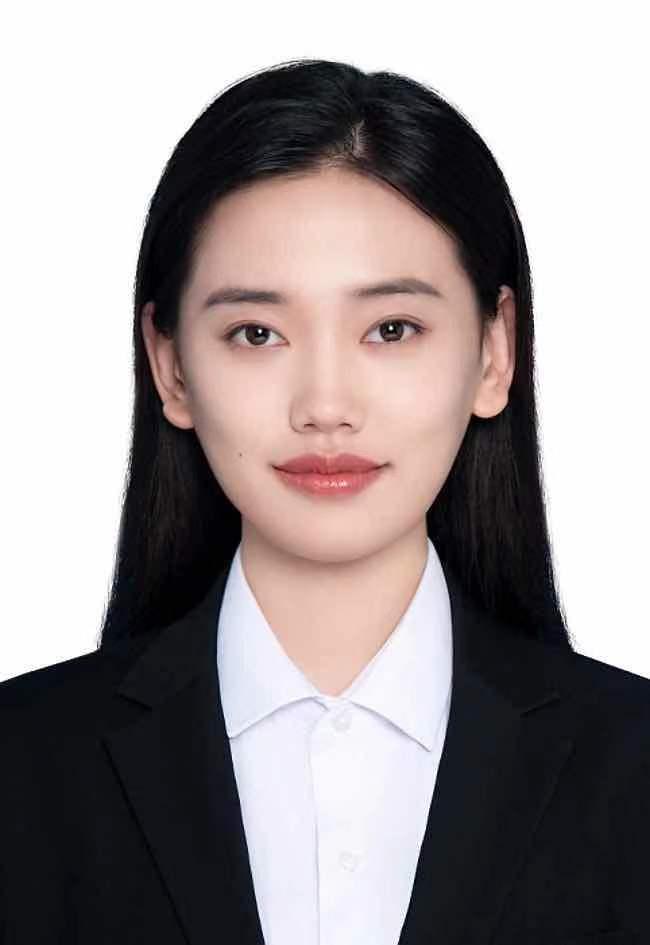}}] {Aixuan Li} currently is a PhD student majoring in signal and information processing in the School of Electronics and Information, Northwestern Polytechnical University. Her main research direction is salient object detection and camouflaged object detection. She received the M.E. degree of signal and information processing from Northwestern Polytechnical University, 2022.
\end{IEEEbiography}
% \begin{IEEEbiographynophoto}{\includegraphics[width=1in,keepaspectratio]{figures/author/aixuan_li.jpg}}
% {Aixuan Li} currently is a PhD student majoring in signal and information processing in the School of Electronics and Information, Northwestern Polytechnical University. Her main research direction is salient object detection and camouflaged object detection. She received the M.E. degree of signal and information processing from Northwestern Polytechnical University, 2021.

\begin{IEEEbiography}[{\includegraphics[width=1in,keepaspectratio]{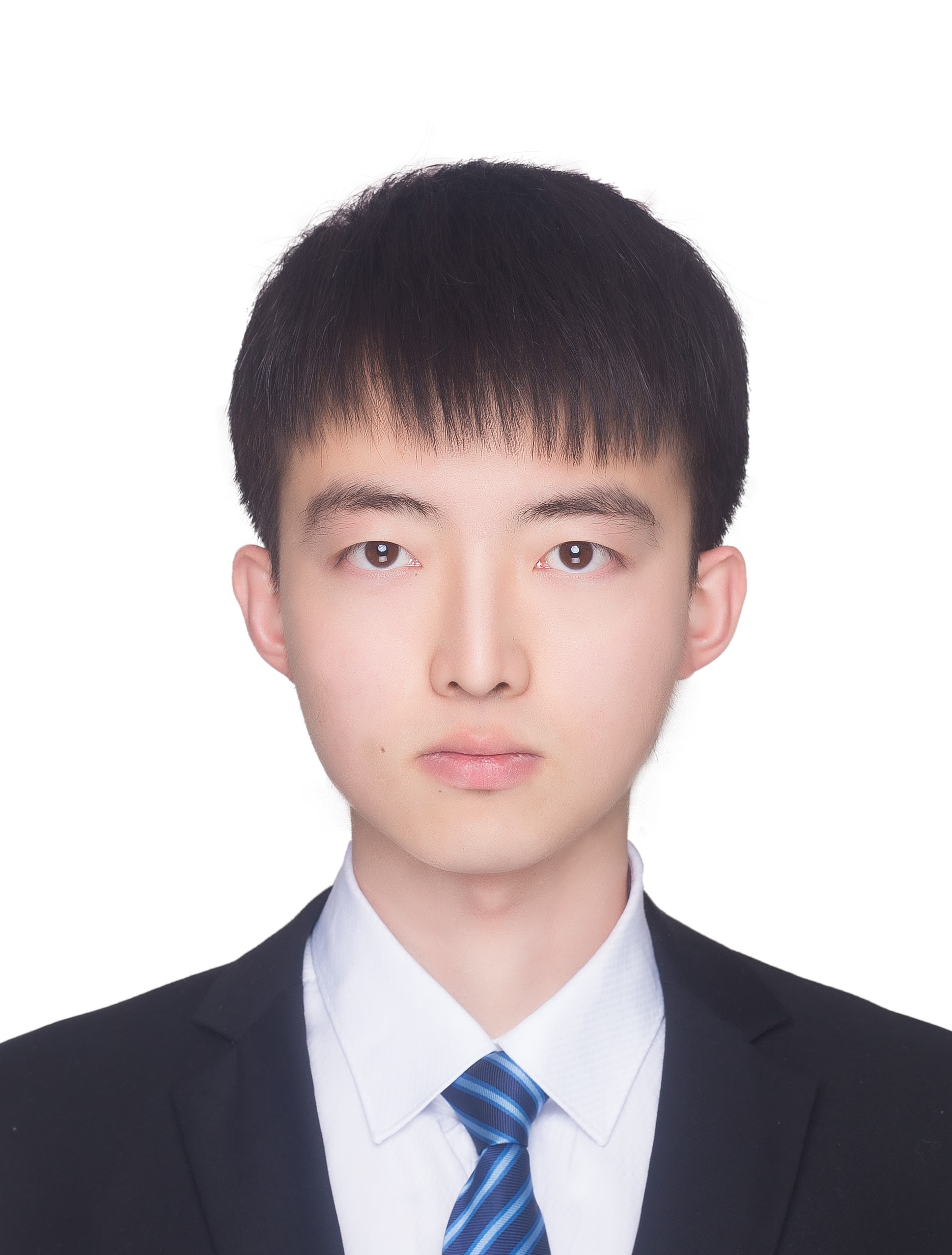}}]{Yuxin Mao}
 is currently a PhD student with School of Electronics and Information, Northwestern Polytechnical University, Xi'an, China. He received his Bachelor of Engineering degree from Southwest Jiaotong University in 2020. He won the best Paper Award Nominee at ICIUS 2019.
\end{IEEEbiography}

% \vspace{-20pt}
\begin{IEEEbiography}[{\includegraphics[width=1in,keepaspectratio]{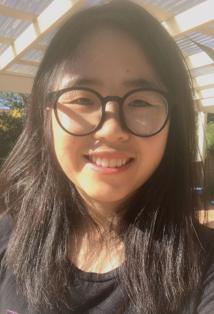}}]{Jing Zhang} is currently a Lecturer with School of Computing, the Australian National University (Canberra, Australia). Her main research interests include saliency detection, weakly supervised learning, generative models. She won the Best Student Paper Prize at DICTA 2017, the Best Deep/Machine Learning Paper Prize at APSIPA ASC 2017 and the Best Paper Award Nominee at IEEE CVPR 2020.
\end{IEEEbiography}

% \vspace{-20pt}
 \begin{IEEEbiography}[{\includegraphics[width=1in,keepaspectratio]{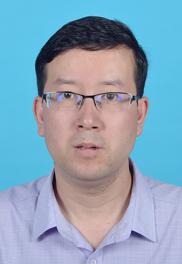}}]{Yuchao Dai}
 is currently a Professor with School of Electronics and Information at the Northwestern Polytechnical University (NPU). He received the B.E. degree, M.E degree and Ph.D. degree all in signal and information processing from NPU, Xi'an, China, in 2005, 2008 and 2012, respectively. He was an ARC DECRA Fellow with the Research School of Engineering at the Australian National University, Canberra, Australia.  His research interests include 3D vision, multi-view geometry, low-level vision, deep learning, and optimization. He won the Best Paper Award in IEEE CVPR 2012, Best Paper Nominee in IEEE CVPR 2020, the DSTO Best Fundamental Contribution to Image Processing Paper Prize at DICTA 2014, the Best Algorithm Prize in NRSFM Challenge at CVPR 2017, the Best Student Paper Prize at DICTA 2017 and the Best Deep/Machine Learning Paper Prize at APSIPA ASC 2017. He served/serves as Area Chair at CVPR, ICCV, NeurIPS, ACM MM, etc. He serves as the Publicity Chair at ACCV 2022 and the Distinguished Lecturer of APSIPA.
 \end{IEEEbiography}

\end{document}